%% file: main.tex
\documentclass{article} %
\usepackage[accepted]{icml2024}

\input{./sections/0_preamble.tex}

\usepackage{subfigure,wrapfigure}

\usepackage{etoc}
\etocdepthtag.toc{mtchapter}
\etocsettagdepth{mtchapter}{subsection}
\etocsettagdepth{mtappendix}{none}

\definecolor{purple}{HTML}{7D2882}
\usepackage{colortbl}
\definecolor{lightskyblue}{HTML}{D3D3D3}

\newcommand{\acts}{ActsTrack\xspace}
\newcommand{\taumu}{Tau3Mu\xspace}
\newcommand{\pdbbind}{PLBind\xspace}
\newcommand{\synbind}{SynMol\xspace}
\newcommand{\pge}{BernMask-P\xspace}
\newcommand{\gexp}{BernMask\xspace}
\newcommand{\gradpos}{GradGeo\xspace}
\newcommand{\gradcam}{GradGAM\xspace}
\newcommand{\pointmask}{PointMask\xspace}

\newcommand{\ours}[0]{\texttt{GMT}\xspace}
\newcommand{\oursl}[0]{\texttt{GMT-lin}\xspace}
\newcommand{\ourslt}[0]{\text{GMT-lin}\xspace}
\newcommand{\ourss}[0]{\texttt{GMT-sam}\xspace}
\newcommand{\oursst}[0]{\text{GMT-sam}\xspace}
\newcommand{\ourst}[0]{\text{GMT}\xspace}	%
\newcommand{\oursfull}[0]{\textbf{G}raph \textbf{M}ultilinear ne\textbf{T}\xspace}

\newcommand{\smtfull}[0]{subgraph multilinear extension\xspace}
\newcommand{\smt}[0]{\texttt{SubMT}\xspace}

\newcommand{\xgnns}[0]{\text{XGNNs}\xspace}
\newcommand{\xgnn}[0]{\text{XGNN}\xspace}

\icmltitlerunning{How Interpretable Are Interpretable Graph Neural Networks?}

\begin{document}

\twocolumn[
  \icmltitle{How Interpretable Are Interpretable Graph Neural Networks? 
  }

  \icmlsetsymbol{intern}{*}
  \icmlsetsymbol{tencent}{1}
  \icmlsetsymbol{cuhk}{2}
  \icmlsetsymbol{hkbu}{3}
  
  \begin{icmlauthorlist}
    \icmlauthor{Yongqiang Chen}{intern,cuhk}
    \icmlauthor{Yatao Bian}{tencent}
    \icmlauthor{Bo Han}{hkbu}
    \icmlauthor{James Cheng}{cuhk}
  \end{icmlauthorlist}

  \icmlcorrespondingauthor{Yatao Bian}{yatao.bian@gmail.com}

  \icmlkeywords{Interpretation, Graph Neural Networks, Out-of-Distribution Generalization, Multilinear Extension, Causality, Geometric Deep Learning}

  \vskip 0.3in
]

\printAffiliationsAndNotice{\textsuperscript{*}Work done during an internship at Tencent AI Lab. \textsuperscript{1}Tencent AI Lab. \textsuperscript{2}The Chinese University of Hong Kong. \textsuperscript{3}Hong Kong Bapist University} %

\begin{abstract}
  Interpretable graph neural networks (\xgnns) are widely adopted in various scientific applications involving graph-structured data. Existing XGNNs predominantly adopt the attention-based mechanism to learn edge or node importance for extracting and making predictions with the interpretable subgraph.
  However, the representational properties and limitations of these methods remain inadequately explored.
  In this work, we present a theoretical framework that formulates interpretable subgraph learning with the multilinear extension of the subgraph distribution, 
  coined as \textit{\smtfull} (\smt).
  Extracting the desired interpretable subgraph requires an accurate approximation of \smt, yet we find that the existing \xgnns can have a huge gap in fitting \smt.
  Consequently, the \smt approximation failure will lead to the degenerated interpretability of the extracted subgraphs.
  To mitigate the issue, we design a new \xgnn architecture called \oursfull (\ours), which is provably more powerful in approximating \smt.
  We empirically validate our theoretical findings on a number of graph classification benchmarks. The results demonstrate that \ours outperforms the state-of-the-art up to $10\%$ in terms of both interpretability and generalizability across $12$ regular and geometric graph benchmarks.
\end{abstract}

\section{Introduction}

Graph Neural Networks (GNNs) have been widely used in scientific applications~\citep{ai4sci0,ai4sci} such as Physics~\citep{ai4sci_phy}, Chemistry~\citep{ai4sci_qchem,ai4sci_bchem}, Quantum mechanics~\citep{ai4sci_qua}, Materials~\citep{ai4sci_mat} and Cosmology~\citep{ai4sci_cos}.
In pursuit of scientific discoveries, it often requires GNNs to be able to generalize to \textit{unseen or Out-of-Distribution} (OOD) graphs~\citep{good_bench,drugood,ai4sci}, and also provide \textit{interpretations} of the predictions that are crucial for scientists to collect insights~\citep{ai4sci_xgnn0,ai4sci_xgnn1,ai4sci_xgnn2} and promote better scientific practice~\citep{xgnn_ai4sci0,xgnn_ai4sci1}.
Recently there has been a surge of interest in developing intrinsically interpretable and generalizable GNNs (\xgnns)~\citep{gib,gsat,dir,ciga,lri}.
In contrast to \textit{post-hoc} explanations~\citep{gnn_explainer,xgnn,pgm_explainer,pge,subgraphxgn,gen_xgnn,orphicx} which are shown to be suboptimal in interpretation and sensitive to pre-trained GNNs performance~\citep{gsat,lri}, \xgnns can provide both reliable explanations and (OOD) generalizable predictions under the proper guidance such as information bottleneck~\citep{gib} and causality~\citep{ciga}.

Indeed, the faithful interpretation and the reliable generalization are the \textit{two sides of the same coin} for \xgnns.
Grounded in the causal assumptions of data generation processes, \xgnns assume that there exists a causal subgraph which holds a causal relation with the target label. Predictions made solely based on the causal subgraph are generalizable under various graph distribution shifts~\citep{eerm,gsat,ciga}.
Therefore, \xgnns typically adopt a two-step paradigm that first extracts a subgraph of the input graph and then predicts the label.
To circumvent the inherent discreteness of subgraphs, \xgnns often learn the sampling probability for each edge or node with the attention mechanism and extract the subgraph with high attention scores~\citep{gsat}.
Predictions are then made via a weighted message passing scheme with the attention scores.
Despite the success of the paradigm in enhancing both interpretability and out-of-distribution (OOD) generalization
~\citep{gsat,lri,ciga},
there is limited theoretical understanding of the representational properties and limitations of \xgnns, and whether they can provide faithful interpretations.

\begin{figure*}[t]
  \centering
  \includegraphics[width=0.9\textwidth]{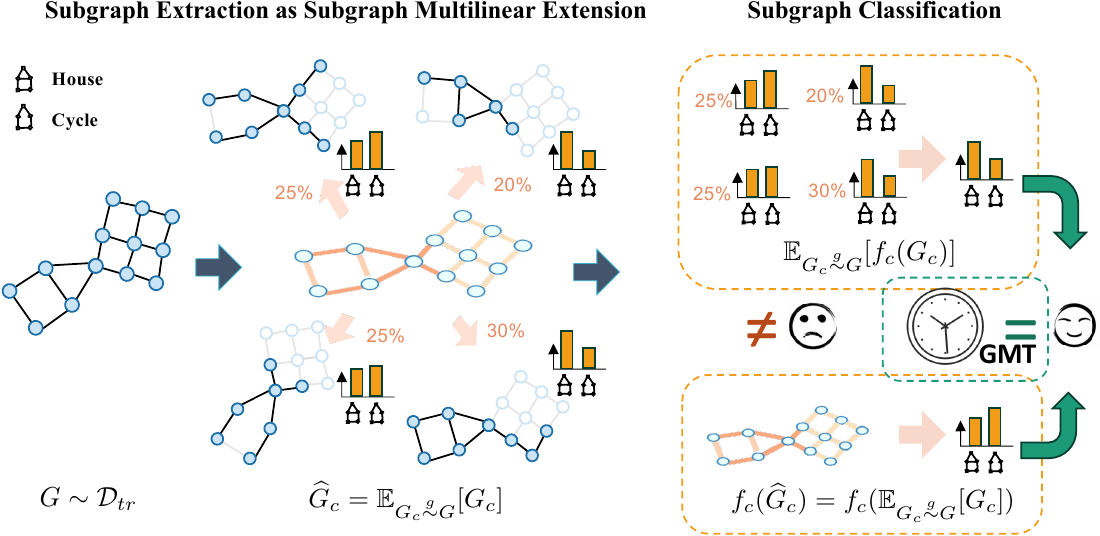}
  \vspace{-0.15in}
  \caption{Illustration of Subgraph Multilinear Extension approximation failure.
    In a binary graph classification task, \xgnns need to classify whether a graph contains a specific “house” or “cycle” motif into two steps: (a) \textbf{Subgraph Extraction}, a subgraph extractor  computes the sampling probability for each edge using the attention mechanism, which further determines the subgraph distribution: part of the “house” or “grid” motif is sampled with a certain probability. The expected subgraph $\widehat{G}_c$ to be sampled according to the subgraph distribution is a “soft” subgraph, with each edge weighting the corresponding sampling probability.
    (b)  \textbf{Subgraph Classification}, each subgraph $G_c^i$ corresponds to a respective label distribution $P(Y|G_c^i)$ (e.g., when some key parts of the “house” motif are sampled, the “house” label probability will be higher). The final label predictions are conditioned on the subgraph distributions, averaged upon each $P(Y|G_c^i)$ with the probability of $G_c^i$ being sampled. The averaged label distribution leads to a prediction of “house” for the example.
    Instead of averaging the subgraph conditional label predictions, previous methods directly take the expected “soft” subgraph as the input to the subgraph classifier GNN $f_c$, which can be biased and lead to an incorrect prediction of “cycle”.
  }
  \label{fig:gmt_illustration}
  \vspace{-0.1in}
\end{figure*}

Inspired by the close connection between interpretable subgraph learning and multilinear extension~\citep{mt}, we present a framework to analyze the expressiveness and evaluate the faithfulness of \xgnns.
In fact, the subgraph learning in \xgnns naturally resembles the multilinear extension of the subgraph predictivity, which we term as \textit{\smtfull} (\smt).
The extracted interpretable subgraph is faithful if the associated prediction is highly correlated with the sampling probability of the subgraph.
However, as demonstrated in Fig.~\ref{fig:gmt_illustration}, we show that the prevalent attention-based paradigm could fail to reliably approximate \smt (Sec.~\ref{sec:expressivity_issue}).
Consequently, the \smt approximation failure will decrease the interpretability of the subgraph for predicting the target label.
More specifically, we instantiate the issue via a causal framework and propose a new interpretability measure called \textit{counterfactual fidelity}, i.e., the sensitivity of the prediction to small perturbations to the extracted subgraphs (Sec.~\ref{sec:counterfactual_faith}).
Although faithful interpretation should have a high counterfactual fidelity with the prediction, we find that \xgnns implemented with the prevalent paradigm only have a low counterfactual fidelity.

To bridge the gap, we propose a simple yet effective \xgnn architecture called \oursfull (\ours). Motivated by the \smt formulation, \ours first performs random subgraph sampling onto the subgraph distribution to approximate \smt, which is provably more powerful in approximating \smt (Sec.~\ref{sec:gmt_sol}). Then, we will train a new classifier onto the trained subgraph extractor without random subgraph sampling, to obtain the final approximator of neural \smt.
Our contributions can be summarized as follows:
\begin{itemize}[leftmargin=*]
  \item We propose the first theoretical framework through the notion of \smt for the expressivity of XGNNs (Sec.~\ref{sec:expressivity});
  \item We propose a new XGNN architecture \ours that is provably more powerful than previous XGNNs. The key differentiator of \ours is a new paradigm to effectively approximate \smt with random subgraph sampling (Sec.~\ref{sec:gmt_sol}).
  \item We validate both our theory and the solution through extensive experiments with $12$ regular and geometric graph benchmarks. The results show that \ours significantly improves the state-of-the-art up to $10\%$ in both interpretability and generalizability (Sec.~\ref{sec:exp}). Our code is available at \url{https://github.com/LFhase/GMT}.
\end{itemize}

\input{sections/1_prelim.tex}

\input{sections/2_issue.tex}

\input{sections/2_understand.tex}
\input{sections/3_solution.tex}

\input{sections/4_exp.tex}

\section{Conclusions}
We developed a theoretical framework for the expressivity of \xgnns by formulating the subgraph learning with multilinear extension (\smt).
We found that existing attention-based \xgnns fail to approximate \smt, leading to unfaithful interpretation and poor generalizability.
Thus, we propose a new architecture called \ours which is provably more powerful in approximating \smt. Extensive experiments on both regular and geometric graphs verify the superior interpretability and generalizability of \ours.

\clearpage
\section*{Acknowledgements}
We thank the reviewers for their valuable comments. This work was supported by Research Grants 8601116, 8601594, and 8601625 from the UGC of Hong Kong. BH was supported by the Guangdong Basic and Applied Basic Research Foundation Nos.~2024A1515012399 and 2022A1515011652, NSFC General Program No.~62376235, HKBU Faculty Niche Research Areas No.~RC-FNRA-IG/22-23/SCI/04, Tencent AI Lab Rhino-Bird Gift Fund, and HKBU CSD Departmental Incentive Grant.

\section*{Impact Statement}
Despite the usefulness of \ours shown in our experiments, the computational overhead of \ourss remains an issue when applied to node classification tasks on a large scale such as millions of nodes, or tasks where fine-grained interpretation in terms of the node features are needed. It would be promising to investigate more efficient approximation approaches of \smt and extend \ours to fine-grained interpretation tasks, or under adversarial scenarios~\citep{hao}. Meanwhile, it is worth noting that the optimization challenge of multiple objectives for interpretation and generalization\citep{pair,feat}.

Considering the wide applications of XGNNs to various industrial and scientific applications, it is crucial to formally characterize and analyze the expressivity of XGNNs.
Built upon the connection between XGNNs and multilinear extension,
our work provides both theoretical and practical tools to understand and improve XGNNs for broader applications and social benefits.
Besides, this paper does not raise any ethical concerns.
This study does not involve
any human subjects, practices to data set releases, potentially harmful insights, methodologies and
applications, potential conflicts of interest and sponsorship, discrimination/bias/fairness concerns,
privacy and security issues, legal compliance, and research integrity issues.
\bibliography{references/graphood,references/ood,references/xgnn,references/references}
\bibliographystyle{icml2024}

\input{sections/9_appdx.tex}
\input{sections/10_appdx_exp.tex}

\end{document}

%% file: sections/0_preamble.tex
\usepackage{microtype}
\usepackage{graphicx}
\usepackage{subfigure}
\usepackage{booktabs} %
\usepackage{arydshln} %
\usepackage{tikz}
\usetikzlibrary{bayesnet} %
\usetikzlibrary{arrows}
\usepackage{amsmath}
\usepackage{amssymb}
\usepackage{mathtools}
\usepackage{amsthm}

\usepackage{enumitem,multirow,adjustbox}

\usepackage{hyperref}
\usepackage{url}
\usepackage{xcolor}		%
\definecolor{darkblue}{rgb}{0, 0, 0.5}
\definecolor{beaublue}{rgb}{0.74, 0.83, 0.9}
\definecolor{gainsboro}{rgb}{0.86, 0.86, 0.86}
\definecolor{kleinblue}{rgb}{0,0.18,0.65}
\hypersetup{colorlinks=true,citecolor=kleinblue, linkcolor=kleinblue, urlcolor=kleinblue}

\usepackage{pifont}
\newtheorem{theorem}{Theorem}[section]
\newtheorem{proposition}[theorem]{Proposition}

\newtheorem{definition}[theorem]{Definition}

\input{math_commands.tex}

\usepackage{enumitem,algorithm,algorithmic}

\usepackage{xspace}

\newcommand{\dataset}{{\cal D}}

\newcommand{\envall}{{\gE_{\text{all}}}}

\newcommand{\gen}{{{\text{gen}}}}

\newcommand{\std}[1]{{$\scriptstyle\pm#1$}}

\makeatletter
\newcommand*\rel@kern[1]{\kern#1\dimexpr\macc@kerna}
\newcommand*\widebar[1]{%
  \begingroup
  \def\mathaccent##1##2{%
    \rel@kern{0.8}%
    \overline{\rel@kern{-0.8}\macc@nucleus\rel@kern{0.2}}%
    \rel@kern{-0.2}%
  }%
  \macc@depth\@ne
  \let\math@bgroup\@empty \let\math@egroup\macc@set@skewchar
  \mathsurround\z@ \frozen@everymath{\mathgroup\macc@group\relax}%
  \macc@set@skewchar\relax
  \let\mathaccentV\macc@nested@a
  \macc@nested@a\relax111{#1}%
  \endgroup
}
\makeatother
\newcommand{\pred}[1]{\widehat{#1}\xspace}

\usepackage[hyperpageref]{backref}

\renewcommand*{\backrefalt}[4]{
  \ifcase #1 \relax
  \or
    (Cited on page #2)
  \else
    (Cited on pages #2)
  \fi
}

\definecolor{Gray}{gray}{0.9}

\graphicspath{{./fig/}}

\newcommand{\gsat}[0]{\texttt{GSAT}\xspace}
\newcommand{\lri}[0]{\texttt{LRI}\xspace}

%% file: math_commands.tex
\usepackage{amsmath,amsfonts,bm}

\newcommand{\ind}{\perp\!\!\!\!\perp}

\def\eqref#1{equation~\ref{#1}}

\def\1{\bm{1}}
\newcommand{\train}{\mathcal{D_{\mathrm{tr}}}}

\def\mT{{\bm{T}}}

\def\mW{{\bm{W}}}

\DeclareMathAlphabet{\mathsfit}{\encodingdefault}{\sfdefault}{m}{sl}
\SetMathAlphabet{\mathsfit}{bold}{\encodingdefault}{\sfdefault}{bx}{n}

\def\gE{{\mathcal{E}}}

\def\gG{{\mathcal{G}}}

\def\gN{{\mathcal{N}}}
\def\gO{{\mathcal{O}}}

\def\gY{{\mathcal{Y}}}

\def\sP{{\mathbb{P}}}

\newcommand{\E}{\mathbb{E}}

\newcommand{\R}{\mathbb{R}}

\newcommand{\sigmoid}{\sigma}

\newcommand{\KL}{D_{\mathrm{KL}}}

%% file: sections/1_prelim.tex
\section{Preliminaries and Related Work}
We begin by introducing preliminary concepts of \xgnns and leave more details to Appendix~\ref{sec:related_appdx}, and also provide a table of notations for key concepts in Appendix~\ref{sec:notations_appdx}.

\textbf{Interpretable GNNs.}
Let $G=(A,X)$ be a graph with node set $V=\{v_1,v_2,...,v_n\}$ and edge set $E=\{e_1,e_2,...,e_m\}$,
where  $A \in \{0,1\}^{n\times n}$  is the adjacency matrix and $X\in \R^{n \times d}$ is the node feature matrix.
In this work, we focus on interpretable GNNs (or \xgnns) for the graph classification task, while the results can be generalized to node-level tasks as well~\citep{gib_node}.
Given each sample from training data $\train=(G^i,Y^i)$,
an interpretable GNN $f:=f_c\circ g$ aims to identify a (causal) subgraph $G_c\subseteq G$ via a subgraph extractor GNN $g:\gG\rightarrow\gG_c$, and then predicts the label via a subgraph classifier GNN $f_c:\gG_c\rightarrow\gY$, where $\gG,\gG_c,\gY$ are the spaces of graphs, subgraphs, and the labels, respectively~\citep{gib}.
Although \textit{post-hoc} explanation approaches also aim to find an interpretable subgraph as the explanation for the model prediction~\citep{gnn_explainer,xgnn,pgm_explainer,pge,subgraphxgn,gen_xgnn,orphicx}, they are shown to be suboptimal in interpretation performance and sensitive to the performance of the pre-trained GNNs~\citep{gsat}.
Therefore, this work focuses on \textit{intrinsic interpretable} GNNs (XGNNs).

A predominant approach to implement \xgnns is to incorporate the idea of information bottleneck~\citep{ib}, such that $G_c$ keeps the minimal sufficient information of $G$ about $Y$~\citep{gib,vgib,gsat,lri,gib_hiera},
which can be formulated as
\begin{equation}
    \text{$\max$}_{G_c}I(G_c;Y)-\lambda I(G_c;G),\ G_c\sim g(G),
\end{equation}
where the maximizing $I(G_c;Y)$ endows the interpretability of $G_c$ while minimizing $I(G_c;G)$ ensures $G_c$ captures only the most necessary information, $\lambda$ is a hyperparamter trade off between the two objectives.
In addition to minimizing $I(G_c;G)$, there are also alternative approaches that impose different constraints such as causal invariance~\citep{ciga,gil} or disentanglement~\citep{dir,cal,grea,disc} to identify the desired subgraphs.
When extracting the subgraph, \xgnns adopts the attention mechanism to learn the sampling probability of each edge or node, which avoids the complicated Monte Carlo tree search used in other alternative implementations~\citep{protGNN}.
Specifically, given node representation learned by message passing $H_i\in\R^h$ for each node $i$, \xgnns either learns a \textbf{node attention} $\alpha_i\in\R_+=\sigma(a(H_i))$ via the attention function $a:\R^h\rightarrow\R_+$, or the \textbf{edge attention} $\alpha_e\in\R_+=\sigma(a([H_u,H_v]))$ for each edge $e=(u,v)$ via the attention function $a:\R^{2h}\rightarrow\R_+$, where $\sigma(\cdot)$ is a sigmoid function. $\boldsymbol{\alpha}=[\alpha_1,...,\alpha_m]^T$ essentially elicits a subgraph distribution of the interpretable subgraph. In this work, we focus on edge-centric subgraph sampling as it is most widely used in \xgnns while our method can be easily generalized to node-centric approaches.

\textbf{Faithful interpretation and (OOD) generalization.}
The faithfulness of interpretation is critical to all interpretable and explainable methods~\citep{fidelity,mythos_inter,robust_xnn,att_not_exp}. There are several metrics developed to measure the faithfulness of graph explanations, such as fidelity~\citep{xgnn_tax,GraphFramEx}, counterfactual robustness~\citep{RCExplainer,counterfactual_xgnn_sur,clear}, and equivalence~\citep{xgnn_equi}, which are however limited to post-hoc graph explanation methods. In contrast, we develop the first faithfulness measure for \xgnns in terms of counterfactual invariance.

In fact, the generalization ability and the faithfulness of the interpretation are naturally intertwined in \xgnns. \xgnns need to extract the underlying ground-truth subgraph in order to make correct predictions on unseen graphs~\citep{gsat}. When distribution shifts are present during testing, the underlying subgraph that has a causal relationship with the target label (or causal subgraphs) naturally becomes the ground-truth subgraph that needs to be learned by \xgnns~\citep{ciga}.

\textbf{Multilinear extension}
serves as a powerful tool
for maximizing combinatorial functions, especially for submodular set function maximization \citep{mt,Vondrak08,optimal_drsub,sets2multisets,neural_set}.
It is the expected value of a set function under the fully factorized Bernoulli distribution, which is also one typical objective for continuous submodular function maximization \citep{2020csfm}.
Our work is the first to identify subgraph multilinear extension as the factorized subgraph distribution for interpretable subgraph learning.

%% file: sections/2_issue.tex
\section{On the Expressivity of Interpretable GNNs}
\label{sec:expressivity}
In this section, we present our theoretical framework for characterizing the expressivity of \xgnns. Since all of the existing approaches need to maximize $I(G_c;Y)$ regardless of the regularization on $G_c$, we focus on the modeling of the subgraph distribution that maximizes $I(G_c;Y)$.

\subsection{Subgraph multilinear extension}
The need for maximizing $I(G_c;Y)$ originates from extracting information in $G$ to predict $Y$ with $f_c$, %
\begin{equation}\label{eq:GI}
	\begin{aligned}
		\text{$\max$}_{f_c} I(G;Y) &= H(Y)-H(Y|G)
	\end{aligned}
\end{equation}
which amounts to $\text{$\min$}_{f_c} H(Y|G)$
due to the irrelevance of $H(Y)$ and $f_c$.
For each sample $(G,Y)$, \xgnn then adopts the subgraph extractor $g$ to extract a subgraph $G_c\sim g(G)$, and take $G_c$ as the input of $f_c$ to predict $Y$.
Then, Eq.~\ref{eq:GI} is realized as follows\footnote{With a bit of abuse of notations, we will omit the unnecessary superscript of samples for the sake of clarity.}: let $L(\cdot)$ be the cross-entropy loss, then
\begin{align}\label{eq:GCE}
    \text{$\min$}_{g,f_c} &  \E_{G, Y} \left[ - \log \E_{G_c\stackrel{g}{\sim}G} P_{f_c}(Y|G_c) \right] \\
   &  =\E_{(G,Y)} [L( \E_{G_c\stackrel{g}{\sim}G} [f_c(G_c)], Y)].
\end{align}
We leave more details about the deduction of Eq.~\ref{eq:GCE} in Appendix~\ref{sec:GCE_deduce_appdx}. 
Let $\boldsymbol{\alpha}\in\R^m_+$ be the attention score elicited from the subgraph extractor $g$, which defines the subgraph distribution. 
Note that $f_c$ is a GNN defined only for \textit{discrete} graph-structured inputs.
Considering $f_c(G_c)$ is a \textit{set function} with respect to node/edge index subsets of $G$ (i.e., subgraphs $G_c$),
and the parameterization of $P(G_c | G)$ in \xgnns~\citep{gsat},
we resort to the \textit{multilinear extension} of $f_c(G_c)$~\citep{mt}.

\begin{definition}[Subgraph multilinear extension (\smt)]\label{def:sub_mt}
	Given the attention score $\boldsymbol{\alpha}\in [0, 1]^m$ as sampling probability of $G_c$, \xgnns factorize $P(G_c | G)$ as independent Bernoulli distributions on edges:
	\[P(G_c|G)=\prod_{e\in G_c}\alpha_e\prod_{e\in G/ G_c}(1-\alpha_e),\]
	which elicits the \textit{multilinear extension} of $f_c(G_c)$ in Eq.~\ref{eq:GCE}:
	\begin{equation}\label{eq:smt}
		\begin{aligned}
			F_c(\boldsymbol{\alpha}; G) &:=\sum_{G_c\in G}f_c(G_c)\prod_{e\in G_c}\alpha_e\prod_{e\in G/G_c}(1-\alpha_e) \\&=\E_{G_c\stackrel{g}{\sim}G}f_c(G_c).
		\end{aligned}
	\end{equation}
\end{definition}

The parameterization of $P(G)$ is widely employed in \xgnns~\citep{gsat,ciga}, which implicitly assumes the random graph data model~\citep{er_graph}. Def.~\ref{def:sub_mt} can also be generalized to other graph models with the corresponding parameterization of $P(G)$~\citep{sbm,graphon}.
When a \xgnn approximates \smt well, we have:
\begin{definition}[$\epsilon$-\smt approximation]\label{def:submt_approx}
	Let $d(\cdot,\cdot)$ be a distribution distance metric, a \xgnn $f=f_c\circ g$ $\epsilon$-approximates \smt (Def.~\ref{def:sub_mt}), if there exists $\epsilon\in\R_+$ such that $d(P_f(Y|G),P(Y|G))\leq\epsilon$
	where $P(Y|G)\in\R^{|\gY|}$ is the ground truth conditional label distribution, and $P_f(Y|G)\in\R^{|\gY|}$ is the predicted label distribution for $G$ via a \xgnn $f$, i.e., $P_f(Y|G)=\E_{G_c\stackrel{g}{\sim}G}f_c(G_c)$.
\end{definition}
Def.~\ref{def:submt_approx} is a natural requirement for \xgnn that approximates \smt properly.
With the definition of \smt, we can write the objective in Eq.~\ref{eq:GCE} as the following:
\begin{equation}\label{eq:GCE_mt}
	\begin{aligned}
		& \ \E_{(G,Y)\sim\train} [L(\E_{G_c\stackrel{g}{\sim}G}f_c(G_c),Y)]\\
        &=\E_{(G,Y)\sim\train}L(F_c(\boldsymbol{\alpha}; G), Y),
	\end{aligned}
\end{equation}
from which it suffices to know that optimizing for $g,f_c$ in Eq.~\ref{eq:GCE} requires an accurate estimation of \smt.

\subsection{Issues of existing approaches}
\label{sec:expressivity_issue}
In general, evaluating \smt requires $\gO(2^m)$ calls of $f_c(G_c)$.
Nonetheless, existing \xgnns introduce a soft subgraph $\widehat{G}_c$ with the adjacency matrix as the attention matrix $\widehat{A}$ where $\widehat{A}_{u,v}\!=\!\alpha_e, \forall e\!=\!(u,\!v)\!\in\! E$, to solve Eq.~\ref{eq:GCE} via weighted message passing~\citep{gsat}:\footnote{With a little abuse of notation, we denote $f_c$ as a generalized GNN that performs weighted message passing if the input graph is a weighted graph.}
\begin{equation}\label{eq:GCE_att}
	\begin{aligned}
		&\E_{(G,Y)\sim\train} [L(\E_{G_c\stackrel{g}{\sim}G}f_c(G_c),Y)]\\
  &= \E_{(G,Y)\sim\train}[L(f_c(\widehat{G}_c),Y)],
	\end{aligned}
\end{equation}
From the edge-centric perspective, the introduction of $\widehat{G}_c$ seems to be natural at first glance, as:
\begin{equation}\label{eq:GCE_att_exp}
	\widehat{G}_c=\E_{G_c\stackrel{g}{\sim}G}G_c.%
\end{equation}
However, Eq.~\ref{eq:GCE_att} holds only when $f_c$ is \textit{linear}. In other words, if Eq.~\ref{eq:GCE_att} holds, we need the following to hold:
\begin{equation}\label{eq:exp_issue}
	f_c(\widehat{G}_c)=f_c(\E[G_c])=\E_{G_c}[f_c(G_c)],
\end{equation}
where the last equality adheres to the equality of Eq.~\ref{eq:GCE_att}.
Obviously $f_c(\cdot)$ is a non-linear function even with a linearized GNN~\citep{sgnn} with more than $1$ layers: 
\begin{equation}\label{eq:linear_gnn}
	f_c(\widehat{G}_c)=\rho(\widehat{A}^kX \mW),
\end{equation}
where $\rho$ is the pooling, $k$ is the number of layers and $\mW\in\R^{h\times h}$ are the learnable weights.
We prove the \smt approximation failure in Appendix~\ref{proof:submt_gap}.
\begin{proposition}\label{thm:submt_gap}
	An XGNN based on linear GNN with $k>1$ cannot satisfy
	Eq.~\ref{eq:exp_issue}, thus cannot  approximate \smt.
\end{proposition}
When given more complicated GNNs, the approximation error to \smt can be even higher, as verified in Appendix~\ref{sec:smt_gap_viz_appdx}.
For example, when $k=2$ and $|\gY|=1$, Eq.~\ref{eq:linear_gnn} is convex, and we have $f_c(\E[A])\leq\E[f_c(A)]$ due to Jensen's inequality when fitting \smt.

%% file: sections/2_understand.tex
\begin{figure*}[t]
	\vspace{-0.1in}
	\centering
	\subfigure[SCM of \xgnns.]{\label{fig:scm_ber}
		\raisebox{0.4\height}{\includegraphics[width=0.35\textwidth]{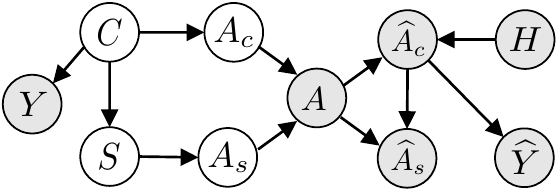}}
	}
	\subfigure[Counterfactual fidelity on BA-2Motifs.]{\label{fig:counterfactual_fidelty_ba}
		\includegraphics[width=0.3\textwidth]{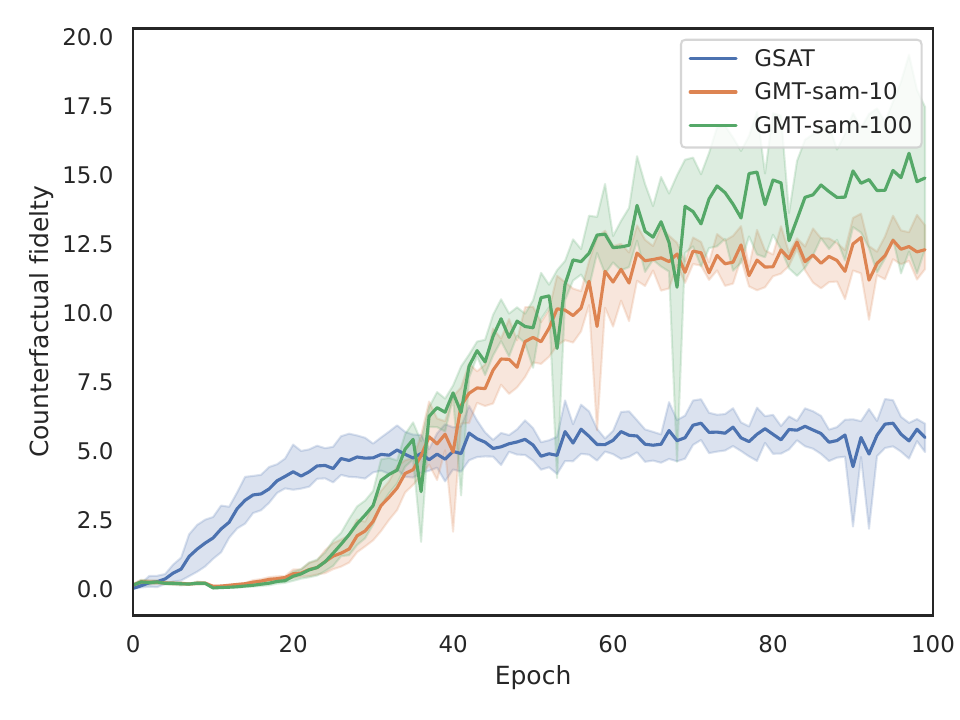}
    	}
	\subfigure[Counterfactual fidelity on Mutag.]{\label{fig:counterfactual_fidelty_mutag}
		\includegraphics[width=0.3\textwidth]{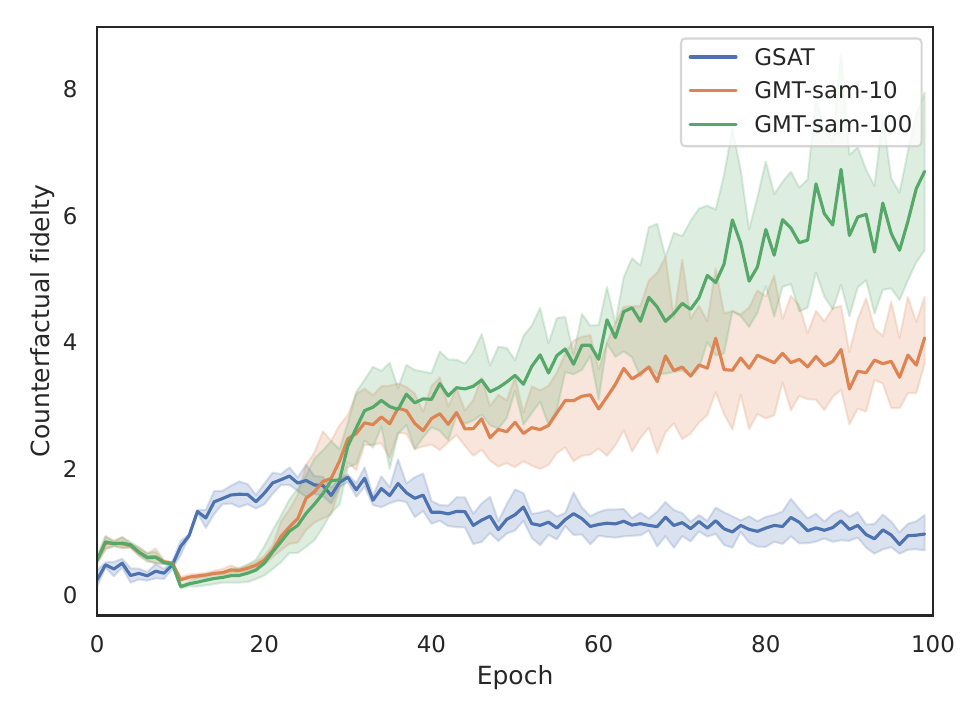}
	}
	\vspace{-0.15in}
	\caption{Comparison of simulated \smt and \gsat in terms of counterfactual faithfulness. }
	\vspace{-0.1in}
\end{figure*}
\section{On the Generalization and Interpretability: A Causal View}
\label{sec:causal_view}
To understand the consequences of the \smt approximation issue,
we conduct a causal analysis of the interpretation faithfulness in \xgnns.
Without loss of generality, we will focus on the edge-centric data generation and interpretation.
\subsection{Causal model of interpretable GNNs}
\label{sec:scm_xgnn}

\textbf{Data generation.} We consider the same data model as previous works~\citep{size_gen2,gsat,ciga},
where the underlying causal subgraph $G_c$ and the spurious subgraph $G_s$ will be assembled via some underlying assembling process. As we focus on the edge-centric view, our following discussion will focus on the graph structures $A_c$ and $A_s$ of the subgraphs. Full details of the structural causal model are deferred to  Appendix~\ref{sec:scm_xgnn_appdx}.

As shown in Fig.~\ref{fig:scm_ber}, there are latent causal and spurious variables $C$ and $S$ that have the invariant and spurious correlations with the label $Y$ across training and test distributions, respectively.
$C$ and $S$ correspondingly control the generation of causal subgraph $G_c$, and the spurious subgraph $G_s$. For example, when generating $A_c$ and $A_s$, $C$ and $S$ will specify the number of nodes in  $A_c$ and $A_s$ and also the edge sampling probability for edges in $A_c$ and $A_s$.

\textbf{Interpretation.} Correspondingly, \xgnns use a subgraph extractor to predict the causal and spurious subgraphs $\widehat{G}_c$ and $\widehat{G}_s$, respectively.
The extraction aims to reverse the generation and recover the structure of the underlying causal subgraph $A_c$.  We denote the \xgnn architecture and the hyperparameter settings as $H$.
$H$ takes $A$ as inputs to learn the edge sampling probability via the attention mechanism and then obtain $\widehat{A}_c$.
Once $\widehat{A}_c$ is determined, $\widehat{A}_s\!=\!A\!-\!\widehat{A}_c$ is also obtained by taking the complementary part. Then, the extracted causal and spurious subgraphs are obtained with $\widehat{G}_c\!=\!(X,\widehat{A}_c)$ and $\widehat{G}_s\!=\!(X,\widehat{A}_s)$, respectively. The classifier then uses $\widehat{G}_c$ to make the prediction $\widehat{Y}$.

\subsection{Causal faithfulness of \xgnns}
\label{sec:counterfactual_faith}

With the aforementioned causal model, we are able to specify the causal desiderata for faithful \xgnns.
When a \xgnn fails to accurately approximate \smt, the estimated label conditional probability will have a huge gap from the ground truth.
The failure will bias the optimization of the subgraph extractor $g$ and lead to the degenerated interpretability of $\widehat{A}$.
More concretely, the recovery of $\widehat{A}$ to the underlying $A$ will be worse, which further affects the extraction of $G_c$ and brings both worse interpretation and (OOD) generalization performance.
As a single measure such as the interpretation or generalization may not fully reflect the consequence or even exhibit conflicted information\footnote{For example, in the experiments of~\citet{gsat}, higher interpretation performance does not necessarily correlate with higher generalization performance.}, we consider a direct notion that jointly consider the interpretability and generalizabiliy to measure the causal faithfulness of \xgnns, inspired by~\citet{att_not_exp}.
\vspace{-0.1in}
\begin{definition}[$(\delta,\epsilon)$-counterfactual fidelity]\label{def:counterfactual_x}
	Given a meaningful minimal distance $\delta>0$,
	let $d(\cdot,\cdot)$ be a distribution distance metric ,
	if a \xgnn $f=f_c\circ g$ commits to the $\epsilon-$counterfactual fidelity, then there exist $\epsilon>0$ such that, $\forall G, \widetilde{G}$ that $d(P(Y|G),P(Y|\widetilde{G}))\geq \delta$, the following holds:
	\[d(P_f(Y|\widetilde{G}),P_f(Y|G))\geq \epsilon\delta.\]
\end{definition}
\vspace{-0.05in}
Intuitively, if the extracted interpretable subgraph $\widehat{G}_c$ is faithful to the target label, then the predictions made based on $\widehat{G}_c$ are sensitive to any perturbations on $\widehat{G}_c$.
Different from counterfactual interpretability~\citep{counterfactual_xgnn_sur,counterfactual_gl} that seeks minimum modifications to change the predictions,
$(\delta,\epsilon)$-counterfactual fidelity measures how sensitive are the predictions to the changes of the interpretable subgraphs. A higher fidelity implies better interpretability and is also a natural behavior of a \xgnn that approximates \smt well.
\begin{proposition}\label{thm:smt_fidelty}
	If a \xgnn $f$ $\epsilon$-approximates \smt,  $f$ satisfies $(\delta,1\!-\!\frac{2\epsilon}{\delta})$-counterfactual fidelity.
\end{proposition}
The proof is given in Appendix~\ref{proof:smt_fidelty}. Intuitively, Proposition~\ref{thm:smt_fidelty} implies that the counterfactual fidelity is an effective measure for the approximation ability of \smt.

\textbf{Practical estimation of counterfactual fidelity.}
Since it is hard to enumerate every possible $\widetilde{G}$,
to verify Def.~\ref{def:counterfactual_x}, we consider a random attention matrix $\widetilde{A}\sim\sigmoid(\gN(\mu_{\widehat{H}_A},\sigma_{\widehat{H}_A}))$,
where $\mu_{\widehat{H}_A}$ and $\sigma_{\widehat{H}_A}$ are the mean and standard deviation
of the pre-attention matrix $\widehat{H}_A$ (The adjacency matrix with the unnormalized attention).
Each non-symmetric entry in $\widetilde{A}$ is sampled independently following the factorization of $P(G)$.
We randomly sample $\widetilde{A}$ by $k$ times and obtain
\vspace{-0.05in}
\begin{equation}
	c_{\widehat{G}_c} = \frac{1}{k}\sum_{i=1}^k d(f_c(Y|\widetilde{G}^i_c),f_c(Y|\widehat{G}_c)),
	\vspace{-0.05in}
\end{equation}
where $\widetilde{G}^i_c=(X,\widetilde{A}^i_c)$ and $d$ is total variation distance.
We compute $c_{\widehat{G}_c}$ for the state-of-the-art \xgnn \gsat~\citep{gsat}.
Shown as in Fig.~\ref{fig:counterfactual_fidelty_ba},~\ref{fig:counterfactual_fidelty_mutag},
we plot the counterfactual fidelity of \gsat on BA-2Motifs and Mutag datasets against
is $2$ to $3$ times lower than the simulated \smt with $10$ and $100$ sampling rounds.
We provide a more detailed discussion in Appendix~\ref{sec:pratical_cf_appdx} and Appendix~\ref{sec:cf_viz_appdx}.

%% file: sections/3_solution.tex
\input{tables/interpretation.tex}

\input{tables/generalization.tex}

\section{Building Reliable \xgnns}
\label{sec:gmt_sol}
The aforementioned gap motivates us to propose a new \xgnn architecture, called \oursfull (\ours), to provide both faithful interpretability and reliable (OOD) generalizability.
\ours have two variants, i.e., \oursl and \ourss, motivated by resolving the failures in Sec.~\ref{sec:expressivity_issue}.

\subsection{Linearized \ourst}
\label{sec:gmt_lin}
Recall that the main reason for the failure of Eq.~\ref{eq:exp_issue} is because of the non-linearity of the expectation to the $k$ weighted message passing with $k>1$. If $k$ can be reduced to $1$, then the linearity can be preserved to ensure a better approximation of \smt, which naturally motivates the following variant:
\begin{equation}\label{eq:gmt_linear}
  (\text{\oursl})\qquad
  f^l(\widehat{G}_c)=\rho(\widehat{A}\odot A^{k-1}X \mW),
\end{equation}
Compared to the previous weighted message passing scheme with linearized GNN (Eq.~\ref{eq:linear_gnn}), \oursl improves the linearity by reducing the number of weighted message passing rounds to $1$. 
If $\exists \mT\in\R^{|\gY|\times |\gY|}$ such that $\mT\cdot f_c(G_c)=P(Y|G_c)$ ($f_c$ is linear), then,we can incorporate \oursl into Eq.~\ref{eq:exp_issue} and have
\[f^l(\widehat{G}_c)=\mT\cdot f(\widehat{G}_c)=\E [f_c(G_c)],\]
due to the linearity of $f^l(G_c)$ with respect to $G_c$.
During training, $\mT$ can be further absorbed into $\mW$, which implies that \oursl is able to fit to \smt.
Empirically, we find that the simple strategy of \oursl already yields better interpretability than the state-of-the-art methods even with non-linear GNNs in experiments.

\subsection{\ourst with random subgraph sampling}
\label{sec:gmt_sam}

To generalize \ours to more general cases, inspired by the \smt formulation, we propose a random subgraph sampling approach, that performs Markov Chain Monte Carlo (MCMC) sampling to approximate \smt. More concretely, given the attention matrix $\widehat{A}$, we perform $t$ rounds of random subgraph sampling from the subgraph distribution elicited by $\widehat{A}$ (or equivalently $\widehat{G}_c=(X,\widehat{A})$ as in \smt(Def.~\ref{def:sub_mt})),
and obtain $t$ i.i.d. random subgraph samples $\{G_c^i\}_{i=1}^t$ for estimating \smt as the following:
\begin{equation}\label{eq:gmt_sampling}
  (\text{\ourss})\qquad
  f_c^s(\widehat{G}_c)=\frac{1}{t}\sum_{i=1}^tf_c(Y|G_c^i),
\end{equation}
where $f_c$ is the classifier taking discrete subgraphs as inputs.

\input{tables/lri.tex}
\input{tables/lri_gen.tex}

\begin{theorem}\label{thm:gmt_success}
  Given the attention matrix $\widehat{A}$,
  and the distribution distance metric $d$ as the total variation distance,
  let $C=|\gY|$,
  for a \ourss with $t$ i.i.d. samples of
  $G_c^i\sim P(G_c|G)$, then,
  there exists $\epsilon\in\R_+$ such that,
  with a probability at least $1-e^{-t\epsilon^2/4}$, \ourss $\frac{\epsilon C}{2}$-approximates \smt and satisfies $(\delta,1-\frac{\epsilon C}{\delta})$ counterfactual fidelity.
\end{theorem}
The proof for Theorem~\ref{thm:gmt_success} is given in Appendix~\ref{proof:gmt_success}.
Intuitively, with more random subgraph samples drawn from $P(G_c|G)$, \ourss obtains a more accurate estimation of \smt.
However, it will incur more practical challenges such as the a) gradient of discrete sampling and b) computational overhead. To overcome the challenges a) and b), we incorporate the following two techniques.

\textbf{Backpropagation of discrete sampling.}
To enable gradient backpropagation with the sampled subgraphs, we also incorporate gradient estimation techniques such as Gumbel softmax and straight-through estimator~\citep{gumbel,gumbel2}.
Compared to the state-of-the-art \xgnn \gsat~\citep{gsat},
this scheme brings two additional benefits: (i) reduces the gradient biases in discrete sampling with Gumbel softmax; (ii) avoids weighted message passing and alleviates the input distribution gap to the graph encoder when shared in both $f_c$ and $g$ as in \gsat.

\textbf{The number of sampling rounds.} Although the estimation of \smt will be more accurate with the increased sampling rounds, it unnecessarily brings improvements. First, as shown in Fig.~\ref{fig:ablation}, the performance may be saturated with moderately sufficient samplings. Besides, the performance may degenerate as more sampling rounds can affect the optimization, as discussed in Appendix~\ref{sec:gmt_impl_appdx}.

\subsection{Learning neural subgraph multilinear extension}
Although \ours trained with \ourss improve interpretability, \ourss still requires multiple random subgraph sampling to approximate \smt and costs much additional overhead.
To this end, we propose to learn a neural \smt that only requires single sampling, based on the trained subgraph extractor $g$ with \ourss.

Learning the neural \smt is essentially to approximate the MCMC with a neural network, though it is inherently challenging to approximate MCMC~\citep{no_free_lunch_MCMC,papamarkou2022a}.
Nevertheless, the feasibility of neural \smt learning is backed by the inherent causal subgraph assumption of~\citep{ciga}, once the causal subgraph is correctly identified, simply learning the statistical correlation between the subgraph and the label is sufficient to recover the causal relation.

Therefore, we propose to simply re-train a new classifier GNN with the frozen subgraph extractor, to distill the knowledge contained in $\widehat{G}_c$ about $Y$.
This scheme also brings additional benefits over the originally trained classifier, which avoid to learn all the available statistical correlations between $G_c$ and $Y$ that can be spurious.
More details and discussions on the implementations are given in Appendix~\ref{sec:gmt_impl_dis_appdx}.

%% file: tables/interpretation.tex
\begin{table*}[t]
    \vspace{-0.1in}
    \caption{Interpretation Performance (AUC) on regular graphs.%
        Results with the mean-1*std larger than the best baselines are shadowed. }
    \small\sc\centering
    \begin{tabular}{llcccccc}
        \toprule
        \multirow{2}{*}{GNN} & \multirow{2}{*}{Method} & \multirow{2}{*}{Ba-2motifs}                        & \multirow{2}{*}{Mutag}                             & \multirow{2}{*}{MNIST-75sp}                        & \multicolumn{3}{c}{Spurious-motif}                                                                                                                           \\
                             &                         &                                                    &                                                    &                                                    & $b=0.5$                                            & $b=0.7$                                            & $b=0.9$                                            \\
        \midrule\multirow{5}{*}{GIN}
                             & GNNExplainer            & $67.35$\std{3.29}                                  & $61.98$\std{5.45}                                  & $59.01$\std{2.04}                                  & $62.62$\std{1.35}                                  & $62.25$\std{3.61}                                  & $58.86$\std{1.93}                                  \\
                             & PGExplainer             & $84.59$\std{9.09}                                  & $60.91$\std{17.10}                                 & $69.34$\std{4.32}                                  & $69.54$\std{5.64}                                  & $72.33$\std{9.18}                                  & $72.34$\std{2.91}                                  \\
                             & GraphMask               & $92.54$\std{8.07}                                  & $62.23$\std{9.01}                                  & $73.10$\std{6.41}                                  & $72.06$\std{5.58}                                  & $73.06$\std{4.91}                                  & $66.68$\std{6.96}                                  \\
                             & IB-Subgraph             & $86.06$\std{28.37}                                 & $91.04$\std{6.59}                                  & $51.20$\std{5.12}                                  & $57.29$\std{14.35}                                 & $62.89$\std{15.59}                                 & $47.29$\std{13.39}                                 \\
                             & DIR                     & $82.78$\std{10.97}                                 & $64.44$\std{28.81}                                 & $32.35$\std{9.39}                                  & $78.15$\std{1.32}                                  & $77.68$\std{1.22}                                  & $49.08$\std{3.66}                                  \\
        \midrule\multirow{3}{*}{GIN}
                             & GSAT                    & $98.85$\std{0.47}                                  & $99.35$\std{0.95}                                  & $80.47$\std{1.86}                                  & $74.49$\std{4.46}                                  & $72.95$\std{6.40}                                  & $65.25$\std{4.42}                                  \\
                             & \ourslt                 & $98.36$\std{0.56}                                  & $99.86$\std{0.09}                                  & $82.98$\std{1.49}                                  & $76.06$\std{6.39}                                  & $76.50$\std{5.63}                                  & \cellcolor{lightskyblue}$\mathbf{80.57}$\std{2.59} \\
                             & \oursst                 & \cellcolor{lightskyblue}$\mathbf{99.62}$\std{0.11} & \cellcolor{lightskyblue}$\mathbf{99.87}$\std{0.11} & \cellcolor{lightskyblue}$\mathbf{86.50}$\std{1.80} & \cellcolor{lightskyblue}$\mathbf{85.50}$\std{2.40} & \cellcolor{lightskyblue}$\mathbf{84.67}$\std{2.38} & $73.49$\std{5.33}                                  \\
        \midrule\multirow{3}{*}{PNA}
                             & GSAT                    & $89.35$\std{5.41}                                  & $99.00$\std{0.37}                                  & $85.72$\std{1.10}                                  & $79.84$\std{3.21}                                  & $79.76$\std{3.66}                                  & $80.70$\std{5.45}                                  \\
                             & \ourslt                 & $95.79$\std{7.30}                                  & $99.58$\std{0.17}                                  & $85.02$\std{1.03}                                  & $80.19$\std{2.22}                                  & $84.74$\std{1.82}                                  & $85.08$\std{3.85}                                  \\
                             & \oursst                 & \cellcolor{lightskyblue}$\mathbf{99.60}$\std{0.48} & \cellcolor{lightskyblue}$\mathbf{99.89}$\std{0.05} & $\mathbf{87.34}$\std{1.79}                         & \cellcolor{lightskyblue}$\mathbf{88.27}$\std{1.71} & \cellcolor{lightskyblue}$\mathbf{86.58}$\std{1.89} & \cellcolor{lightskyblue}$\mathbf{85.26}$\std{1.92} \\
        \bottomrule
        \label{table:Interpretation}
    \end{tabular}%
    \vspace{-0.1in}
\end{table*}

%% file: tables/generalization.tex
\begin{table*}[t]
    \vspace{-0.15in}
    \caption{Prediction Performance (Acc.) on regular graphs.
    The shadowed entries are the results with the mean-1*std larger than the mean of the corresponding best baselines. }
    \small\sc\centering
    \begin{tabular}{llcccccc}
        \toprule
        \multirow{2}{*}{GNN} & \multirow{2}{*}{Method} & \multirow{2}{*}{MolHiv (AUC)}                      & \multirow{2}{*}{Graph-SST2}                        & \multirow{2}{*}{MNIST-75sp}                        & \multicolumn{3}{c}{Spurious-motif}                                                                                                                           \\
                             &                         &                                                    &                                                    &                                                    & $b=0.5$                                            & $b=0.7$                                            & $b=0.9$                                            \\
        \midrule
        \multirow{3}{*}{GIN}
                             & GIN                     & $76.69$\std{1.25}                                  & $82.73$\std{0.77}                                  & $95.74$\std{0.36}                                  & $39.87$\std{1.30}                                  & $39.04$\std{1.62}                                  & $38.57$\std{2.31}                                  \\
                             & IB-subgraph             & $76.43$\std{2.65}                                  & $82.99$\std{0.67}                                  & $93.10$\std{1.32}                                  & $54.36$\std{7.09}                                  & $48.51$\std{5.76}                                  & $46.19$\std{5.63}                                  \\
                             & DIR                     & $76.34$\std{1.01}                                  & $82.32$\std{0.85}                                  & $88.51$\std{2.57}                                  & $45.49$\std{3.81}                                  & $41.13$\std{2.62}                                  & $37.61$\std{2.02}                                  \\
        \midrule\multirow{3}{*}{GIN}
                             & GSAT                    & $76.12$\std{0.91}                                  & $83.14$\std{0.96}                                  & $96.20$\std{1.48}                                  & $47.45$\std{5.87}                                  & $43.57$\std{2.43}                                  & $45.39$\std{5.02}                                  \\
                             & \ourslt                 & $76.87$\std{1.12}                                  & $83.19$\std{1.28}                                  & $96.01$\std{0.25}                                  & $47.69$\std{4.93}                                  & $53.11$\std{4.12}                                  & $46.22$\std{4.18}                                  \\
                             & \oursst                 & \cellcolor{lightskyblue}$\mathbf{77.22}$\std{0.93} & $\mathbf{83.62}$\std{0.50}                         & \cellcolor{lightskyblue}$\mathbf{96.50}$\std{0.19} & \cellcolor{lightskyblue}$\mathbf{60.09}$\std{2.40} & \cellcolor{lightskyblue}$\mathbf{54.34}$\std{4.04} & \cellcolor{lightskyblue}$\mathbf{55.83}$\std{5.68} \\
        \midrule\multirow{4}{*}{PNA}
                             & PNA                     & $78.91$\std{1.04}                                  & $79.87$\std{1.02}                                  & $87.20$\std{5.61}                                  & $68.15$\std{2.39}                                  & $66.35$\std{3.34}                                  & $61.40$\std{3.56}                                  \\
                             & GSAT                    & $79.82$\std{0.67}                                  & $80.90$\std{0.37}                                  & $93.69$\std{0.73}                                  & $68.41$\std{1.76}                                  & $67.78$\std{3.22}                                  & $51.51$\std{2.98}                                  \\
                             & \ourslt                 & $80.05$\std{0.71}                                  & $81.18$\std{0.47}                                  & $94.44$\std{0.49}                                  & $69.33$\std{1.42}                                  & $64.49$\std{3.51}                                  & $58.30$\std{6.61}                                  \\
                             & \oursst                 & $\mathbf{80.58}$\std{0.83}                         & \cellcolor{lightskyblue}$\mathbf{82.36}$\std{0.96} & \cellcolor{lightskyblue}$\mathbf{95.75}$\std{0.42} & \cellcolor{lightskyblue}$\mathbf{71.98}$\std{3.44} & $\mathbf{69.68}$\std{3.99}                         & \cellcolor{lightskyblue}$\mathbf{67.90}$\std{3.60} \\
        \bottomrule
        \label{table:Generalization}
    \end{tabular}%
    \vspace{-0.1in}
\end{table*}

%% file: tables/lri.tex
\begin{table*}[t]
    \vspace{-0.1in}
    \caption{Interpretation performance on geometric graphs.
        Results with the mean-1*std larger than the best baselines are shadowed. }
    \small
    \centering\sc
    \label{tab:lri_inter}
    \resizebox{\textwidth}{!}{%
        \begin{tabular}{lcccccccc}
            \toprule
            \multicolumn{1}{c}{\multirow{2}{*}{}}
                                 & \multicolumn{2}{c}{\textsc{\acts}}                 & \multicolumn{2}{c}{\textsc{\taumu}}                & \multicolumn{2}{c}{\textsc{\synbind}}              & \multicolumn{2}{c}{\textsc{\pdbbind}}                        \\
            \cmidrule(l{5pt}r{5pt}){2-3} \cmidrule(l{5pt}r{5pt}){4-5} \cmidrule(l{5pt}r{5pt}){6-7} \cmidrule(l{5pt}r{5pt}){8-9}
            \multicolumn{1}{c}{} & ROC AUC                                            & Prec@12                                            & ROC AUC
                                 & Prec@12                                            & ROC AUC                                            & Prec@12                                            & ROC AUC                                            & Prec@12 \\
            \midrule
            Random               & 50                                                 & 21                                                 & 50                                                 & 35
                                 & 50                                                 & 31                                                 & 50                                                 & 45                                                           \\
            \gradpos             & $69.31 $\std{0.89}                                 & $33.54$\std{1.23}                                  & $78.04$\std{0.57}                                  & $64.18$\std{1.25 }
                                 & $76.38$\std{4.96}                                  & $64.72$\std{3.75 }                                 & $58.11$\std{2.91 }                                 & $64.78$\std{4.73 }
            \\
            \gexp                & $54.23 $\std{4.31}                                 & $20.46$\std{5.46}                                  & $71.58$\std{0.69}                                  & $60.51$\std{0.76 }
                                 & $76.38$\std{4.96}                                  & $64.72$\std{3.75 }                                 & $52.23$\std{4.45 }                                 & $41.50$\std{9.77 }
            \\
            \pge                 & $22.87 $\std{3.33}                                 & $11.29 $\std{5.46}                                 & $70.72$\std{5.10 }                                 & $55.50$\std{6.26}
                                 & $87.06$\std{7.12 }                                 & $77.11$\std{7.58 }                                 & $51.98 $\std{4.66}                                 & $59.20 $\std{5.48}
            \\
            \pointmask           & $49.20$\std{1.51}                                  & $20.54 $\std{1.71}                                 & $55.93$\std{4.85 }                                 & $39.65$\std{7.14 }
                                 & $66.46$\std{6.86 }                                 & $53.93$\std{1.94 }                                 & $50.00 $\std{0.00}                                 & $45.10 $\std{0.00}
            \\
            \gradcam             & $75.19$\std{1.91}                                  & $75.94$\std{2.16}                                  & $76.18$\std{2.62 }                                 & $62.05$\std{2.16 }
                                 & $60.31$\std{4.95 }                                 & $52.35$\std{11.02 }                                & $48.61$\std{2.34}                                  & $55.10$\std{10.57 }
            \\
            \midrule
            LRI-Bernoulli        & $74.38$\std{4.33}                                  & $81.42$\std{1.52}                                  & $78.23$\std{1.11}                                  & $65.64$\std{2.44}
                                 & $89.22$\std{3.58}                                  & $68.76$\std{7.35}                                  & $54.87$\std{1.89}                                  & $72.12$\std{2.60}
            \\
            \ourslt              & \cellcolor{lightskyblue}$\mathbf{77.45}$\std{1.69} & $\mathbf{81.81}$\std{1.57}                         & \cellcolor{lightskyblue}$\mathbf{79.17}$\std{0.82} & \cellcolor{lightskyblue}$\mathbf{68.94}$\std{1.08}
                                 & \cellcolor{lightskyblue}$\mathbf{96.17}$\std{1.44} & \cellcolor{lightskyblue}$\mathbf{86.33}$\std{6.16} & $59.70$\std{1.10}                                  & $70.62$\std{3.59}
            \\
            \oursst              & $75.61$\std{1.86}                                  & $81.96$\std{1.35}                                  & $78.28$\std{1.34}                                  & $65.69$\std{2.61}
                                 & $93.93$\std{3.59}                                  & $83.20$\std{4.74}                                  & \cellcolor{lightskyblue}$\mathbf{60.03}$\std{1.02} & $\mathbf{72.56}$\std{2.27}
            \\
            \bottomrule
        \end{tabular}%
    }
    \vspace{-0.1in}
\end{table*}

%% file: tables/lri_gen.tex
\begin{table}[t]
    \vspace{-0.1in}
    \caption{Prediction performance (AUC) on geometric graphs.}
    \centering
    \label{tab:lri_gen}
    \small\sc
    \resizebox{\columnwidth}{!}{%
        \begin{tabular}{lcccc}
            \toprule
            \multicolumn{1}{c}{\multirow{1}{*}{}} & \multicolumn{1}{c}{\text{\acts}}                   & \multicolumn{1}{c}{\text{\taumu}}                  & \multicolumn{1}{c}{\text{\synbind}}                 & \multicolumn{1}{c}{\text{\pdbbind}}
            \\
            \midrule
            ERM                                   & $97.40$\std{0.32}                                  & $82.75$\std{0.16}                                  & $99.30$\std{0.20}                                   & $85.31$\std{2.21 }                  \\
            LRI-Bernoulli                         & $94.00 $\std{0.78}                                 & $86.36$\std{0.06}                                  & $99.30 $\std{0.15}                                  & $85.80 $\std{0.70}                  \\
            \ourslt                               & $93.92 $\std{0.98}                                 & $82.60 $\std{0.17}                                 & $99.26 $\std{0.27}                                  & $86.29 $\std{0.80}
            \\
            \oursst                               & \cellcolor{lightskyblue}$\mathbf{98.55}$\std{0.11} & \cellcolor{lightskyblue}$\mathbf{86.42}$\std{0.08} & \cellcolor{lightskyblue}$\mathbf{99.89} $\std{0.03} & $\mathbf{87.19} $\std{1.86}
            \\
            \bottomrule
        \end{tabular}%
    }
    \vspace{-0.1in}
\end{table}

%% file: sections/4_exp.tex
\section{Experimental Evaluations}\label{sec:exp}
We conduct extensive experiments to evaluate \ours with different backbones and on multiple  benchmarks, and compare both the interpretability and (OOD) generalizability against the baselines.
We will briefly introduce the datasets, baselines, and setups, and leave more details in Appendix~\ref{sec:exp_appdx}.

\subsection{Experimental settings}
\label{sec:exp_setting}

\textbf{Datasets.}
We consider both the regular and geometric graph classification benchmarks following the \xgnn  literature~\citep{gsat,lri}.
For regular graphs, we include \textsc{BA-2Motifs}~\citep{pge}, \textsc{Mutag}~\citep{mutag}, \textsc{MNIST-75sp}~\citep{understand_att}, which are widely evaluated by post-hoc explanation approaches~\citep{xgnn_tax}, as well as \textsc{Spurious-Motif}~\citep{dir}, \textsc{Graph-SST2}~\citep{sst25,xgnn_tax} and \textsc{OGBG-Molhiv}~\citep{ogb} where there exist various graph distribution shifts.
For geometric graphs, we consider \textsc{ActsTrack}, \textsc{Tau3Mu}, \textsc{SynMol} and \textsc{PLBind} curated by~\citet{lri}.

\textbf{Baselines.}
For post-hoc methods, we mainly adopt the results from the previous works~\citep{gsat,lri}, including \text{ GNNExplainer}~\citep{gnn_explainer}, \text{PGExplainer}~\citep{pge}, \text{GraphMask}~\citep{graphmask} for regular graph benchmarks, and \text{BernMask}, \text{BernMask-P}, that are modified from \text{GNNExplainer} and \text{PGExplainer}, GradGeo~\citep{gradgeo}, and GradCam~\citep{gradcam} that are extended for geometric data, as well as PointMask~\citep{pointmask} developed specifically for geometric data.
For \xgnns, since we focus on the interpretation performance, we mainly compared with \xgnns that have the state-of-the-art interpretation abilities, i.e., \gsat~\citep{gsat} and \lri~\citep{lri}, which also have excellent OOD generalizability than other \xgnns~\citep{good_bench}.
We also include two representative \xgnns baselines, \text{DIR}~\citep{dir} and \text{IB-subgraph}~\citep{gib} for regular graphs.

\textbf{Training and evaluation.}
We consider three backbones GIN~\citep{gin} and PNA~\citep{pna} for regular graph data, EGNN~\citep{egnn} for geometric data.
All methods adopted the identical graph encoder, and optimization protocol for fair comparisons.
We tune the hyperparameters as recommended by previous works.
More details are given in Appendix~\ref{sec:eval_appdx}.

\subsection{Experimental results and analysis}
\label{sec:results}

\textbf{Interpretation performance.} As shown in Table.~\ref{table:Interpretation}, compared to post-hoc methods (in the first row) and \gsat, both \oursl and \ourss lead to non-trivial improvements for interpretation performance.
Especially, in challenging Spurious-Motif datasets with distribution shifts, \ourss brings improvements than \gsat up to $15\%$ with GIN, and up to $8\%$ with PNA. In challenging realistic dataset MNIST-75sp, \ourss also improves \gsat up to $6\%$.

\textbf{Generalization performance.} Table \ref{table:Generalization} illustrates the prediction accuracy on regular graph datasets.  We again observe consistent improvements by \ours spanning from molecule graphs to image-converted datasets. Despite distribution shifts, \ourss still brings improvements up to $13\%$ with GIN, and up to $16\%$ against \gsat in Spurious-Motif.

\textbf{Results on geometric graphs.}
Tables \ref{tab:lri_inter} and \ref{tab:lri_gen} show the interpretation and generalization performances of various methods. Again, we observe consistent non-trivial improvements of \oursl and \ourss in most cases than \gsat and post-hoc methods.
Interestingly, \oursl brings more improvements than \ourss in terms of interpretation performance despite its simplicity. In terms of generalization performance, \ourss remains the best method.
The results on geometric datasets further demonstrate the strong generality of \ours across different tasks and backbones.

\subsection{Ablation studies}
\begin{figure*}[t]
    \vspace{-0.1in}
    \centering
    \subfigure[Counterfactual fidelity.]{
        \includegraphics[width=0.27\textwidth]{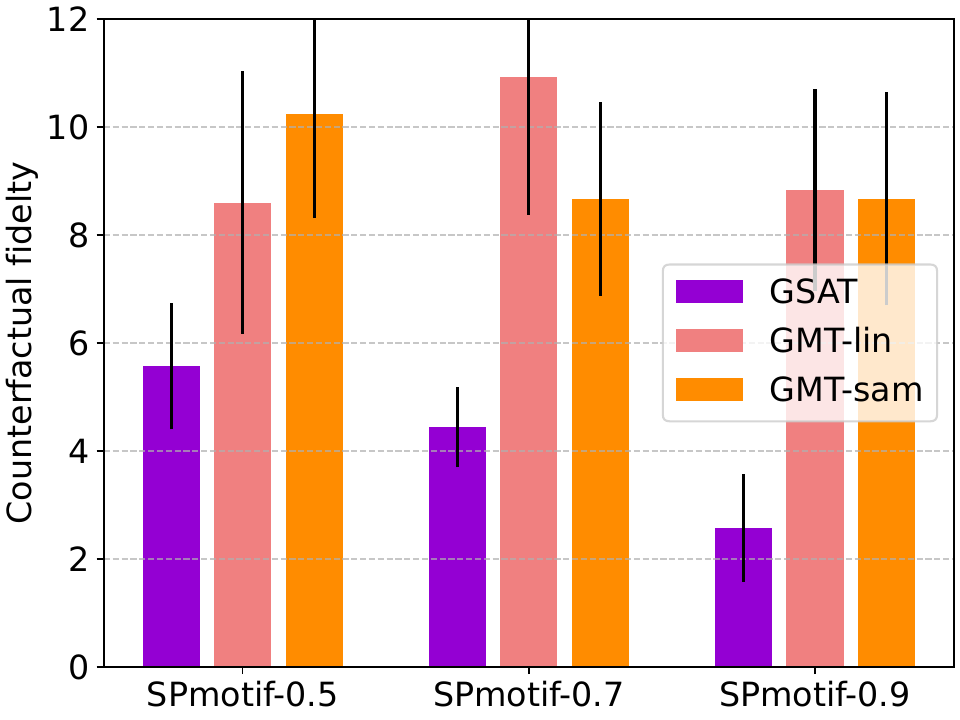}
        \label{fig:cf_results}
    }
    \subfigure[Interpretation sensitivity.]{
        \includegraphics[width=0.3\textwidth]{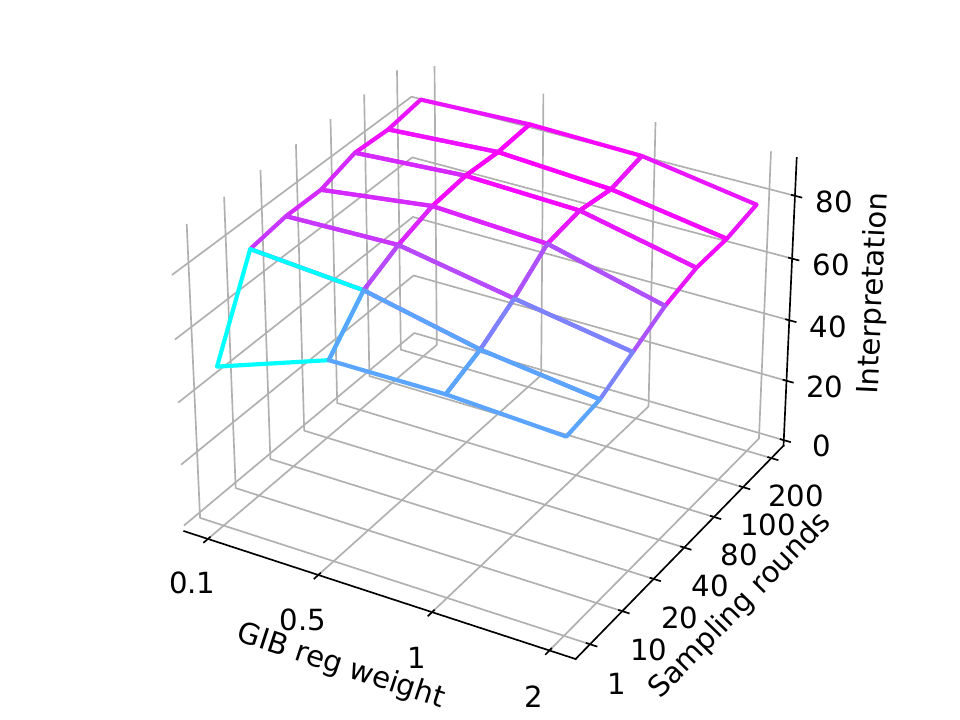}
        \label{fig:inter_sens}
    }
    \subfigure[Generalization sensitivity.]{
        \includegraphics[width=0.3\textwidth]{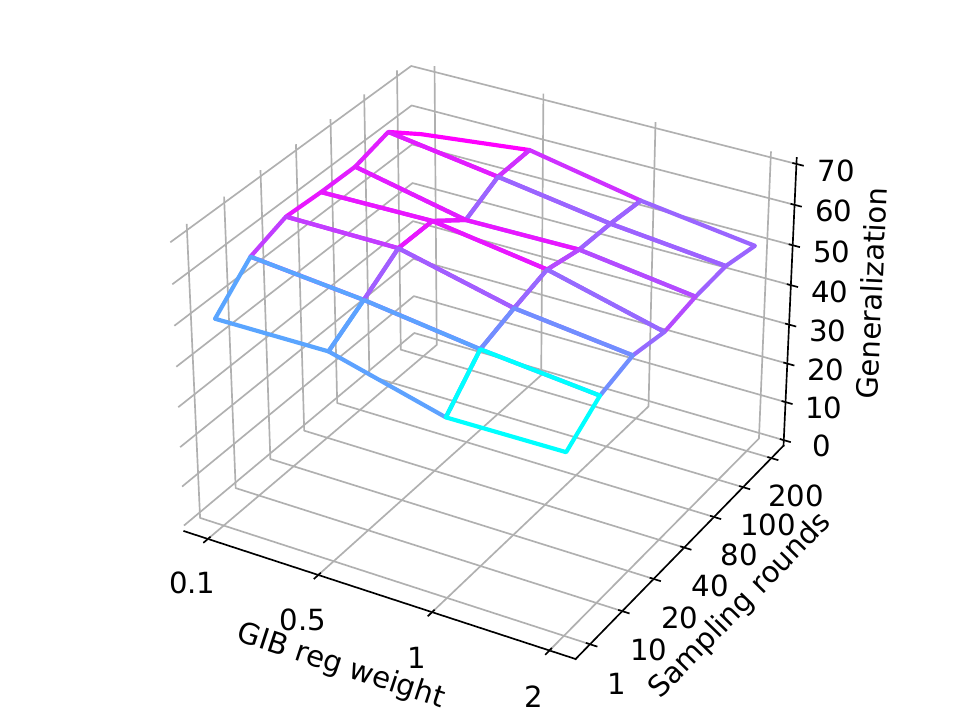}
        \label{fig:gen_sens}
    }
    \vspace{-0.15in}
    \caption{Ablation studies.}
    \label{fig:ablation}
\end{figure*}

In complementary to the interpretability and generalizability study, we conduct further ablation studies to better understand the results. Fig.~\ref{fig:cf_results} shows the counterfactual fidelity of \gsat, \oursl, and \ourss in Spurious-Motif (SPmotif) test sets.
As shown in Fig.~\ref{fig:cf_results} that \gsat achieves a lower counterfactual fidelity.
In contrast, \oursl and \ourss improve a higher counterfactual fidelity, which explains the reason for the improved interpretability of \ours.

\textbf{Hyperparameter sensitivity.} We also examine the hyperparameter sensitivity of \ourss in SPMotif-0.5 dataset.
As shown in Fig.~\ref{fig:inter_sens},~\ref{fig:gen_sens}, \ourss maintains strong robustness against the hyperparameter choices.
The interpretation performance gets improved along with the sampling rounds,
while using too-large GIB information regularizer weights will affect the optimization of \ours and the generalizability.

\begin{table}[t]
    \vspace{-0.1in}
    \caption{Comparison of prediction performances of \ourss and \gsat with and without retraining on SPMotif datasets. ``O'' refers to the originally trained classifier when training the subgraph extractor. ``R'' refers to the retrained classifier with a frozen subgraph extractor that has been trained with the corresponding methods.}
    \label{tab:retrain}
    \small\sc\centering
    \resizebox{0.48\textwidth}{!}{
        \begin{tabular}{@{}lllll@{}}
            \toprule
            \textbf{}       & \textbf{} & {SPMotif 0.5}      & {SPMotif 0.7}      & {SPMotif 0.9}      \\ \midrule
            { \gsat (GIN)}  & O         & { 47.45\std{5.87}} & { 43.57\std{2.43}} & { 45.39\std{5.02}} \\
            { }             & R         & { 46.54\std{3.90}} & { 45.09\std{4.48}} & { 47.76\std{4.50}} \\
            { \ourss (GIN)} & O         & { 46.85\std{4.57}} & { 48.82\std{2.87}} & { 46.77\std{2.93}} \\
            { }             & R         & { 60.09\std{5.57}} & { 54.34\std{4.04}} & { 55.83\std{5.68}} \\
            { \ourss (PNA)} & O         & { 60.02\std{3.97}} & { 61.67\std{3.16}} & { 52.98\std{2.55}} \\
            { }             & R         & { 71.98\std{3.44}} & { 69.68\std{3.99}} & { 67.90\std{3.60}} \\ \bottomrule
        \end{tabular}}
    \vspace{-0.15in}
\end{table}

\textbf{Effectiveness of classifier retraining for \ourss.}
As discussed in Sec.~\ref{sec:gmt_sam}, if the subgraph extractor extracts the desired causal subgraph, training the subgraph classifier with ERM could capture the desired correlations between the causal subgraph and the target label~\cite{ciga}.

\input{tables/interpretation_p2.tex}

To understand the rationale of retraining in \ourss, we conduct a comparison of the prediction performances of \ourss to \gsat with and without retraining. As shown in Table~\ref{tab:retrain}, with the classifier retraining, the prediction performances of \ourss increase significantly, since the subgraph extractor already gets rid of the spurious signals in the original graphs, compared to \gsat. The effectiveness of the retraining strategy also generalizes to PNA.

\textbf{More detailed interpretation performance comparison.} We conduct additional experiments with other post-hoc explanation metrics including Precision and mask entropies~\citep{Zorro}.
The results are given in Table~\ref{table:Interpretation_detailed}. From the table, we can find that, both Precision@5 and Average Precision align with the AUROC when reflecting the interpretation performance, despite of the numerical scale.
While \gsat suffers from lower Precision scores, \ourss demonstrates consistent and significant improvements than \gsat across both Precision@5 and Average Precision scores.
Meanwhile, both \gsat and \ourss have a relatively high sparsity, due to the mutual information regularization. Nevertheless, compared to \gsat, \ourss distributes the attention to more semantically meaningful edges in the underlying causal subgraph. Therefore, \ourss enjoys both high precision in terms of interpretability, and better generalizability, under a high sparsity.

\textbf{Results with XGNNs for node classification} are also conducted for which the details are left in Appendix~\ref{sec:node_xgnn}.
Specifically, we follow the experimental setup of a recent work called SunnyGNN~\citep{sunny_gnn} to evaluate the interpretation and prediction performances.
Since ground truth labels are not available, we mainly compare the counterfactual fidelity (Def.~\ref{def:counterfactual_x}) proposed in our work and sufficiency fidelity~\citep{sunny_gnn}.
The scale of the datasets ranges from Cora~\citep{cora} with $2,708$ nodes with $10,556$ edges to Coauthor-Physics~\citep{Shchur2018PitfallsOG} with $34,493$ nodes and $495,924$ edges.

In Table~\ref{tab:node} of Appendix~\ref{sec:node_xgnn}, we find that, compared to the state-of-the-art XGNNs in node classification, \ourss and \oursl achieve a competitive prediction performance while bringing significant improvements in terms of interpretation performance (including both sufficiency fidelty~\citep{sunny_gnn} and counterfactual fidelity (Def.~\ref{def:counterfactual_x}), aligned with our discussion. We also note that in some cases \oursl may underperform \gsat, highlighting an interesting direction for future investigation.

\textbf{More baseline results in PNA backbones} are given in Appendix~\ref{sec:more_x_appdx}, including two representative post-hoc methods GNNExplainer and PGExplainer, and one representative XGNN baseline DIR.
The results show that most of the baselines still significantly underperform \gsat and \ours.

\textbf{Computational analysis} is given in Appendix~\ref{sec:comp_appdx}. Although \ourss takes a longer time for training, but the absolute values are not high even for the largest dataset MNIST-75sp. When compared to other intrinsic interpretable methods,  \ourss consumes a similar training time of around 6 hours on MNIST-75sp as DIR.
As for inference,  \ourss enjoys a similar latency as others.

%% file: tables/interpretation_p2.tex
\begin{table*}[t]
    \vspace{-0.1in}
    \caption{More detailed comparion of interpretation performance (AUC) on regular graphs.}%
    \small\sc\centering
    \resizebox{\textwidth}{!}{
        \begin{tabular}{llcccccc}
            \toprule
            \multirow{2}{*}{GNN} & \multirow{2}{*}{Method} & \multicolumn{3}{c}{GIN} & \multicolumn{3}{c}{PNA}                                                                                 \\
                                 &                         & SPMotif-$0.5$           & SPMotif-$0.7$           & SPMotif-$0.9$     & SPMotif-$0.5$     & SPMotif-$0.7$     & SPMotif-$0.9$     \\
            \midrule\multirow{4}{*}{GSAT}
                                 & AUROC                   & {74.49\std{4.46}}       & {72.95\std{6.40}}       & {72.95\std{6.40}} & {79.84\std{3.21}} & {79.76\std{3.66}} & {80.70\std{5.45}} \\
                                 & Precision@5             & {46.69\std{2.26}}       & {42.96\std{7.42}}       & {35.34\std{4.71}} & {56.11\std{1.80}} & {58.95\std{0.99}} & {51.83\std{2.32}} \\
                                 & Avg. Precision          & {28.43\std{9.51}}       & {24.83\std{11.07}}      & {17.08\std{5.71}} & {35.12\std{5.00}} & {39.73\std{7.17}} & {40.20\std{4.86}} \\
                                 & Mask Entropy            & {4.70\std{0.21}}        & {4.71\std{0.13}}        & {4.70\std{0.17}}  & {4.70\std{0.03}}  & {4.72\std{0.09}}  & {4.71\std{0.07}}  \\
            \midrule\multirow{4}{*}{GMT-sam}
                                 & AUROC                   & {85.50\std{2.40}}       & {84.67\std{2.38}}       & {73.49\std{5.33}} & {88.27\std{1.71}} & {86.58\std{1.89}} & {85.26\std{1.92}} \\
                                 & Precision@5             & {56.56\std{1.80}}       & {53.11\std{3.41}}       & {40.43\std{5.66}} & {62.99\std{1.35}} & {61.44\std{1.46}} & {63.53\std{1.98}} \\
                                 & Avg. Precision          & {52.75\std{2.66}}       & {51.49\std{4.93}}       & {28.79\std{9.84}} & {58.11\std{1.81}} & {54.77\std{2.57}} & {54.12\std{3.00}} \\
                                 & Mask Entropy            & {4.70\std{0.21}}        & {4.71\std{0.11}}        & {4.71\std{0.13}}  & {4.70\std{0.06}}  & {4.71\std{0.13}}  & {4.71\std{0.17}}  \\
            \bottomrule
            \label{table:Interpretation_detailed}
        \end{tabular}}
    \vspace{-0.2in}
\end{table*}

%% file: sections/9_appdx.tex
\onecolumn
\appendix

\begin{center}
    \LARGE \bf {Appendix of \ourst}
\end{center}

\etocdepthtag.toc{mtappendix}
\etocsettagdepth{mtchapter}{none}
\etocsettagdepth{mtappendix}{subsection}
\tableofcontents

\clearpage
\section{Notations}
\label{sec:notations_appdx}
In the following, we list notations for key concepts that have appeared in this paper.
\begin{table}[ht]%
    \caption{Notations for key concepts involved in this paper.}
    \centering
    \resizebox{\textwidth}{!}{
        \begin{tabular}{ll}
            \toprule
            \(\gG\)                     & the graph space                                                                                          \\\midrule
            \(\gG_c\)                   & the space of subgraphs with respect to the graphs from $\gG$                                             \\\midrule
            \(\gY\)                     & the label space                                                                                          \\\midrule
            \(\rho\)                    & the pooling function of the GNN                                                                          \\\midrule
            \(d(\cdot,\cdot)\)          & a distribution distance metric                                                                           \\\midrule
            \(L(\cdot,\cdot)\)          & the loss       function                                                                                  \\\midrule
            \(G\in\gG\)                 & a graph                                                                                                  \\\midrule
            \(G=(A,X)\)                 & a graph with the adjacency matrix $A\in\{0,1\}^{n\times n}$ and node feature matrix $X\in\R^{n\times d}$ \\
                                        & for brevity, we also use $G$ and $Y$ to denote the random variables as the graphs and labels             \\\midrule
            \(f=f_c\circ g\)            & a \xgnn with a subgraph extractor $g$ and a classifier $f_c$                                             \\\midrule
            \(g\)                       & a subgraph extractor $g:\gG\rightarrow\gG_c$                                                             \\\midrule
            \(f_c\)                     & a classifier GNN $f_c:\gG_c\rightarrow\gY$                                                               \\\midrule
            \(G_c\)                     & the invariant subgraph with respect to $G$                                                               \\\midrule
            \(G_s\)                     & the spurious subgraph with respect to $G$                                                                \\\midrule
            \(\pred{A}_c,\pred{A}\)     & the weighted adjacency matrix for causal subgraph with entries $A_{u,v}=\alpha_e$                        \\
                                        & as the sampling probability predicted by $g$                                                             \\\midrule
            \(\pred{A}_s\)              & the weighted adjacency matrix for spurious subgraph with entries $A_{u,v}=1-\alpha_e$                    \\
                                        & as the sampling probability predicted by $g$                                                             \\\midrule
            \(\pred{G}_c\)              & the estimated invariant subgraph produced by $g$                                                         \\
                                        & if the subgraph partitioning is conducted in an edge-centric view, then $\pred{G}_c=(X,\pred{A}_c)$      \\\midrule
            \(\pred{G}_s\)              & the estimated spurious subgraph produced by tacking the complementary of $\pred{G}_c$                    \\
                                        & if the subgraph partitioning is conducted in an edge-centric view, then $\pred{G}_s=(X,\pred{A}_s)$      \\\midrule
            \(I(G_c;Y)\)                & mutual information between the extracted subgraph $G_c$ and $Y$, specialized for maximizing $I(G;Y)$     \\\midrule
            \(P(G_c|G)\in\R_+\)         & the probability for sampling $G_c$ from $G$ with the subgraph extractor $g$                              \\\midrule
            \(P(Y|G)\in\R^{|\gY|}_+\)   & the label distribution of $Y$ conditioned on $G$                                                         \\\midrule
            \(P_f(Y|G)\in\R^{|\gY|}_+\) & the predicted label distribution of $Y$ conditioned on $G$                                               \\\midrule
            \(f_c(G_c)\in\R^{|\gY|}_+\) & the predicted label distribution of $Y$ with $f_c$ by taking the input $G_c$. \\
            &If the input graph is a weighted graph, $f_c$ computes the label prediction with weighted message passing. 
            \\\midrule
            \bottomrule
        \end{tabular}}
\end{table}
\clearpage
\section{More Details about the Background}
\label{sec:prelim_appdx}
We begin by introducing related works in Appendix~\ref{sec:related_appdx} and then more backgrounds about graph information bottleneck in Appendix~\ref{sec:GCE_deduce_appdx}, especially for how to obtain the formulas in the main text.
\subsection{More related works}
\label{sec:related_appdx}
We give a more detailed background introduction of interpretable and generalizable GNNs (\xgnns) in this section.

\paragraph{Graph Neural Networks.}
We use $G=(A,X)$ to denote a graph with $n$ nodes and $m$ edges.
Within $G$, $A \in \{0,1\}^{n\times n}$ is the adjacency matrix, and $X\in \R^{n \times d}$ is the node feature matrix with a node feature dimension of $d$.
This work focuses on the task of graph classification.
Specifically, we are given a set of $N$ graphs $\{G_i\}_{i=1}^N\subseteq \gG$
and their labels $\{Y_i\}_{i=1}^N\subseteq\gY=\R^c$ from $c$ classes.
Then, we need to train a GNN $\rho \circ h$ with an encoder $h:\gG\rightarrow\R^h$ that learns a meaningful representation $h_G$ for each graph $G$ to help predict their labels $y_G=\rho(h_G)$ with a downstream classifier $\rho:\R^h\rightarrow\gY$.
The representation $h_G$ is typically obtained by performing pooling with a $\text{READOUT}$ function on the learned node representations:
\begin{equation}
    \label{eq:gnn_pooling}
    h_G = \text{READOUT}(\{h^{(K)}_u|u\in V\}),
\end{equation}
where the $\text{READOUT}$ is a permutation invariant function (e.g., $\text{SUM}$, $\text{MEAN}$)~\citep{gin},
and $h^{(K)}_u$ stands for the node representation of $u\in V$ at $K$-th layer that is obtained by neighbor aggregation:
\begin{equation}
    \label{eq:gnn}
    h^{(K)}_u = \sigma(W_K\cdot a(\{h^{(K-1)}_v\}| v\in\mathcal{N}(u)\cup\{u\})),
\end{equation}
where $\mathcal{N}(u)$ is the set of neighbors of node $u$,
$\sigma(\cdot)$ is an activation function, e.g., $\text{ReLU}$, and $a(\cdot)$ is an aggregation function over neighbors, e.g., $\text{MEAN}$.

\paragraph{Interpretable GNNs.}
Let $G=(A,X)$ be a graph with node set $V=\{v_1,v_2,...,v_n\}$ and edge set $E=\{e_1,e_2,...,e_m\}$,
where  $A \in \{0,1\}^{n\times n}$  is the adjacency matrix and $X\in \R^{n \times d}$ is the node feature matrix.
In this work, we focus on interpretable GNNs (or \xgnns) for the graph classification task, while the results can be generalized to node-level tasks as well~\citep{gib_node}.
Given each sample from training data $\train=(G^i,Y^i)$,
an interpretable GNN $f:=h\circ g$ aims to identify a (causal) subgraph $G_c\subseteq G$ via a subgraph extractor GNN $g:\gG\rightarrow\gG_c$, and then predicts the label via a subgraph classifier GNN $f_c:\gG_c\rightarrow\gY$, where $\gG,\gG_c,\gY$ are the spaces of graphs, subgraphs, and the labels, respectively~\citep{gib}.
Although \textit{post-hoc} explanation approaches also aim to find an interpretable subgraph as the explanation for the model prediction~\citep{gnn_explainer,xgnn,pgm_explainer,pge,subgraphxgn,gen_xgnn,orphicx,Zorro,meta_ex_gnn,sunny_gnn}, they are shown to be suboptimal in interpretation performance and sensitive to the performance of the pre-trained GNNs~\citep{gsat}.
Therefore, this work focuses on \textit{intrinsic interpretable} GNNs (XGNNs).

A predominant approach to implement \xgnns is to incorporate the idea of information bottleneck~\citep{ib}, such that $G_c$ keeps the minimal sufficient information of $G$ about $Y$~\citep{gib,vgib,gsat,lri,gib_hiera},
which can be formulated as
\begin{equation}
    \max_{G_c}I(G_c;Y)-\lambda I(G_c;G),\ G_c\sim g(G),
\end{equation}
where  maximizing the mutual information between $G_c$ and $Y$ endows the interpretability of $G_c$ while minimizing $I(G_c;G)$ ensures $G_c$ captures only the most necessary information, $\lambda$ is a hyperparamter trade off between the two objectives.
In addition to minimizing $I(G_c;G)$, there are also alternative approaches that impose different constraints such as causal invariance~\citep{ciga,gil} or disentanglement~\citep{dir,cal,grea,disc} to identify the desired subgraphs.
When extracting the subgraph, \xgnns adopts the attention mechanism to learn the sampling probability of each edge or node, which avoids the complicated Monte Carlo tree search used in other alternative implementations~\citep{protGNN}.
Specifically, given node representation learned by message passing $H_i\in\R^h$ for each node $i$, \xgnns either learns a \textbf{node attention} $\alpha_i\in\R_+=\sigma(a(H_i))$ via the attention function $a:\R^h\rightarrow\R_+$, or the \textbf{edge attention} $\alpha_e\in\R_+=\sigma(a([H_u,H_v]))$ for each edge $e=(u,v)$ via the attention function $a:\R^{2h}\rightarrow\R_+$, where $\sigma(\cdot)$ is a sigmoid function. $\boldsymbol{\alpha}=[\alpha_1,...,\alpha_m]^T$ essentially elicits a subgraph distribution of the interpretable subgraph. In this work, we focus on edge attention-based subgraph distribution as it is most widely used in \xgnns while our method can be easily generalized to node attention-based subgraph approaches as demonstrated in the experiments with geometric learning datasets.

Besides, \citet{gatv2,aero_gnn} find the failures of graph attention networks in properly propagating messages with the attention mechanism. They differ from our work as they focus on node classification tasks.

\textbf{Other interpretable GNNs.}
\citet{meta_ex_gnn} develop interpretable and efficient GNNs that leverage the attention mechanism to fully and efficiently exploit the useful meta-paths in heterogeneous graphs. In our aforementioned preliminary experiments, we have demonstrated the promising results of GMT in node classification tasks. Therefore, we believe it’s a promising future direction to extend GMT to heterogeneous graphs based on the framework proposed in~\citet{meta_ex_gnn}.
\citet{graphlime} incorporate Hilbert-Schmidt Independence Criterion Lasso to find interpretable features as local explanations for the task of node classification. Different from \citet{graphlime}, we focus on finding subgraphs of the inputs as interpretations.

\textbf{Faithful interpretation and (OOD) generalization.}
The faithfulness of interpretation is critical to all interpretable and explainable methods~\citep{fidelity,mythos_inter,robust_xnn,Rudin2018StopEB,att_not_exp,relation_exp_pred}.
Yet, there are many failure cases found especially when with attention mechanisms. For example, \citet{att_not_exp} reveals that in NLP, randomly shuffling or imposing adversarial noises will not affect the predictions too much, highlighting a weak correlation between attention and prediction.
\citet{relation_exp_pred} present a causal analysis showing the hyperparameters and the architecture setup could be a cofounder that affects the causal analysis. \citet{inv_rat} show interpretations will fail when  distribution shifts are presented.
Although the faithfulness of explanation/interpretations has been widely a concern for Euclidean data, whether and how GNNs and \xgnns suffer from the same issue is under-explored.

Talking about the progress in graph data, there are several metrics developed to measure the faithfulness of graph explanations, such as fidelity~\citep{xgnn_tax,GraphFramEx}, counterfactual robustness~\citep{RCExplainer,counterfactual_xgnn_sur,clear},  equivalence~\citep{xgnn_equi}, and robust fidelity~\citep{zheng2024robust_fid}, which are however limited to post-hoc graph explanation methods.
In fact, post-hoc explanation methods are mostly developed to adhere the faithfulness measures such as fidelity. However, as shown by~\citet{gsat}, the post-hoc methods are suboptimal in finding the interpretable subgraph and sensitive to the pre-trained model, which highlights a drawback of the existing faithfulness measure.
In contrast, we develop the first faithfulness measure for \xgnns in terms of counterfactual invariance.

Although \citet{RCExplainer,counterfactual_xgnn_sur,clear} also adopt the concept of counterfactual to develop post-hoc explanation methods, they focus on finding the minimal perturbations that will change the predictions. Counterfactual is also widely used to improve graph representation learning~\citep{counterfactual_gl}.
In contrast, we adopt the concept of counterfactual to measure the sensitivity of the \xgnns predictions to the predicted attention.

\input{figures/full_scm.tex}

\paragraph{On the natural connection of \xgnns and OOD generalization on graphs.}
In the context of graph classification, the generalization ability and the faithfulness of the interpretation are naturally intertwined in \xgnns.
In many realistic graph classification practices such as drug property prediction~\citep{drugood,ai4sci}, the property of a drug molecule can naturally be represented by a subgraph, termed as causal subgraph. The causal subgraph, in return, holds a causal relationship with the drug property. Therefore, it is natural to identify the underlying causal subgraph to provide OOD generalizable predictions and interpretations.

Typically, \xgnns need to extract the underlying ground truth subgraph in order to make correct predictions on unseen test graphs~\citep{gsat}. When distribution shifts are presented in the test data, it is critical to find the underlying subgraph that has a causal relationship with the target label (or causal subgraphs) ~\citep{inv_rat,ciga}.

We now briefly introduce the background of causal subgraph and OOD generalization.
Specifically,
we are given a set of graph datasets $\dataset=\{\dataset_e\}_e$ collected from multiple environments $\envall$.
Samples $(G^e_i, Y^e_i)\in \dataset^e$ from the same
environment are considered as drawn independently from an identical distribution $\sP^e$.
We consider the graph generation process proposed
by~\citet{ciga} that covers a broad case of graph distribution shifts.
Fig.~\ref{fig:scm_appdx} shows the full graph generation process considered in~\citet{ciga}.
The generation of the observed graph $G$ and labels $Y$
are controlled by a set of latent causal variable $C$ and spurious variable $S$, i.e.,
\[G\coloneqq f_\gen(C,S).\]
$C$ and $S$ control the generation of $G$ by controlling the underlying invariant subgraph $G_c$
and spurious subgraph $G_s$, respectively.
Since $S$ can be affected by the environment $E$,
the correlation between $Y$, $S$ and $G_s$ can change arbitrarily
when the environment changes.
$C$ and $S$ control the generation of the underlying invariant subgraph $G_c$
and spurious subgraph $G_s$, respectively.
Since $S$ can be affected by the environment $E$,
the correlation between $Y$, $S$ and $G_s$ can change arbitrarily
when the environment changes.
Besides, the latent interaction among $C$, $S$ and $Y$
can be further categorized into \emph{Full Informative Invariant Features} (\emph{FIIF})
when $Y\ind S|C$ and \emph{Partially Informative Invariant Features} (\emph{PIIF}) when $Y \not\ind S|C$. Furthermore, PIIF and FIIF shifts can be mixed together and yield \emph{Mixed Informative Invariant Features} (\emph{MIIF}), as shown in Fig.~\ref{fig:scm_appdx}.
We refer interested readers to~\citet{ciga} for a detailed introduction to the graph generation process.

To tackle the OOD generalization challenge on graphs generated following in Fig.~\ref{fig:scm_appdx},
the existing invariant graph learning approaches generically
aim to identify the underlying invariant subgraph $G_c$ to predict the label $Y$~\citep{eerm,ciga}.
Specifically, the goal of OOD generalization on graphs
is to learn an invariant \xgnn $f\coloneqq f_c\circ g$, with the following objective:
\begin{equation}
    \label{eq:inv_cond_appdx}
    \text{$\max$}_{f_c, \; g} \ I(\pred{G}_{c};Y), \ \text{s.t.}\ \pred{G}_{c}\ind E,\ \pred{G}_{c}=g(G).
\end{equation}
Since $E$ is not observed, many strategies are proposed to
impose the independence of $\pred{G}_c$ and $E$.
A common approach is to augment the environment information.
For example, based on the estimated invariant subgraphs $\pred{G}_c$ and spurious subgraphs $\pred{G}_s$,
\citet{dir,grea,eerm,dps} propose to generate new environments, while \citet{gil} propose to infer the underlying environment labels via clustering.
\citet{moleood} propose a variational framework to infer the environment labels.
\citet{joint_causal_indep} propose to learn causal independence between labels and environments.
\citet{gib,vgib,gsat,lri,gib_hiera} adopt graph information bottleneck to tackle FIIF graph shifts, and they cannot generalize to PIIF shifts. \citet{gala} show the pitfall of the environment generation or augmentation methods for PIIF shifts, and propose to adopt an environment assistant to resolve the issues.
Nevertheless, since most of the existing works adopt the backbone of \xgnns, and \xgnns with information bottleneck is the state-of-the-art method with both high interpretation performance and OOD generalization performance, the focus in this work will be around tackling FIIF shifts with the principle of graph information bottleneck. More details are given in the next section.

In addition to the aforementioned approaches, \citet{size_gen1,size_gen2,size_gen3} study the OOD generalization as an extrapolation from small graphs to larger graphs in the task of graph classification and link prediction. In contrast, we study OOD generalization against various graph distribution shifts formulated in Fig.~\ref{fig:scm_appdx}.
\citet{graph_joint_extra} propose an extrapolation strategy to improve OOD generalization on graphs.
In addition to the standard OOD generalization tasks studied in this paper, \citet{nn_extrapo,OOD_CLRS} study the OOD generalization in tasks of algorithmic reasoning on graphs. \citet{graph_ttt} study the test-time adaption in the graph regime. \citet{shape_matching} study the 3D shape matching under the presence of noises.

\textbf{Multilinear extension.}
Multilinear extension serves as a powerful tool
for maximizing combinatorial functions, especially for submodular set function maximization \citep{mt_game,mt,Vondrak08,Calinescu11,Chekuri14,Chekuri15,optimal_drsub,sets2multisets,bian2022energybased,neural_set}.
For example, \citet{Vondrak08,Calinescu11} study the multilinear extension in the context of social welfare. \citet{bian2022energybased} study the multilinear extension for cooperative games.
It is the expected value of a set function under the fully factorized i.i.d. Bernoulli distribution.
The closest work to ours is~\citet{neural_set} that builds neural set function extensions for multiple discrete functions.
Nevertheless, to the best of our knowledge, the notion of multilinear extensions for \xgnns is yet underexplored.
In contrast, in this work, we are the first to identify subgraph multilinear extension as the factorized subgraph distribution for interpretable subgraph learning.

\subsection{Variational bounds and realization of the IB principle}
\label{sec:GCE_deduce_appdx}
We first introduce how to derive Eq.~\ref{eq:GCE} in the main text, and then discuss how to implement the graph information bottleneck regularization $\min I(G_c;G)$ following the state-of-the-art architecture \gsat~\citep{gsat,lri}.

\paragraph{Variational bounds for $I(G;Y)$.}
For the term $I(G;Y)$, notice that

\begin{equation}\label{}
    \begin{aligned}
        I(G;Y) = \E_{G, Y} \left[ \log \frac{P(Y|G)}{P(Y)}  \right]
    \end{aligned}
\end{equation}
Since the true $P(Y|G)$ is intractable, through XGNN modelling we introduce a variational approximation $P_{f_c, g}(Y|G)$. Then,
\begin{align}
    I(G;Y) & = \E_{G, Y} \left[ \log \frac{P_{f_c, g}(Y|G)}{P(Y)} \right]  +  \E_{G, Y} \left[ \log \frac{P(Y|G)}{P_{f_c, g}(Y|G)} \right] \\
           & = \E_{G, Y} \left[ \log \frac{P_{f_c, g}(Y|G)}{P(Y)} \right]  +  \KL (P(Y|G) || P_{f_c, g}(Y|G))                              \\
           & \geq  \E_{G, Y} \left[ \log P_{f_c, g}(Y|G)  \right]  + H(Y)
\end{align}
Since the optimization does not involve $H(Y) $, we continue with
$\E_{G, Y} \left[ \log P_{f_c, g}(Y|G)  \right] $,
\begin{align}
    \E_{G, Y} \left[ \log P_{f_c, g}(Y|G)  \right] & =  \E_{G, Y} \left[ \log \sum_{G_c} P_{f_c, g}(Y, G_c|G)  \right]                   \\
                                                   & =  \E_{G, Y} \left[ \log \sum_{G_c} P_{f_c, g}(Y|G, G_c)  P_{f_c, g}(G_c|G) \right] \\  \label{overall_xgnn_imple}
                                                   & = \E_{G, Y} \left[ \log \sum_{G_c} P_{f_c}(Y|G_c)  P_{g}(G_c|G) \right]
\end{align}
where Eq. \ref{overall_xgnn_imple} is due to the implementation of XGNNs.
Eq. \ref{overall_xgnn_imple} can also be written with expectations:
\[
    \E_{G, Y} \left[ \log \sum_{G_c} P_{f_c}(Y|G_c)  P_{g}(G_c|G) \right]=
    \E_{G, Y} \left[ \log \E_{G_c\sim \mathbb{P}(G_c|G)} P_{f_c}(Y|G_c) \right].
\]
Maximizing $I(G;Y)$ is  equivalent to minimizing $-I(G;Y)$, and further minimizing $\E_{G,Y}[-\log P_{f_c,g}(Y|G)]$.
This achieves to Eq.~\ref{eq:GCE} in the main text, i.e.,
\begin{align}
    \E_{G, Y} \left[ - \log \E_{G_c\sim \mathbb{P}(G_c|G)} P_{f_c}(Y|G_c) \right]  =\E_{(G,Y)} [L( \E_{G_c\sim \mathbb{P}(G_c|G)} [f_c(G_c)], Y)],
\end{align}
with $L$ as the cross entropy loss.
$\boldsymbol{\alpha}$ factorizes the sampling probability of the subgraphs as independent Bernoulli distributions on edges $e\sim \text{Bern}(\alpha_e), \forall e\in E$:
\[
    P(G_c|G)=\prod_{e\in G_c}\alpha_e\prod_{e\in G/G_c}(1-\alpha_e).
\]

\paragraph{Variational bounds for $I(G_c;G)$.}
For the term $I(G_c;G)$, since we factorize graph distribution as multiple independent Bernoulli distributions on edges, we are able to calculate the KL divergence to upper bound $I(G_c;G)$:
\begin{align}
    I(G_c;G)\leq D_\text{KL}(P(G_c|G)||Q(G_c)),
\end{align}
where $Q(G_c)$ is a variational approximation to $P(G_c)$. $D_\text{KL}$ can be obtained via
\begin{align}
    D_\text{KL}(P(G_c|G)||Q(G_c))=\sum_{e\in G_c}D_\text{KL}(\text{Bern}(\alpha_e)||\text{Bern}(r))+c(n,r),
\end{align}
where $c(n,r)$ is a small constant, $r$ is a hyperparameter to specify the prior for subgraph distributions. To minimize $I(G_c;G)$ is essentially to minimize $D_\text{KL}(\text{Bern}(\alpha_e)||\text{Bern}(r))$. The KL divergence can be directly calculated as
\begin{equation}\label{eq:gib_reg_appdx}
    D_\text{KL}(\text{Bern}(\alpha_e)||\text{Bern}(r))=\sum_{e}\alpha_e\log\frac{\alpha_e}{r}+(1-\alpha_e)\log\frac{(1-\alpha_e)}{(1-r)}.
\end{equation}
\citet{gsat} find the mutual information based regularization can effectively regularize the information contained in $G_c$ than previous implementations such as vanilla size constraints with the norm of attention scores or connectivity constraints~\citep{gib}.

Besides, we would like to note that  \gsat implementation provided by the author  does not exactly equal to the mathematical formulation, i.e., they directly take the unormalized attention to Eq.~\ref{eq:gib_reg_appdx}, as acknowledged by the authors~\footnote{\url{https://github.com/Graph-COM/GSAT/issues/10}}.
The reason for using another form of information regularization is because the latter empirically performs better.
Nevertheless, \lri adopts the mathematically correct form and obtains better empirical performance. In our experiments, we adopt the mathematically correct form for both regular and geometric learning tasks, in order to align with the theory. Empirically, we find the two forms perform competitively well with the suggested hyperparemters and hence stick to the mathematically correct form.

\section{On the Generalization and Interpretability: A Causal View}
\subsection{Structural Causal Model for \xgnns}
\label{sec:scm_xgnn_appdx}
We provide a detailed description and the full structural causal model of \xgnns in complementary to the causal analysis in Sec.~\ref{sec:causal_view}.
\begin{figure}[H]
    \centering
    \includegraphics[scale=0.8]{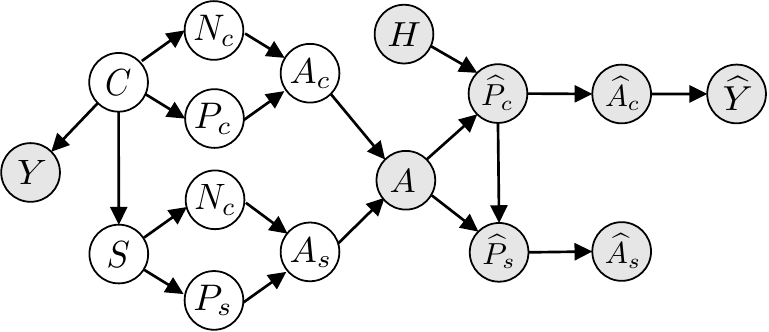}
    \caption{Bernoulli Parameterized SCM for interpretable GNN}
    \label{fig:scm_ber_full_appdx}
\end{figure}
\paragraph{Data generation.} We consider the same data model as previous works~\citep{size_gen2,gsat,ciga},
where the underlying causal subgraph $G_c$ and the spurious subgraph $G_s$ will be assembled via some underlying assembling process $G = f_g(G_c,G_s)$,
as illustrated in Appendix~\ref{sec:prelim_appdx} Fig.~\ref{fig:scm_appdx}.

We focus on the FIIF distribution shifts (Fig.~\ref{fig:scm_fiif_appdx}) that can be resolved by graph information bottleneck~\citep{gsat,ciga}.
As shown in the figure, there are latent causal and spurious variables $C$ and $S$ that have an invariant and spurious correlation with the label $Y$, respectively. $C$ and $S$ further control the generation of the graph structure of the causal subgraph $G_c$, and the spurious subgraph $G_s$.
Specifically, $C$ and $S$ will specify the number of nodes in $G_c$ and $G_s$ as $N_c$ and $N_s$. Then, $C$ and $S$ further control the underlying Bernoulli distributions on edges, by specifying the sampling probability as $P_c$ and $P_s$. With $N_c$ and $P_c$ (or $N_s$ and $P_s$), $A_c$ (or $A_s$) can be sampled and then assembled into the observed graph structure $A$.
As we focus on the edge-centric view, our discussion focuses on the graph structures $A_c$ and $A_s$ of the subgraphs.
Nevertheless, a similar generation model can also be developed for the node-centric view.

\paragraph{Interpretation.} Correspondingly, \xgnns first uses a subgraph extractor to predict the causal and spurious subgraphs $\widehat{G}_c$ and $\widehat{G}_s$, respectively.
The extraction aims to reverse the generation and recover the underlying $P_c$, by learning the $\widehat{P}_c$ via the attention $\boldsymbol{\alpha}$. We denote the architecture and the hyperparameter settings  as $H$.
Once $\widehat{P}_c$ is determined, $\widehat{P}_s=1-\widehat{P}_c$ is also obtained by finding the complementary part. Then, the estimated causal and spurious subgraphs are sampled from $\widehat{P}_c$ and $\widehat{P}_s$, respectively.
With the estimated causal subgraph $\widehat{G}_c=(X,\widehat{A}_c)$, the classifier GNN $c(\cdot)$ will use it to make a prediction $\widehat{Y}$.

\subsection{Practical Estimation of Counterfactual Fidelity}
\label{sec:pratical_cf_appdx}
Since it is prohibitively expensive to enumerate all possible $\widetilde{G}$ and the distance $\delta$ to examine the counterfactual fidelity. We instead consider an alternative notion that adopts random perturbation onto the learned attention score.
Specifically, we consider a random attention matrix $\widetilde{A}\sim\sigmoid(\gN(\mu_{\widehat{H}_A},\sigma_{\widehat{H}_A}))$,
where $\mu_{\widehat{H}_A}$ and $\sigma_{\widehat{H}_A}$ are the mean and standard deviation of the pre-attention matrix $\widehat{H}_A$ (The adjacency matrix with the unnormalized attention).
Since each non-symmetric entry in the attention is generated independently,
each non-symmetric entry in $\widetilde{A}$ is sampled independently following the factorization of $P(G)$.
We randomly sample $\widetilde{A}$ by $k$ times and calculate the following:
\begin{equation}\label{eq:cf_appdx}
    c_{\widehat{G}_c} = \frac{1}{k}\sum_{i=1}^k d(f_c(Y|\widetilde{G}^i_c),f_c(Y|\widehat{G}_c)),
\end{equation}
where $\widetilde{G}^i_c=(X,\widetilde{A}^i_c)$ and $d$ is total variation distance.
The detailed computation of the practical counterfactual fidelity is provided in Algorithm~\ref{alg:counterfactual_fidelity}.

\begin{algorithm}[ht]
    \caption{Practical estimation of counterfactual fidelity. }
    \label{alg:counterfactual_fidelity}
    \begin{algorithmic}[1]
        \STATE \textbf{Input:} Training data $\train$;
        a trained \xgnn $f$ with subgraph extractor $g$, and classifier $f_c$;
        sampling times $e_s$;
        batch size $b$;
        total variation distance $d(\cdot)$;
        \STATE \texttt{// Minibatch sampling.}
        \FOR{$j=1$ to $|\train|/b$}
        \STATE Sample a batch of data $\{G^i,Y^i\}_{i=1}^b$ from $\train$;
        \STATE Obtain the pre-attention matrix $\widehat{H}_A$;
        \STATE Obtain the attention matrix $\widehat{A}=\sigmoid(\widehat{H}_A)$;
        \STATE Obtain the original prediction with $f_c$ based on the attention matrix $\widehat{A}$ as $\{\hat{y}^i\}_{i=1}^b$;
        \STATE \texttt{// Random noises injection.}
        \FOR{$k=1$ to $e_s$}
        \STATE Sample a random attention matrix $\widetilde{A}\sim\sigmoid(\gN(\mu_{\widehat{H}_A},\sigma_{\widehat{H}_A}))$;
        \STATE Obtain sampling attention $\{\boldsymbol{\alpha}^i\}_{i=1}^b$;
        \STATE Obtain the perturbed prediction with $f_c$ based on the attention matrix $\widetilde{A}$ as $\{\hat{y}^i_k\}_{i=1}^b$;
        \ENDFOR
        \STATE Calculate $\{c_{\widehat{G}_c}^i\}_{i=1}^b$ with $k$ groups of $\{\hat{y}^i_k\}_{i=1}^b$ and $\{\hat{y}^i\}_{i=1}^b$;
        \STATE Obtain the averaged $c_{\widehat{G}_c}^j$ within the batch;
        \ENDFOR
        \STATE Obtain the averaged $c_{\widehat{G}_c}$ within the training data;
        \STATE \textbf{Return} estimated $c_{\widehat{G}_c}$;
    \end{algorithmic}
\end{algorithm}

\begin{figure}[ht]
    \centering
    \subfigure[\smt on BA-2Motifs trainset.]{
        \includegraphics[width=0.31\textwidth]{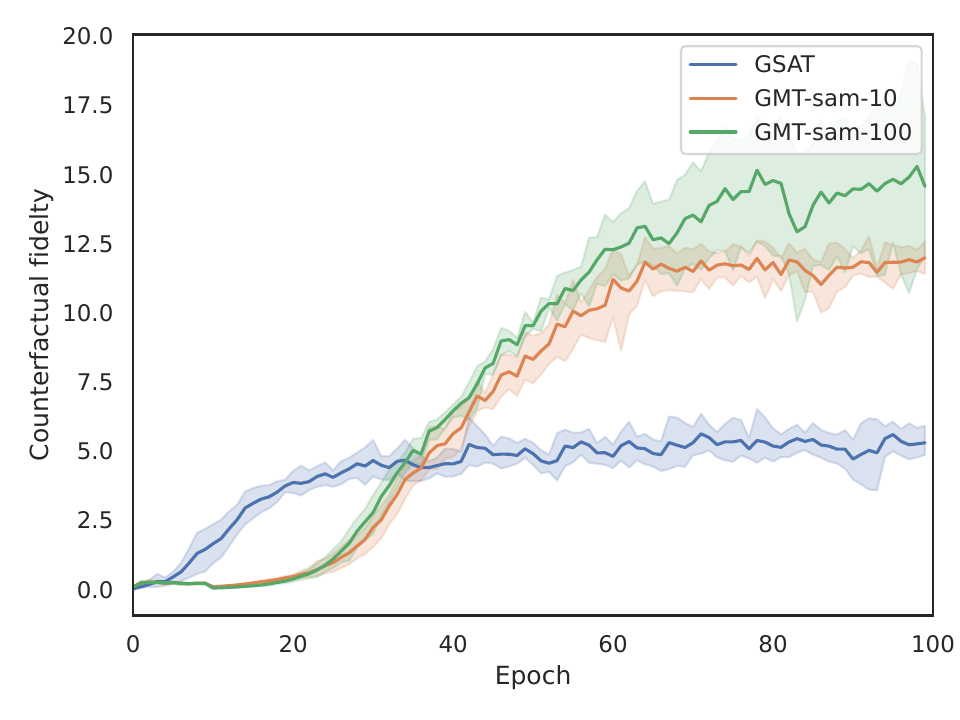}
        \label{fig:cf_ba_train_appdx}
    }
    \subfigure[\smt on BA-2Motifs valset.]{
        \includegraphics[width=0.31\textwidth]{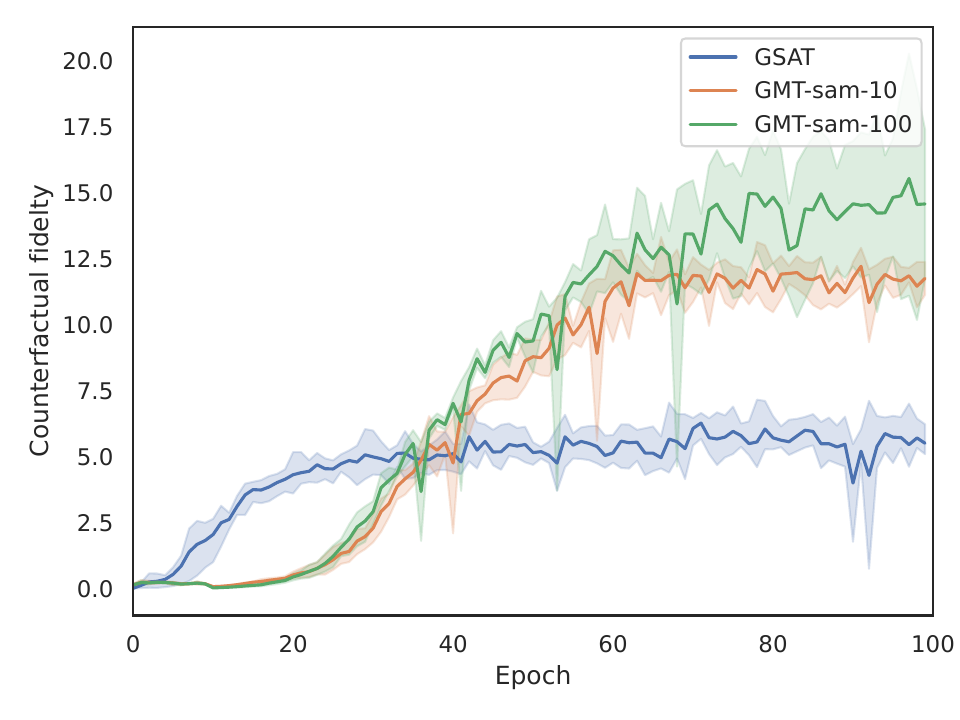}
        \label{fig:cf_ba_val_appdx}
    }
    \subfigure[\smt on BA-2Motifs test set.]{
        \includegraphics[width=0.31\textwidth]{figures/cf_ba_test_logits_tvd.pdf}
        \label{fig:cf_ba_test_appdx}
    }
    \caption{Counterfactual fidelity on BA-2Motifs.}
    \label{fig:cf_ba_appdx}
\end{figure}
\begin{figure}[ht]
    \centering
    \subfigure[\smt on Mutag trainset.]{
        \includegraphics[width=0.31\textwidth]{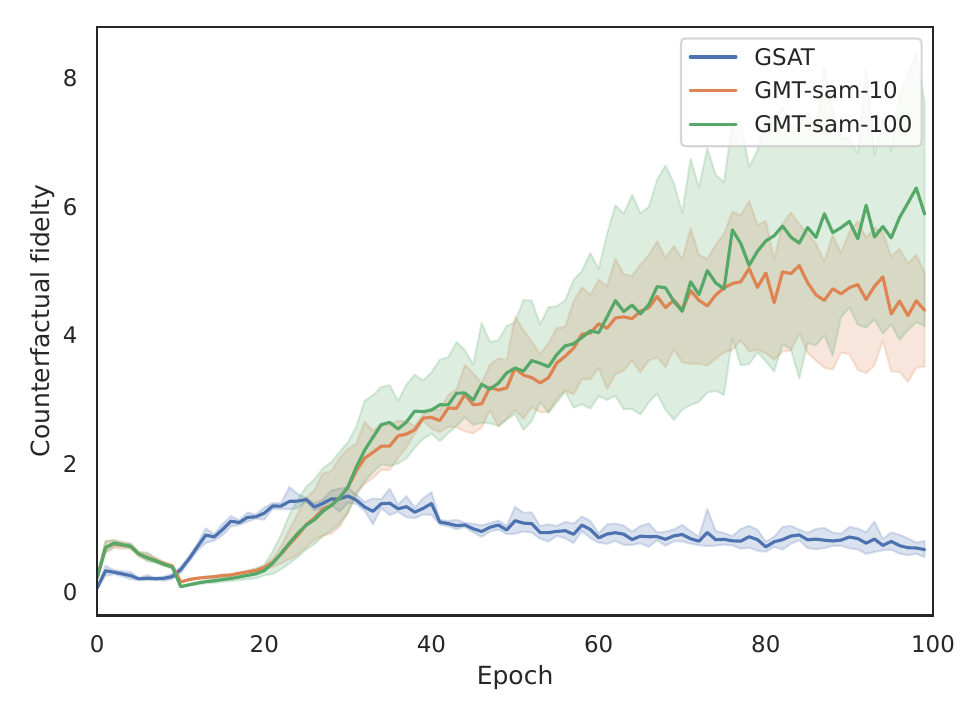}
        \label{fig:cf_mu_train_appdx}
    }
    \subfigure[\smt on Mutag validation set.]{
        \includegraphics[width=0.31\textwidth]{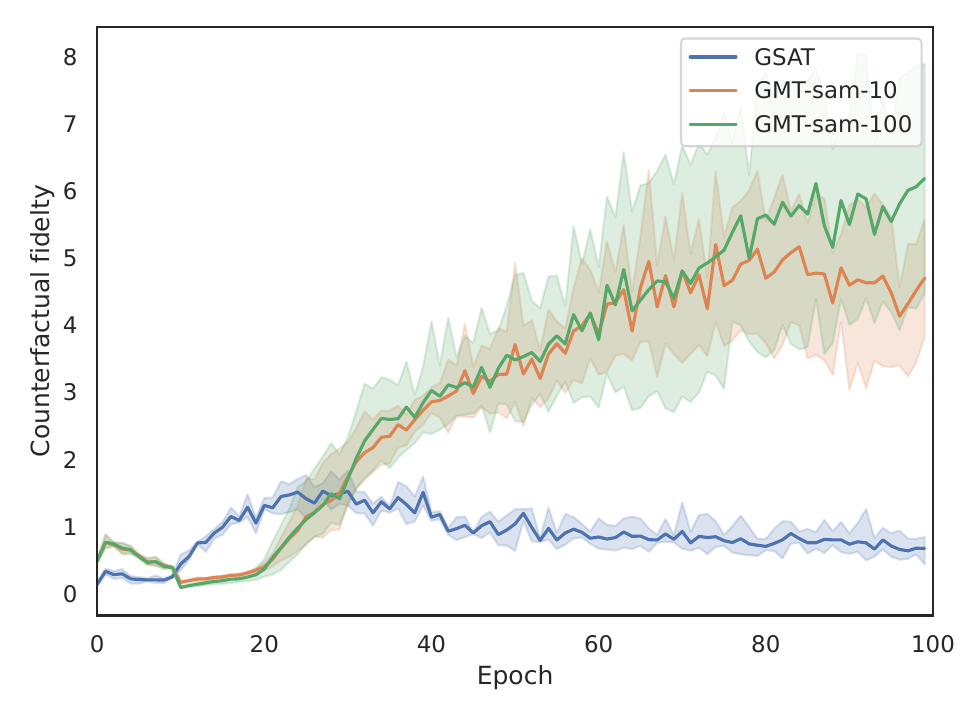}
        \label{fig:cf_mu_val_appdx}
    }
    \subfigure[\smt on Mutag test set.]{
        \includegraphics[width=0.31\textwidth]{figures/cf_mu_test_logits_tvd.pdf}
        \label{fig:cf_mu_test_appdx}
    }
    \caption{Counterfactual fidelity on Mutag.}
    \label{fig:cf_mu_appdx}
\end{figure}

Shown as in Fig.~\ref{fig:cf_ba_appdx},~\ref{fig:cf_mu_appdx},
    we plot the counterfactual fidelity of \gsat and the simulated \smt with $10$ and $100$ sampling rounds on BA-2Motifs and Mutag datasets.
    The \smt is approximated via \ourss with different sampling rounds.
    It can be found that \gsat achieves a counterfactual fidelity that is $2$ to $3$ times lower than the simulated \smt via \ourss with $10$ and $100$ sampling rounds.
    Moreover, in simple tasks such as BA-2Motifs and Mutag, using larger sampling rounds like $100$ does not necessarily bring more counterfactual fidelity. One reason can be using small sampling rounds to touch the upper bounds of counterfactual fidelity measured in our work.
    We also provide a discussion on why the counterfactual fidelity grows slowly at the initial epochs in BA-2Motif datasets in Appendix~\ref{sec:gmt_impl_appdx}.
    More counterfactual fidelty studies can be found in Appendix~\ref{sec:cf_viz_appdx}.

\section{Theories and Proofs}
\label{sec:proof_appdx}

\subsection{Useful definitions}
We give the relevant definitions here for ease of reference when reading our proofs.
\begin{definition}[Subgraph multilinear extension (\smt)]\label{def:sub_mt_appdx}
    Given the attention $\boldsymbol{\alpha}\in\R^m_+$ as edge sampling probability of $G_c$, \xgnns factorize $P(G)$ as independent Bernoulli distributions on edges:
    \[P(G_c|G)=\prod_{e\in G_c}\alpha_e\prod_{e\in G/ G_c}(1-\alpha_e),\]
    which elicits the \textit{multilinear extension} of $f_c(G_c)$ in Eq.~\ref{eq:GCE} as:
    \begin{equation}\label{eq:smt_appdx}
        \begin{aligned}
            F_c(\boldsymbol{\alpha}; G) :=\sum_{G_c\in G}f_c(G_c)\prod_{e\in G_c}\alpha_e\prod_{e\in G/G_c}(1-\alpha_e) =\E_{G_c\stackrel{g}{\sim}G}f_c(G_c).
        \end{aligned}
    \end{equation}
\end{definition}

\begin{definition}[$\epsilon$-\smt approximation]\label{def:submt_approx_appdx}
    Let $d(\cdot,\cdot)$ be a distribution distance metric, a \xgnn $f=f_c\circ g$ $\epsilon$-approximates \smt (Def.~\ref{def:sub_mt}), if there exists $\epsilon\in\R_+$ such that $d(P_f(Y|G),P(Y|G))\leq\epsilon$
    where $P(Y|G)\in\R^{|\gY|}$ is the ground truth conditional label distribution, and $P_f(Y|G)\in\R^{|\gY|}$ is the predicted label distribution for $G$ via a \xgnn $f$, i.e., $P_f(Y|G)=f_c(\E_{G_c\stackrel{g}{\sim}G}G_c)$.
\end{definition}

\begin{definition}[$(\delta,\epsilon)$-counterfactual fidelity]\label{def:counterfactual_x_appdx}
    Given a meaningful minimal distance $\delta>0$,
    let $d(\cdot,\cdot)$ be a distribution distance metric ,
    if a \xgnn $f=f_c\circ g$ commits to the $\epsilon-$counterfactual fidelity, then there exist $\epsilon>0$ such that, $\forall G, \widetilde{G}$ that $d(P(Y|G),P(Y|\widetilde{G}))\geq \delta$, the following holds:
    \[d(P_f(Y|\widetilde{G}),P_f(Y|G))\geq \epsilon\delta.\]
\end{definition}

\subsection{Proof for Proposition~\ref{thm:submt_gap}}
\label{proof:submt_gap}
\begin{proposition}\label{thm:submt_gap_appdx}
    Consider a linearized GNN~\citep{sgnn} with number of message passing layers $k>1$, linear activations and pooling,
        \begin{equation}\label{eq:linear_gnn_appdx}
            f_c(G_c)=\rho(\widehat{A}^kX \mW),
        \end{equation}
        if there exists $1\leq i,j\leq n$ that $0<\widehat{A}_{i,j}<1$,
        Eq.~\ref{eq:exp_issue} can not hold, thus Eq.~\ref{eq:linear_gnn_appdx}
        can not  approximate \smt (Def.~\ref{def:sub_mt}).
\end{proposition}
\begin{proof}
    To begin with, given a linear pooling function $\rho$, one could write the outcomes of $f_c(A)=\rho(A^kXW)$ as a summation in $A^k_{i,j}v_{i,j}$, with $v_{i,j}$ is the weight that accounting for the pooling as well as $XW$:
    \begin{align}
        f_c(A)=\sum_{i}\sum{j}A_{i,j}v_{i,j}.
    \end{align}
    Given the linearity of expectations, the comparison between $E[f_c(A)]$ and $f_c(E[A])$ now turns into the comparison between $E[A^k_{i,j}v_j]$ and $(E[A_{i,j}])^kv_j$.
    Since $A_ij$ is drawn from the Bernoulli distribution, with the expectation as $\widehat{A}_{i,j}$, it suffices to know that \begin{align}
        E[A^k_{i,j}v_j]=1^k\widehat{A}_{i,j}+0^k(1-\widehat{A}_{i,j})=\widehat{A}_{i,j},
    \end{align}
    while $(E[A_{i,j}])^k=\widehat{A}_{i,j}^k$. Then, we know that $E[f_c(A)] \neq f_c(E[A])$.
\end{proof}
We also conduct empirical verifications with \gsat implemented in GIN and SGC with various layers in Appendix~\ref{sec:smt_gap_viz_appdx}.

\subsection{Proof for Proposition~\ref{thm:smt_fidelty}}
\label{proof:smt_fidelty}
\begin{proposition}\label{thm:smt_fidelty_appdx}
    If a \xgnn $f$ $\epsilon$-approximates \smt (Def.~\ref{def:submt_approx_appdx}),  then $f$ also satisfies $(\delta,1\!-\!\frac{2\epsilon}{\delta})$-counterfactual fidelity (Def.~\ref{def:counterfactual_x_appdx}).
\end{proposition}
\begin{proof}
    Considering any two graphs $G$ and $\widetilde{G}$ that $d(P(Y|G),P(Y|\widetilde{G})\geq \delta$,
    since $d$ is a distance metric, we have the following inequality holds:
    \begin{align}
        d(P(Y|G),P_f(Y|\widetilde{G})) \leq d(P_f(Y|G),P(Y|G))+d(P_f(Y|G),P_f(Y|\widetilde{G})),
    \end{align}
    by the triangle inequality.
    Furthermore, we have
    \begin{equation}
        \begin{aligned}
            d(P(Y|G),P_f(Y|\widetilde{G}))-d(P_f(Y|G),P(Y|G)) & \leq d(P_f(Y|G),P_f(Y|\widetilde{G}))
        \end{aligned}
    \end{equation}
    As \xgnn $f$ that $\epsilon$-approximates \smt,
    we have the following by definition:
    \[ d(P_f(Y|\widetilde{G}),P(Y|\widetilde{G}))\leq \epsilon, d(P_f(Y|G),P(Y|G))\leq \epsilon.\]
    Then, call the triangle inequality again, we have
    \begin{equation}
        \begin{aligned}
            d(P(Y|G),P(Y|\widetilde{G}))                                            & \leq d(P_f(Y|\widetilde{G}),P(Y|G))+d(P_f(Y|\widetilde{G}),P(Y|\widetilde{G})) \\
            d(P(Y|G),P(Y|\widetilde{G}))-d(P_f(Y|\widetilde{G}),P(Y|\widetilde{G})) & \leq d(P_f(Y|\widetilde{G}),P(Y|G))                                            \\
            \delta-d(P_f(Y|\widetilde{G}),P(Y|\widetilde{G}))                       & \leq d(P_f(Y|\widetilde{G}),P(Y|G))                                            \\
            \delta-\epsilon                                                         & \leq d(P_f(Y|\widetilde{G}),P(Y|G)).                                           \\
        \end{aligned}
    \end{equation}
    Combining the aforementioned three inequalities, we have
    \[d(P_f(Y|\widetilde{G}),P(Y|G))-d(P_f(Y|G),P(Y|G))\geq \delta-2\epsilon,\]
    Then, it suffices to know that
    \begin{equation}
        \begin{aligned}
            \delta-2\epsilon & \leq d(P_f(Y|G),P_f(Y|\widetilde{G})).
        \end{aligned}
    \end{equation}
\end{proof}

\subsection{Proof for Theorem~\ref{thm:gmt_success}}
\label{proof:gmt_success}
\begin{theorem}\label{thm:gmt_success_appdx}
    Given the attention matrix $\widehat{A}$,
    and the distribution distance metric $d$ as the total variation distance,
    let $C=|\gY|$,
    for a \ourss with $t$ i.i.d. samples of
    $G_c^i\sim P(G_c|G)$, then,
    there exists $\epsilon\in\R_+$ such that,
    with a probability at least $1-e^{-t\epsilon^2/4}$, \ourss $\frac{\epsilon C}{2}$-approximates \smt (Def.~\ref{def:submt_approx_appdx}) and satisfies $(\delta,1-\frac{\epsilon C}{\delta})$ counterfactual fidelity (Def.~\ref{def:counterfactual_x_appdx}).
\end{theorem}
\begin{proof}
    Recall the \smt objective:
    \[F_c(\boldsymbol{\alpha}; G) :=\sum_{G_c\in G}f_c(G_c)\prod_{e\in G_c}\alpha_e\prod_{e\in G/G_c}(1-\alpha_e),\]
    which is the expanded form of  $\E[f_c(G_c)]$, $G_c\sim P(G_c|\widehat{A})$.
    Then, denote $M=\max|f_c(G_c)|$, $f_c(G_c)$ can be considered as a random variable within the range of $[-M,M]$.
    Considering $t$ random i.i.d. examples of $\{G_c^i\}_{i=1}^t$ drawn from $P(G_c|\widehat{A})$,
    and the predicted probability for each class, denoted as $Y_i=\frac{1}{M}f_c(G_c^i)$,
    we then have $Y_i\in[-1,1]$ and $\sum_{i=1}^t\E[Y_i]=\frac{t}{M}F(\boldsymbol{\alpha};G)$.
    It allows us to adopt the Markov's inequality and obtain the following Chernoff bound:
    \[
        \text{Pr}(|\sum_{i=1}^tY_i-\frac{t}{M}F(\boldsymbol{\alpha};G)>t\epsilon|)<e^{-t^2\epsilon^2/4t}=e^{-t\epsilon^2/4}.
    \]
    Since by definition of \ourss, i.e.,
    \[f_c^s(\widehat{G}_c)=\frac{1}{t}\sum_{i=1}^tf_c(Y|G_c^i),\]
    we have \[\sum_{i=1}^tY_i=\frac{t}{M}\sum_{i=1}^tf_c(G_c^i)=\frac{t}{M}f_c^s(\widehat{G}_c),\]
    the bound can be written as:
    \begin{equation}
        \begin{aligned}
            \text{Pr}(|\frac{t}{M}f_c^s(\widehat{G}_c)-\frac{t}{M}F(\boldsymbol{\alpha};G)>t\epsilon|) & <e^{-t^2\epsilon^2/4t}=e^{-t\epsilon^2/4} \\
            \text{Pr}(|f_c^s(\widehat{G}_c)-F(\boldsymbol{\alpha};G)>\epsilon M|)                      & <e^{-t\epsilon^2/4}                       \\
            \text{Pr}(|f_c^s(\widehat{G}_c)-F(\boldsymbol{\alpha};G)\leq\epsilon M|)                   & \geq 1-e^{-t\epsilon^2/4}.
        \end{aligned}
    \end{equation}
    In other words, with a probability at least $1-e^{-t\epsilon^2/4}$, we have the following holds:
    \begin{equation}
        |f_c^s(\widehat{G}_c)-F_c(\boldsymbol{\alpha}; G)]|\leq \epsilon M.
    \end{equation}
    Since $M$ is defined as the maximal probability for each class,
    \[M= \max\E[f_cP(Y|G_c)],\]
    it suffices to know that $M\leq 1$.
    Therefore, it follows that
    \[|f_c^s(\widehat{G}_c)-F_c(\boldsymbol{\alpha}; G)]|\leq \epsilon,\]
    for each class, which further implies that
    \[|f_c^s(\widehat{G}_c)-F_c(\boldsymbol{\alpha}; G)]|\leq \epsilon |\gY|=\epsilon C,\]
    which commits to the $\frac{\epsilon C}{2}$ \smt approximation under the total variation distance.
    Then, using the results of Proposition~\ref{thm:smt_fidelty}, we know \ourss also commits to the $1-\frac{\epsilon C}{\delta}$ counterfactual fidelity.
\end{proof}

%% file: figures/full_scm.tex
\begin{figure*}[ht]
	\centering\hfill
	\subfigure[Graph generation SCM]{\label{fig:graph_gen_appdx}
		\resizebox{!}{0.225\textwidth}{\tikz{
				\node[latent] (S) {$S$};%
				\node[latent,left=of S,xshift=-1.5cm] (C) {$C$};%
				\node[latent,below=of C,xshift=-0.75cm,yshift=0.5cm] (ZCA) {$Z_X^c$}; %
				\node[latent,below=of C,xshift=0.75cm,yshift=0.5cm] (ZCX) {$Z_A^c$}; %
				\node[latent,below=of S,xshift=-0.75cm,yshift=0.5cm] (ZSA) {$Z_X^s$}; %
				\node[latent,below=of S,xshift=0.75cm,yshift=0.5cm] (ZSX) {$Z_A^s$}; %
				\node[latent,below=of ZCX,xshift=-0.75cm,yshift=0.5cm] (GC) {$G_c$}; %
				\node[latent,below=of ZSX,xshift=-0.75cm,yshift=0.5cm] (GS) {$G_s$}; %
				\node[obs,below=of GC,xshift=1.6cm,yshift=0.5cm] (G) {$G$}; %
				\edge[dashed,-] {C} {S}
				\edge {C} {ZCX,ZCA}
				\edge {S} {ZSX,ZSA}
				\edge {ZCX,ZCA} {GC}
				\edge {ZSX,ZSA} {GS}
				\edge {GC,GS} {G}
			}}}
	\subfigure[FIIF SCM]{\label{fig:scm_fiif_appdx}
		\resizebox{!}{0.18\textwidth}{\tikz{
				\node[latent] (E) {$E$};%
				\node[latent,below=of E,yshift=0.5cm] (S) {$S$}; %
				\node[obs,below=of E,xshift=-1.2cm,yshift=0.5cm] (Y) {$Y$}; %
				\node[obs,below=of E,xshift=1.2cm,yshift=0.5cm] (G) {$G$}; %
				\node[latent,below=of Y,xshift=1.2cm,yshift=0.5cm] (C) {$C$}; %
				\edge {E} {S}
				\edge {C} {Y,G}
				\edge {S} {G}
				\edge {C} {S}
			}}}
	\subfigure[PIIF SCM]{\label{fig:scm_piif_appdx}
		\resizebox{!}{0.18\textwidth}{\tikz{
				\node[latent] (E) {$E$};%
				\node[latent,below=of E,yshift=0.5cm] (S) {$S$}; %
				\node[obs,below=of E,xshift=-1.2cm,yshift=0.5cm] (Y) {$Y$}; %
				\node[obs,below=of E,xshift=1.2cm,yshift=0.5cm] (G) {$G$}; %
				\node[latent,below=of Y,xshift=1.2cm,yshift=0.5cm] (C) {$C$}; %
				\edge {E} {S}
				\edge {C} {Y,G}
				\edge {S} {G}
				\edge {Y} {S}
			}}}
	\subfigure[MIIF SCM]{\label{fig:scm_miif_appdx}
		\resizebox{!}{0.24\textwidth}{\tikz{
				\node[latent] (E) {$E$};%
				\node[latent,below=of E,xshift=-2cm] (S1) {$S_1$}; %
				\node[latent,below=of E,xshift=-0.6cm] (C) {$C$}; %
				\node[latent,below=of E,xshift=2cm] (S2) {$S_2$}; %
				\node[obs,below=of E,xshift=0.6cm] (Y) {$Y$}; %
				\node[obs,below=of C,xshift=0.6cm] (G) {$G$}; %
				\edge {E} {S1,S2}
				\edge {C} {S1,Y,G}
				\edge {Y} {S2}
				\edge {S1,S2} {G}
			}}}
	\caption{Full SCMs on Graph Distribution Shifts~\citep{ciga}.}
	\label{fig:scm_appdx}
\end{figure*}

%% file: sections/10_appdx_exp.tex
\clearpage
\section{More Discussions on Practical Implementations of \ourst}
\label{sec:gmt_impl_dis_appdx}
We provide more discussion in complementary to the description of Sec.~\ref{sec:gmt_sol} in the main text.

\subsection{Algorithms of \ourst}
\label{sec:gmt_alg_appdx}
\paragraph{Training subgraph extractor with random subgraph sampling.}
We focus on discussing the implementation details of \ourss since \oursl differs from \gsat only in the number of weighted message passing times.
\ourss contains two stages: i) subgraph extractor training, and ii) neural subgraph extension learning. The first stage aims to train the subgraph extractor to extract the desired subgraphs, while the second stage aims to reduce the additional computation overhead of the random subgraph sampling, and further better learn the correlations between the soft subgraphs and the labels. The algorithm for stage i) is given in Algorithm~\ref{alg:gmt_pretraining} and for stage ii) is given in Algorithm~\ref{alg:gmt_finetune}, respectively.

\begin{algorithm}[H]
    \caption{Subgraph extractor training algorithm of \oursfull (\textbf{\ourst}). }
    \label{alg:gmt_pretraining}
    \begin{algorithmic}[1]
        \STATE \textbf{Input:} Training data $\train$;
        a \xgnn $f$ with subgraph extractor $g$, and classifier $f_c$;
        subgraph sampling epochs $e_s$;
        length of maximum subgraph learning epochs $e_l$;
        batch size $b$;
        loss function $l(\cdot)$;
        subgraph regularization $o(\cdot)$;
        subgraph regularization weight $\gamma$;
        \STATE Randomly initialize $f$;
        \STATE \texttt{// Stage I:  subgraph learning.}
        \FOR{$j=1$ to $e_l$}
        \STATE Sample a batch of data $\{G^i,Y^i\}_{i=1}^b$ from $\train$;
        \STATE Obtain sampling attention $\{\boldsymbol{\alpha}^i\}_{i=1}^b$ via Eq.~\ref{eq:subgraph_att};
        \STATE \texttt{// MCMC subgraph sampling.}
        \FOR{$k=1$ to $e_s$}
        \STATE Obtain the sampling probability $\{\boldsymbol{\beta}^i\}_{i=1}^b$ via Eq.~\ref{eq:gumbel_appdx} using Gumbel-softmax;
        \STATE Randomly sample subgraphs $\{G_c^i\sim\text{Ber}(\boldsymbol{\beta}^i)\}_{i=1}^b$ via Eq.~\ref{eq:subgraph_sampling_appdx};
        \STATE Obtain predictions as logits $\{\hat{y}^i_k\}_{i=1}^b$;
        \ENDFOR
        \STATE Obtain simulated prediction $\{\hat{y}^i=\frac{1}{e_s}\sum_{k=1}^{e_s}\hat{y}_k^i\}_{i=1}^b$;
        \STATE Obtain prediction loss $l_p$ with $l(\cdot)$ and $\{\hat{y}^i\}_{i=1}^b$;
        \STATE Obtain subgraph regularization loss $l_o$ with $o(\cdot)$ and  $\{\boldsymbol{\alpha}^i\}_{i=1}^b$;
        \STATE Obtain the final loss $l_f = l_p+\eta \cdot l_o$;
        \STATE Updated model via backpropagation with $l_f$;
        \ENDFOR
        \STATE \textbf{Return} trained subgraph extraction model $f_c\circ g$;
    \end{algorithmic}
\end{algorithm}

For each input graph along with the label $(G,Y)$, the subgraph extractor $g$ first propagates among $G$ and obtains the node representations $H_i\in\R^h$ for each node.
Then, the (edge-centric) sampling attention is obtained as the following
\begin{equation}\label{eq:subgraph_att}
    \alpha_e=a([H_u,H_v]),
\end{equation}
for each edge $e=(u,v)\in E$, where $a(\cdot)$ is the attention function and can be simply implemented as a MLP. Note that $\alpha_e$ is slightly different from that in the main text, since we will discuss in detail the discrete sampling process in the implementation.

To enable the gradient backpropagation along with the discrete sampling of subgraphs, we will adopt the Gumbel-softmax trick and straight-through estimator~\citep{gumbel,gumbel2}. With the attention from Eq.~\ref{eq:subgraph_att}, the sampling probability $\boldsymbol{\beta}$ is then obtained as follows
\begin{equation}\label{eq:gumbel_appdx}
    \beta_e=\sigma((\alpha_e+D)/\tau),
\end{equation}
where $\tau$ is the temperature, $\sigma$ is the sigmoid function, and
\[
    D=\log U-\log(1-U),
\]
with $U\sim\text{Uniform}(0,1)$. To sample the discrete subgraph, we sample from the Bernoulli distributions on edges independently
\[A_e\sim \text{Bern}(\beta_e)\]
and obtain the discrete subgraph with each entry as
\begin{equation}\label{eq:subgraph_sampling_appdx}
    A_e = \text{StopGrad}(A_e-\alpha_e)+\alpha_e,
\end{equation}
which allows computing the gradients along with the subgraph sampling probability. Although the trick works empirically well, the estimated gradients are approximated ones that have biases from the ground truth.
It might be of independent interest to analyze whether the random subgraph sampling in \ourss can also reduce the gradient estimator biases during discrete sampling.

\begin{algorithm}[H]
    \caption{Subgraph classifier training algorithm of \oursfull (\textbf{\ourst}). }
    \label{alg:gmt_finetune}
    \begin{algorithmic}[1]
        \STATE \textbf{Input:} Training data $\train$;
        trained \xgnn $f$ with subgraph extractor $g$, and classifier $f_c$ by Alg.~\ref{alg:gmt_pretraining};
        length of maximum subgraph classifier training epochs $e_l$;
        batch size $b$;
        loss function $l(\cdot)$;
        subgraph regularization $o(\cdot)$;
        subgraph regularization weight $\gamma$;
        \STATE Initialize $f_c$;  Keep $g$ frozen;
        \STATE \texttt{// Stage II:  subgraph classifier learning.}
        \FOR{$j=1$ to $e_l$}
        \STATE Sample a batch of data $\{G^i,Y^i\}_{i=1}^b$ from $\train$;
        \STATE Obtain sampling attention $\{\boldsymbol{\alpha}^i\}_{i=1}^b$ via Eq.~\ref{eq:subgraph_att};
        \STATE \texttt{// Soft subgraph propagation.}
        \STATE Obtain edge sampling probability $\{\boldsymbol{\beta}^i=\text{StopGrad}(\boldsymbol{\alpha}^i)\}_{i=1}^b$; \texttt{//  subgraph extractor frozen}
        \STATE Obtain prediction with subgraph $\{\hat{y}^i\}_{i=1}^b$ via weighted message passing with $\{\boldsymbol{\beta}^i\}_{i=1}^b$;
        \STATE Obtain prediction loss $l_p$ with $l(\cdot)$ and $\{\hat{y}^i\}_{i=1}^b$;
        \STATE Obtain final loss $l_f = l_p$;
        \STATE Updated model via backpropagation with $l_f$;
        \ENDFOR
        \STATE \textbf{Return} final model $f_c\circ g$;
    \end{algorithmic}
\end{algorithm}

\paragraph{Learning neural subgraph multilinear extension.}
When the subgraph extractor is trained, we then enter into stage two, which focuses on extracting the learned subgraph information for better predicting the label with a single pass forward.
More concretely, although \ours trained with \ourss improves interpretability, \ourss still requires multiple random subgraph sampling to approximate \smt and costs much additional overhead.
To this end, we propose to learn a neural \smt that only requires a single sampling, based on the trained subgraph extractor $g$ by \ourss.

Learning the neural \smt is essentially to approximate the MCMC with a neural network, though it is inherently challenging to approximate MCMC~\citep{no_free_lunch_MCMC,papamarkou2022a}.
Nevertheless, the feasibility of neural \smt learning is backed by the inherent causal subgraph assumption of~\cite{ciga}, once the causal subgraph is correctly identified, simply learning the statistical correlation between the subgraph and the label is sufficient to recover the causal relation.

Therefore, we propose to simply re-train a new classifier GNN with the frozen subgraph extractor, to distill the knowledge contained in $\widehat{G}_c$ about $Y$.
The implementation is simply to stop the gradients of the subgraph extractor, while only optimizing the classifier GNN with the predicted sampling probability. Note that it breaks the shared encoder structure of the \gsat, which could avoid potential representation conflicts for a graph encoder shared by both the subgraph extractor and the classifier.
Under this consideration, we also enable the BatchNorm~\citep{batch_norm} in the subgraph extractor to keep count of the running stats when training the new classifier.

Empirically, the weighted message passing can effectively capture the desired information from $g$ and lead to a performance boost.
This scheme also brings additional benefits over the originally trained classifier, which focuses on providing the gradient guidance for finding proper $G_c$ instead of learning all the available statistical correlations between $G_c$ and $Y$.

\subsection{Discussions on  \ourst Implementations}
\label{sec:gmt_impl_appdx}
With the overall algorithm training the subgraph extractor and the classifier, we then discuss in more detail the specific implementation choices of \ourss.

\paragraph{Transforming node-centric random subgraph sampling.}
In the task of geometric learning, the input graphs are initially represented as point clouds. The graph structures are built upon the node features and geometric knowledge. Therefore, \lri adopts the node-centric sampling and learns sampling probabilities for nodes when implementing the graph information bottleneck.
However, when sampling concrete subgraphs from a node-centric view, it will often lead to a too aggressive sampling. Otherwise, one has to increase the sampling probability $r$ of the variational distribution $Q(G_c)$ in Eq.~\ref{eq:gib_reg_appdx}.
To this end, we  transform the node-centric sampling to edge-centric sampling. Let $\alpha_i$ denote the sampling probability for node $i$, then the edge sampling probabilities can be obtained via:
\begin{align}
    \beta_e = \alpha_u \cdot \alpha_v,
\end{align}
for each edge $e=(u,v)\in E$. It thus enables the subgraph sampling from the node-centric view. Empirically, in geometric datasets, we observe a lower variance when transforming the node-centric sampling to edge-centric sampling.

\paragraph{Warmup of \ourss.}
We first explain the motivation of using warmup for \ourss.  
Although more sampling rounds can improve the approximation precision of \ourss to \smt, it would also affect the optimization of the interpretable subgraph learning, in addition to the additional unnecessary computational overhead. For example, at the beginning of the interpretable subgraph learning, the subgraph extractor will yield random probabilities like $0.5$.
\begin{itemize}
    \item First, a more accurate estimation based on random \smt is unnecessary.
    \item Second, at such random probabilities, every subgraph gets a nearly equal chance of being sampled, and gets gradients backpropagated. Since neural networks are universal approximators, the whole network can easily be misled by the noises, which will slow down the learning speed of the meaningful subgraphs.
    \item Third, when spurious correlations exist between subgraphs and the labels, the learning process will be more easily misled by the potential spurious correlations at the beginning of the subgraph learning.
\end{itemize}
More importantly, sampling multiple times can lead to trivial solutions with degenerated performance in the \gsat objective. Specifically, the formulation of the mutual information regularizer in \gsat has a trivial solution where all $\alpha_e$ directly collapses to the given $r$. More formally, let $\alpha_e=r$ in the following objective obviously lead to zero loss that appears to be a Pareto optimal solution~\citep{pair} that can be selected as the output:
\[
    D_\text{KL}(\text{Bern}(\alpha_e)||\text{Bern}(r))=\sum_{e}\alpha_e\log\frac{r}{r}+(1-r)\log\frac{(1-\alpha_e)}{(1-r)}=0.
\]
The trivial solutions can occur to \ours more easily with more rounds of subgraph sampling, especially in too simple or too complicated tasks.

To tackle the above problem, we propose two warmup strategies:
\begin{itemize}
    \item Larger initial prior $r$ of $Q(G_c)$ in Eq.~\ref{eq:gib_reg_appdx}: \gsat achieves the objective of graph information bottleneck with a schedule of $r$ in $Q(G_c)$ as $0.9$, which could promote the random sampling probabilities to meaningful subgraph signals. As the random subgraph sampling will slow the optimization, we can warm up the initial subgraph learning with a larger initial $r$. In experiments, we try with $r=1.0$ and $r=0.9$, and find $r=1.0$ can effectively warm up and speed up the subgraph learning, which is especially meaningful for too simple tasks where \xgnns can easily overfit to, or too hard tasks where \xgnns learns the meaningful subgraph signals in a quite slow speed. We can also use a larger regularization penalty at the initial stage to speed up meaningful subgraph learning.
    \item Single subgraph sampling: As sampling too many subgraphs can bring many drawbacks such as overfitting and slow learning, we propose warm up the initial subgraph learning with a single sampling during the first stage of $r$ (i.e., when $r$ still equals to the initial $r$ in the schedule of \gsat). The single subgraph sampling also implicitly promotes meaningful subgraph learning, as it encourages a higher chance even for a small difference in the sampling probability.
\end{itemize}

In addition to helping with the warmup of the interpretable subgraph, single subgraph sampling also has some additional benefits and effectively tackles the trivial solution of \gsat objective. It also brings more variance between meaningful subgraph learning and noisy subgraph learning, and we find using a single random subgraph learning is extremely helpful for simple tasks such as BA\_2motifs in our experiments. The implicit variance of single random subgraph sampling also brings additional benefits to maintaining high variance between the signal subgraph and noisy subgraph, which might be of independent interest.
It turns out that  the variance in single subgraph learning can have an implicit regularization preventing the trivial solution.

In experiments, we will use all of the warmup strategies together (i.e., a larger initial $r$, a larger penalty score, and single subgraph sampling) when we observe a performance degeneration in the validation set. Otherwise, we will stick to the original receipt. More details are given in Sec.~\ref{sec:eval_appdx}.

\paragraph{Single weighted message passing in \oursl.}
Although it has been shown that propagation with the attention only once can effectively reduce the \smt approximation error, it remains unknown which layer the attention should be applied.
Empirically, we examine the following three strategies:
\begin{itemize}
    \item Weighted message passing on the first layer;
    \item Weighted message passing on the last layer;
    \item Single weighted message passing of all layers, and then average the logits;
\end{itemize}
We find applying weighted message passing to the first layer outperforms the other two strategies in experiments, and thus we stick to the first layer weighted message passing scheme. Exploring the reasons behind the intriguing phenomenon will be an interesting future extension.

\paragraph{Subgraph sampling for neural \smt.}
Although the weighted message passing with $\boldsymbol{\alpha}$ produced by the trained subgraph extractor already achieves better performance, it may not maximally extract the full underlying information of the learned subgraph and the labels, since the original function is a MCMC that is not easy to be fitted~\citep{no_free_lunch_MCMC}. Besides, the weighted message passing itself may not be expressive enough due to the expressivity constraints of GNNs~\citep{gin}, and also the limitations of the attention-based GNNs~\citep{gatv2,aero_gnn}.

Therefore, we propose more subgraph sampling strategies along with alternative architecture of the new classifier, in order to best fit the underlying MCMC function.
Specifically, we consider the following aspects:
\begin{itemize}
    \item Initialization: the graph encoder of the new classifier can be initialized from scratch and avoids overfitting, or initialized from the random subgraph sampling trained models;
    \item Architecture: weighted message passing, or single weighted message passing as that of \oursl;
    \item Attention sampling: set the minimum $p\%$ attention scores directly to $0$; set the maximum $p\%$ attention scores directly to $1$; set the maximum $p\%$ attention scores directly to $1$ while set the minimum $(1-p)\%$ attention scores directly to $0$;
\end{itemize}
We examine the aforementioned strategies and choose the one according to the validation performance in experiments. We exhibit the detailed hyperparameter setup in Appendix~\ref{sec:eval_appdx}.

\section{More Details about the Experiments}
\label{sec:exp_appdx}
In this section, we provide more details about the experiments, including the dataset preparation, baseline implementations, models and hyperparameters selection as well as the evaluation protocols.

\bgroup
\def\arraystretch{1}
\begin{table}[ht]
    \centering
    \caption{Information about the datasets used in experiments. The number of nodes and edges are respectively taking average among all graphs.}\small\sc
    \label{tab:datasets_stats_appdx}
    \resizebox{\textwidth}{!}{
        \begin{small}
            \begin{tabular}{l|ccccccc}
                \toprule
                \textbf{Datasets}      & \textbf{\# Training} & \textbf{\# Validation} & \textbf{\# Testing} & \textbf{\# Classes} & \textbf{ \# Nodes} & \textbf{ \# Edges}
                                       & \textbf{  Metrics}                                                                                                                        \\\midrule
                BA-2Motifs             & $800$                & $100$                  & $100$               & $2$                 & $25$               & $50.96$            & ACC \\
                Mutag                  & $2,360$              & $591$                  & $1,015$             & $2$                 & $30.13$            & $60.91$            & ACC \\
                Suprious-Motif $b=0.5$ & $9,000$              & $3,000$                & $6,000$             & $3$                 & $45.05$            & $65.72$            & ACC \\
                Suprious-Motif $b=0.7$ & $9,000$              & $3,000$                & $6,000$             & $3$                 & $46.36$            & $67.10$            & ACC \\
                Suprious-Motif $b=0.9$ & $9,000$              & $3,000$                & $6,000$             & $3$                 & $46.58$            & $67.59$            & ACC \\
                MNIST-75sp             & $20,000$             & $5,000$                & $10,000$            & $10$                & $70.57$            & $590.52$           & ACC \\
                Graph-SST2             & $28,327$             & $3,147 $               & $12,305$            & $2$                 & $10.20$            & $18.40$            & ACC \\
                OGBG-MolHiv            & $32,901 $            & $4,113 $               & $4,113$             & $2$                 & $25.51$            & $54.94$            & AUC \\
                \bottomrule
            \end{tabular}	\end{small}}
\end{table}
\egroup

\begin{table}[t]
    \caption{Statistics of the four geometric datasets from~\citet{lri}.}
    \label{tab:lri_stat}
    \centering
    \resizebox{\textwidth}{!}{%
        \begin{tabular}{lccccccccc}
            \toprule
                              & \# Classes & \# Features in $\mathbf{X}$ & \# Dimensions in $\mathbf{r}$ & \# Samples & Avg. \# Points/Sample & Avg. \# Important Points/Sample & Class Ratio & Split Scheme & Split Ratio \\
            \midrule
            \texttt{\acts}    & 2          & 0                           & 3                             & 3241       & 109.1                 & 22.8                            & 39/61       & Random       & 70/15/15    \\
            \texttt{\taumu}   & 2          & 1                           & 2                             & 129687     & 16.9                  & 5.5                             & 24/76       & Random       & 70/15/15    \\
            \texttt{\synbind} & 2          & 1                           & 3                             & 8663       & 21.9                  & 6.6                             & 18/82       & Patterns     & 78/11/11    \\
            \texttt{\pdbbind} & 2          & 3                           & 3                             & 10891      & 339.8                 & 132.2                           & 29/71       & Time         & 92/6/2      \\
            \bottomrule
        \end{tabular}%
    }
\end{table}

\subsection{Datasets}
\label{sec:dataset_appdx}

We provide more details about the motivation and construction method of the datasets that are used in our experiments. Statistics of the regular graph datasets are presented in Table~\ref{tab:datasets_stats_appdx}, and statistics of the geometric graph datasets are presented in Table~\ref{tab:lri_stat}.

\textbf{BA-2Motifs}~\citep{pge} is a synthetic dataset that adopts the Barabási–Albert (BA) graph data model to generate subgraphs in specific shapes. Each graph contains a motif subgraph that is either a five-node cycle or a house. The class labels are determined by the motif, and the motif itself serves as the interpretation of ground truth. The motif is then attached to a large base graph.

\textbf{Mutag}~\citep{mutag} is a typical molecular property prediction dataset. The nodes represent atoms and the edges represent chemical bonds. The label of each graph is binary and is determined based on its mutagenic effect. Following~\citet{pge,gsat}, -NO2 and -NH2 in mutagen graphs are labeled as ground-truth explanations.

\textbf{MNIST-sp}~\citep{understand_att} is a graph dataset converted from MNIST dataset via superpixel transformation. The nodes of MNIST-75sp graphs are the superpixels and the edges are generated according to the spatial distance of nodes in the original image. The ground truth explanations of MNIST-75sp are simply the non-zero pixels. As the original digits are hand-written, the interpretation subgraphs could be in varying sizes.

\textbf{Suprious-Motif datasets}~\citep{dir} is a 3-class synthetic datasets based on BA-2Motifs~\citep{gnn_explainer,pge} with structural distribution shifts.
The model needs to tell which one of three motifs (House, Cycle, Crane) the graph contains.
For each dataset, $3000$ graphs are generated for each class at the training set, $1000$ graphs for each class at the validation set and testing set, respectively.
During the construction of the training data, the motif and one of the three base graphs (Tree, Ladder, Wheel) are artificially (spuriously) correlated with a probability of various biases, and equally correlated with the other two. Specifically, given a predefined bias $b$, the probability of a specific motif (e.g., House) and a specific base graph (Tree) will co-occur is $b$ while for the others is $(1-b)/2$ (e.g., House-Ladder, House-Wheel).
The test data does not have spurious correlations with the base graphs, however, test data will use larger base graphs that contain graph size distribution shifts.
Following~\citet{gsat}, we select datasets with a bias of $b=0.5$, $b=0.7$, and $b=0.9$.  The interpretation ground truth is therefore the motif itself.

\textbf{Graph-SST2}~\citep{sst25,xgnn_tax} is converted from a sentiment analysis dataset in texts. Each sentence in SST2 will be converted to a graph. In the converted graph, the nodes are the words and the edges are the relations between different words. Bode features are generated using BERT~\citep{bert} and the edges are parsed by a Biaffine parser~\citep{biaffine}.
Following previous works~\citep{dir,gsat,ciga}, our splits are created according to the averaged degrees of each graph.
Specifically, we assign the graphs as follows: Those that have smaller or equal to $50$-th percentile averaged degree are assigned to training, those that have averaged degree larger than $50$-th percentile while smaller than $80$-th percentile are assigned to the validation set, and the left are assigned to test set.
Since the original dataset does not have the ground truth interpretations, we report only the classification results.

\textbf{OGBG-Molhiv}~\citep{ogb} is also a molecular property prediction dataset. The nodes represent atoms and the edges represent chemical bonds. The label of each graph is binary and is determined based on whether a molecule inhibits HIV virus replication or not. The training, validation and test splits are constructed according to the scaffolds~\citep{ogb} hence there also exist distribution shifts across different splits. Since the original dataset does not have the ground truth interpretations, we report only the classification results.

In what follows we continue to introduce the four geometric learning datasets. We refer interested readers to~\citet{lri} for more details.

\textbf{ActsTrack} dataset~\citep{lri}:
\begin{itemize}[leftmargin=*]
    \item Background: \textbf{ActsTrack} involves a fundamental resource in High Energy Physics (HEP), employed for the purpose of reconstructing various properties, including the kinematics, of charged particles based on a series of positional measurements obtained from a tracking detector. Within the realm of HEP experimental data analysis, particle tracking is an essential procedure, and it also finds application in medical contexts, such as proton therapy~\citep{act}. \text{ActsTrack} is formulated differently by~\citet{lri} from traditional track reconstruction tasks: It requires predicting the existence of a z → µµ decay and using the set of points from the µ ’s to verify model interpretation, which can be used to reconstruct µ tracks.
    \item Construction: In the \textbf{ActsTrack} dataset, each data point corresponds to a detector hit left by a particle, and it is associated with a 3D coordinate. Notably, the data points in ActsTrack lack any features in the X dimension, necessitating the use of a placeholder feature with all values set to one during model training. Additionally, the dataset provides information about the momenta of particles as measured by the detectors, which has the potential to be employed for assessing fine-grained geometric patterns in the data; however, it is not utilized as part of the model training process. Given that, on average, each particle generates approximately 12 hits, and a model can perform well by capturing the trajectory of any one of the µ (muon) particles resulting from the decay, we report performance metrics in precision@12 following~\citet{lri}. The dataset was randomly split into training, validation, and test sets, maintaining a distribution ratio of 70\% for training, 15\% for validation, and 15\% for testing.
\end{itemize}

\textbf{Tau3Mu} dataset~\citep{lri}:
\begin{itemize}[leftmargin=*]
    \item Background: \textbf{Tau3Mu} involves another application in High Energy Physics (HEP) dedicated to identifying a particularly challenging signature – charged lepton flavor-violating decays, specifically $\tau\rightarrow\mu\mu\mu$ decay.
          This task involves the analysis of simulated muon detector hits resulting from proton-proton collisions. It's worth noting that such decays are heavily suppressed within the framework of the Standard Model (SM) of particle physics~\citep{tau3mu}, making their detection a strong indicator of physics phenomena beyond the Standard Model~\citep{tau3mu_discovery}.
          Unfortunately, $\tau\rightarrow\mu\mu\mu$ decay involves particles with extremely low momentum, rendering them technically impossible to trigger using conventional human-engineered algorithms. Consequently, the online detection of these decays necessitates the utilization of advanced models that explore the correlations between signal hits and background hits, particularly in the context of the Large Hadron Collider. Our specific objective is twofold: to predict the occurrence of $\tau\rightarrow\mu\mu\mu$ decay and to employ the detector hits generated by the $\mu$ (muon) particles to validate the model's interpretations.

    \item Construction: \textbf{Tau3Mu} uses the data simulated via the PYTHIA generator~\citep{Bierlich2022ACG}.
          The interpretation labels are using the signal sample with the background samples on a per-event basis (per point cloud) while preserving the ground-truth labels. The hits originating from $\mu$ (muon) particles resulting from the $\tau\rightarrow\mu\mu\mu$ decay are designated as ground-truth interpretation. The training data only include hits from the initial layer of detectors, ensuring that each sample in the dataset contains a minimum of three detector hits. Each data point in the samples comprises measurements of a local bending angle and a 2D coordinate within the pseudorapidity-azimuth ($\eta-\phi$) space.
          Given that, in the most favorable scenario, the model is required to capture hits from each $\mu$ particle, we report precision@3 following~\citet{lri}.  Lastly, the dataset is randomly split into training, validation, and test sets, maintaining a distribution ratio of 70\% for training, 15\% for validation, and 15\% for testing.

\end{itemize}

\textbf{SynMol} dataset~\citep{lri}:
\begin{itemize}[leftmargin=*]
    \item Background: \textbf{SynMol} is a molecular property prediction task. While prior research efforts have explored model interpretability within this domain~\citep{McCloskey2018UsingAT}, their emphasis has been primarily on examining chemical bond graph representations of molecules, often overlooking the consideration of geometric attributes. In our present study, we shift our attention towards 3D representations of molecules. Our specific objective is to predict a property associated with two functional groups, namely carbonyl and unbranched alkane (as defined by~\citet{McCloskey2018UsingAT}), and subsequently employ the atoms within these functional groups to validate our model's interpretations.
    \item Construction: \textbf{SynMol} is constructed based on ZINC~\citep{zin} following~\citet{McCloskey2018UsingAT} that creates synthetic properties based on the existence of certain functional groups. The labeling criteria involve classifying a molecule as a positive sample if it contains both an unbranched alkane and a carbonyl group. Conversely, molecules lacking this combination are categorized as negative samples. Consequently, the atoms within branched alkanes and carbonyl groups serve as the designated ground-truth interpretation.
          In addition to specifying a 3D coordinate, each data point within a sample is also associated with a categorical feature signifying the type of atom it represents. While the combined total of atoms in the two functional groups may be limited to just five, it is important to note that certain molecules may contain multiple instances of such functional groups. Consequently, we report precision metric at precision@5 following~\citet{lri}.
          Finally, to mitigate dataset bias, the dataset is split into training, validation, and test sets using a distribution strategy following~\citet{McCloskey2018UsingAT,lri}. This approach ensures a uniform distribution of molecules containing or lacking either of these functional groups.
\end{itemize}

\textbf{PLBind} dataset~\citep{lri}:
\begin{itemize}[leftmargin=*]
    \item Background: \textbf{PLBind} is to predict protein-ligand binding afﬁnities leveraging the 3D structural information of both proteins and ligands. This task holds paramount significance in the field of drug discovery, as a high affinity between a protein and a ligand is a critical criterion in the drug selection process~\citep{Wang2017ImprovingSP,Karimi2018DeepAffinityID}. The accurate prediction of these affinities using interpretable models serves as a valuable resource for rational drug design and contributes to a deeper comprehension of the underlying biophysical mechanisms governing protein-ligand binding~\citep{Du2016InsightsIP}. Our specific mission is to forecast whether the affinity surpasses a predefined threshold, and we achieve this by examining the amino acids situated within the binding site of the test protein to corroborate our model's interpretations.
    \item Construction: \textbf{PLBind} is constructed protein-ligand complexes from PDBind~\citep{PDBind}. PDBind annotates binding afﬁnities for a subset of complexes in the Protein Data Bank (PDB)~\citep{PDB}, therefore, a threshold on the binding afﬁnity between a pair of protein and ligand can be used to construct a binary classification task. The ground-truth interpretation is then the part of the protein that are within 15A of the ligand to be the binding site~\citep{Liu2022Generating3M}. Besides, PLBind also includes all atomic contacts (hydrogen bond and hydrophobic contact) for every protein-ligand pair from PDBsum~\citep{Laskowski2001PDBsumSS}, where the ground-truth interpretations are the corresponding amino acids in the protein.
          Every amino acid in a protein is linked to a 3D coordinate, an amino acid type designation, the solvent-accessible surface area (SASA), and the B-factor. Likewise, each atom within a ligand is associated with a 3D coordinate, an atom type classification, and Gasteiger charges. The whole dataset is then partitioned into training, validation, and test sets, adopting a division based on the year of discovery for the complexes, following~\citet{Strk2022EquiBindGD}.
\end{itemize}

\subsection{Baselines and Evaluation Setup}
\label{sec:eval_appdx}

During the experiments, we do not tune the hyperparameters exhaustively while following the common recipes for optimizing GNNs, and also the recommendation setups by previous works.
Details are as follows.

\textbf{GNN encoder.} For fair comparison, we use the same GNN architecture as graph encoders for all methods, following~\citet{gsat,lri}.
For the backbone of GIN, we use $2$-layer GIN~\citep{gin} with Batch Normalization~\citep{batch_norm} between layers, a hidden dimension of $64$ and a dropout ratio of $0.3$.
For the backbone of PNA, we use $4$-layer PNA~\citep{pna} with Batch Normalization~\citep{batch_norm} between layers, a hidden dimension of $80$ and a dropout ratio of $0.3$. The PNA network does not use scalars, while using (mean, min, max, std aggregators.
For the backbone of EGNN~\citep{egnn}, we use $4$-layer EGNN with Batch Normalization~\citep{batch_norm} between layers, a hidden dimension of $64$ and a dropout ratio of $0.2$. The pooling functions are all sum pooling.

\textbf{Dataset splits.} We follow previous works~\citep{pge,gsat} to split BA-2Motifs randomly into three sets as (80\%/10\%/10\%), Mutag randomly into 80\%/20\% as train and validation sets where the test data are the mutagen molecules with -NO2 or -NH2. We use the default split for MNIST-75sp given by~\citep{understand_att} with a smaller sampling size following~\citep{gsat}. We use the default splits for Graph-SST2~\citep{xgnn_tax}, Spurious-Motifs~\citep{dir} and OGBG-Molhiv~\citep{ogb} datasets.
For geometric datasets, we use the author provided default splits.

\textbf{Baseline implementations.} We use the author provided codes to implement the baselines \gsat~\citep{gsat}\footnote{\url{https://github.com/Graph-COM/GSAT}} and \lri~\citep{lri}\footnote{\url{https://github.com/Graph-COM/LRI}}.
We re-run \gsat and \lri under the same environment using the author-recommended hyperparameters for a fair comparison.
Specifically, BA-2Motif, Mutag and PLBind use $r = 0.5$, and all other datasets use $r = 0.7$. The $\lambda$ of information regularizer is set to be $1$ for regular graphs, $0.01$ for \taumu, and $0.1$ for \acts, \synbind and \pdbbind as recommended by the authors. $r$ will initially be set to $0.9$ and gradually decay to the tuned value. We adopt a step decay, where $r$ will decay $0.1$ for every $10$ epochs.
As for the implementation of explanation methods,
for regular graphs, we directly adopt the results reported.
For geometric graphs, we re-run the baselines to obtain the results, as previous results are obtained according to the best validation interpretation performance that may mismatch the practical scenario where the interpretation labels are usually not available.

\textbf{Optimization and model selection.}
Following previous works, by default, we use Adam optimizer~\citep{adam} with a learning rate of $1e-3$ and a batch size of $128$ for all models at all datasets, except for Spurious-Motif with GIN and PNA, Graph-SST2 with PNA that we will use a learning rate of $3e-3$.
When GIN is used as the backbone model, MNIST-75sp is trained for 200 epochs, and all other datasets are trained for 100 epochs, as we observe that 100 epochs are sufficient for convergence at OGBG-Molhiv.
When PNA is used, Mutag and Ba-2Motifs are trained for 50 epochs and all other datasets are trained for 200 epochs. We report the performance of the epoch that achieves the best validation prediction performance and use the models that achieve such best validation performance as the pre-trained models.
All datasets use a batch size of 128; except for MNIST-75sp with GIN, we use a batch size of 256 to speed up training due to its large size in the graph setting.

The final model is selected according to the best validation classification performance. We report the mean and standard deviation of $10$ runs with random seeds from $0$ to $9$.

\textbf{Implementations of \ourst.}
For a fair comparison, \ours uses the same GNN architecture for GNN encoders as the baseline methods.
We search for the hyperparameters of $r$ from $[r_0-0.1,r_0,r_0+0.1]$ according to the default $r_0$ given by~\citet{gsat,lri}. We search the weights of graph information regularizers from $[0.1,0.5,1,2]$ for regular graphs and from $[0.01,0.1,1]$ for geometric datasets.
To avoid trivial solutions of the subgraph extractor at the early stage, we search for warm-up strategies mentioned in Appendix~\ref{sec:gmt_impl_appdx}. Besides, we also search for the decay epochs of the $r$ scheduler to avoid trivial solutions.
We search for the sampling rounds from $[1,20,40,80,100,200]$ when the memory allows.
In experiments, we find \ours already achieves the state-of-the-art results in most of the set-ups without the warm-up. Only in BA-2Motifs and MNIST-75sp with GIN, and in Tau3Mu with EGNN, \ours needs the warmups.

\begin{table}[H]
    \caption{Sensitivity to different subgraph decoding strategies.}
    \label{tab:decoding_strategy}
    \resizebox{\textwidth}{!}{
        \begin{tabular}{@{}lll|llllll@{}}
            \toprule
                           &              &           & Generalization             &                            &                            & Interpretation             &                            &                            \\
            Initialization & Architecture & Attention & spmotif-0.5                & spmotif-0.7                & spmotif-0.9                & spmotif-0.5                & spmotif-0.7                & spmotif-0.9                \\\midrule
                           &              & \gsat     & $47.45$\std{5.87}          & $43.57$\std{3.05}          & $45.39$\std{5.02}          & $74.49$\std{4.46}          & $72.95$\std{6.40}          & $65.25$\std{4.42}          \\
            new            & mul          & min0      & $\mathbf{60.09}$\std{5.57} & $54.34$\std{4.04}          & $\mathbf{55.83}$\std{5.68} & $85.50$\std{2.40}          & $\mathbf{84.67}$\std{2.38} & $73.49$\std{5.33}          \\
            old            & mul          & min0      & $58.83$\std{7.22}          & $\mathbf{55.04}$\std{4.73} & $55.77$\std{5.97}          & $\mathbf{85.52}$\std{2.41} & $84.65$\std{2.42}          & $73.49$\std{5.33}          \\
            new            & mul          & max1      & $44.49$\std{2.65}          & $49.77$\std{2.31}          & $50.22$\std{2.79}          & $85.50$\std{2.39}          & $84.66$\std{2.37}          & $73.50$\std{5.31}          \\
            old            & mul          & max1      & $45.91$\std{2.86}          & $49.11$\std{3.04}          & $50.30$\std{2.07}          & $85.49$\std{2.39}          & $84.64$\std{2.39}          & $73.50$\std{5.35}          \\
            old            & mul          & min0max1  & $51.21$\std{6.46}          & $50.91$\std{6.50}          & $53.13$\std{4.46}          & $\mathbf{85.52}$\std{2.41} & $84.66$\std{2.43}          & $73.49$\std{5.34}          \\
            new            & mul          & normal    & $47.69$\std{5.72}          & $44.12$\std{5.44}          & $40.69$\std{4.84}          & $84.69$\std{2.40}          & $80.08$\std{5.37}          & $73.48$\std{5.34}          \\
            old            & mul          & normal    & $45.36$\std{2.65}          & $44.25$\std{5.41}          & $43.43$\std{5.44}          & $83.52$\std{3.41}          & $80.07$\std{5.35}          & $73.49$\std{5.36}          \\
            new            & lin          & normal    & $43.54$\std{5.02}          & $47.59$\std{4.78}          & $46.53$\std{3.27}          & $85.47$\std{2.39}          & $80.07$\std{5.37}          & $\mathbf{73.52}$\std{5.34} \\
            old            & lin          & normal    & $46.18$\std{3.03}          & $46.42$\std{5.63}          & $49.00$\std{3.34}          & $83.51$\std{3.39}          & $80.09$\std{5.34}          & $73.46$\std{5.35}          \\\bottomrule
        \end{tabular}}
\end{table}
To better extract the subgraph information, we also search for subgraph sampling strategies mentioned in Appendix~\ref{sec:gmt_impl_appdx}.
Note that the hyperparameter search and training of the classifier is independent of the hyperparameter search of the subgraph extractor. One could select the best subgraph extractor and train the new classifier onto it. When training the classifier, we search for the following $9$ subgraph decoding strategies as shown in Table~\ref{tab:decoding_strategy}.
Specifically,
\begin{itemize}[leftmargin=*]
    \item Initialization: "new" refers to that  the classifier is initialized from scratch; "old" refers to that the classifier is initialized from the subgraph extractor;
    \item Architecture: "mul" refers to the default message passing architecture; "lin" refers to the \oursl architecture;
    \item Attention: "normal" refers to the default weighted message passing scheme; "min0" refers to setting the minimum $p\%$ attention scores directly to $0$; "max0" refers to setting the maximum $p\%$ attention scores directly to $1$; "min0max1" refers to setting the maximum $p\%$ attention scores directly to $1$ while set the minimum $(1-p)\%$ attention scores directly to $0$;
\end{itemize}
Table~\ref{tab:decoding_strategy} demonstrates the generalization and interpretation performance of \ourss in spurious motif datasets~\citep{dir}, denoted as "spmotif" with different levels of spurious correlations. It can be found that \ourss is generically robust to the different choices of the decoding scheme and leads to improvements in terms of OOD generalizability and interpretability.

\subsection{More interpretation results}
\label{sec:more_x_appdx}
We additionally conduct experiments with post-hoc explanation methods based on the PNA backbone.
Specifically, we selected two representative post-hoc methods GNNExplainer and PGExplainer, and one representative intrinsic interpretable baseline DIR.
The results are given in the table below. It can be found that most of the baselines still significantly underperform GSAT and GMT.
One exception is that DIR obtains highly competitive (though unstable) interpretation results in spurious motif datasets,
nevertheless, the generalization performance of DIR remains highly degenerated ($53.03$\std{8.05} on spmotif\_0.9).

\begin{table}[ht]\caption{More interpretation results of baselines using PNA}\centering\small
    \begin{tabular}{@{}lllllll@{}}
        \toprule
               & BA\_2Motifs      & Mutag            & MNIST-75sp       & spmotif\_0.5    & spmotif\_0.7    & spmotif\_0.9    \\\midrule
        GNNExp & 54.14\std{3.30}  & 73.10\std{7.44}  & 53.91\std{2.67}  & 59.40\std{3.88} & 56.20\std{6.30} & 57.39\std{5.95} \\
        PGE    & 48.80\std{14.58} & 76.02\std{7.37}  & 56.61\std{3.38}  & 59.46\std{1.57} & 59.65\std{1.19} & 60.57\std{0.85} \\
        DIR    & 72.33\std{23.87} & 87.57\std{27.87} & 43.12\std{10.07} & 85.90\std{2.24} & 83.13\std{4.26} & 85.10\std{4.15} \\
        GSAT   & 89.35\std{5.41}  & 99.00\std{0.37}  & 85.72\std{1.10}  & 79.84\std{3.21} & 79.76\std{3.66} & 80.70\std{5.45} \\
        \oursl & 95.79\std{7.30}  & 99.58\std{0.17}  & 85.02\std{1.03}  & 80.19\std{2.22} & 84.74\std{1.82} & 85.08\std{3.85} \\
        \ourss & 99.60\std{0.48}  & 99.89\std{0.05}  & 87.34\std{1.79}  & 88.27\std{1.71} & 86.58\std{1.89} & 85.26\std{1.92} \\\bottomrule
    \end{tabular}
\end{table}

\subsection{Computational analysis}
\label{sec:comp_appdx}
We provide more discussion and analysis about the computational overhead required by \ours, when compared to \gsat.
As \oursl differs only in the number of weighted message passing rounds from \gsat, and has the same number of total message passing rounds, hence \oursl and \gsat have the same time complexity as $O(E)$ for each epoch, or for inference.
When comparing \ourss to \oursl and \gsat,
During training, \ourss needs to process $k$ rounds of random subgraph sampling, resulting in $O(k|E|)$ time complexity;
During inference, \ourss with normal subgraph decoding methods requires the same complexity as \oursl and \gsat, as $O(|E|)$. When with special decoding strategy such as setting part of the attention entries to $1$ or $0$,
\ourss additionally needs to sort the attention weights, and requires $O(|E|+|E|\log |E|)$ time complexity.

\begin{table}[ht]
    \centering\small
    \begin{tabular}{@{}llllll@{}}
        \toprule
                        & BA\_2Motifs    &                 & MNIST-75sp       &                  & ActsTrack      \\
        Training        & GIN            & PNA             & GIN              & PNA              & EGNN           \\  \midrule
        \gsat           & 0.70\std{0.12} & 1.00\std{0.13}  & 41.28\std{0.61}  & 80.98\std{10.5}5 & 3.57\std{1.41} \\
        \oursl          & 0.68\std{0.12} & 1.02\std{0.15}  & 41.12\std{0.69}  & 81.11\std{10.4}4 & 3.69\std{0.93} \\
        \ourss          & 6.25\std{0.48} & 17.03\std{0.91} & 136.60\std{1.21} & 280.77\std{4.00} & 5.38\std{0.59} \\ \midrule
        Inference       &                &                 &                  &                  &                \\ \midrule
        \gsat           & 0.07\std{0.05} & 0.11\std{0.12}  & 18.69\std{0.35}  & 24.40\std{2.06}  & 0.84\std{0.38} \\
        \oursl          & 0.08\std{0.07} & 0.07\std{0.01}  & 18.72\std{0.41}  & 23.81\std{1.89}  & 0.80\std{0.21} \\
        \ourss (normal) & 0.05\std{0.01} & 0.12\std{0.01}  & 18.72\std{0.35}  & 18.01\std{1.47}  & 0.50\std{0.13} \\
        \ourss (sort)   & 0.07\std{0.01} & 0.21\std{0.06}  & 19.07\std{0.55}  & 18.69\std{3.35}  & 0.54\std{0.10} \\ \bottomrule
    \end{tabular}
\end{table}

In the table above, we benchmarked the real training/inference time of \gsat, \oursl and \ourss in different datasets,
where each entry demonstrates the time in seconds for one epoch.
We benchmark the latency of \gsat, \oursl and \ourss based on GIN, PNA and EGNN on different scales of datasets.
The sampling rounds of \ourss are set to $20$ for PNA on MNIST-sp, $10$ for EGNN, and $100$ to other setups.
From the table, it can be found that, although \ourss takes longer time for training, but the absolute values are not high even for the largest dataset MNIST-sp.
As for inference, \ourss enjoys a similar latency as others, aligned with our discussion.

\subsection{More counterfactual fidelity studies}
\label{sec:cf_viz_appdx}

To better understand the results, we provide more counterfactual fidelity results in supplementary to Sec.~\ref{sec:expressivity_issue} and Fig.~\ref{fig:cf_ba_appdx} and~\ref{fig:cf_mu_appdx}.

\begin{figure}[H]
    \centering
    \subfigure[\smt on BA-2Motifs trainset.]{
        \includegraphics[width=0.31\textwidth]{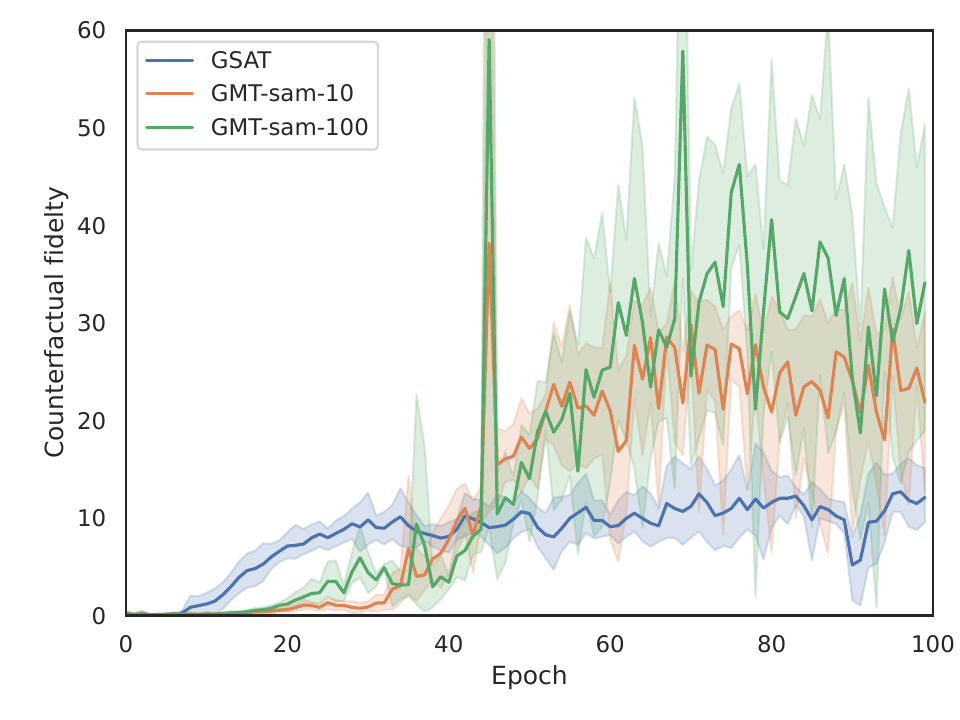}
    }
    \subfigure[\smt on BA-2Motifs valset.]{
        \includegraphics[width=0.31\textwidth]{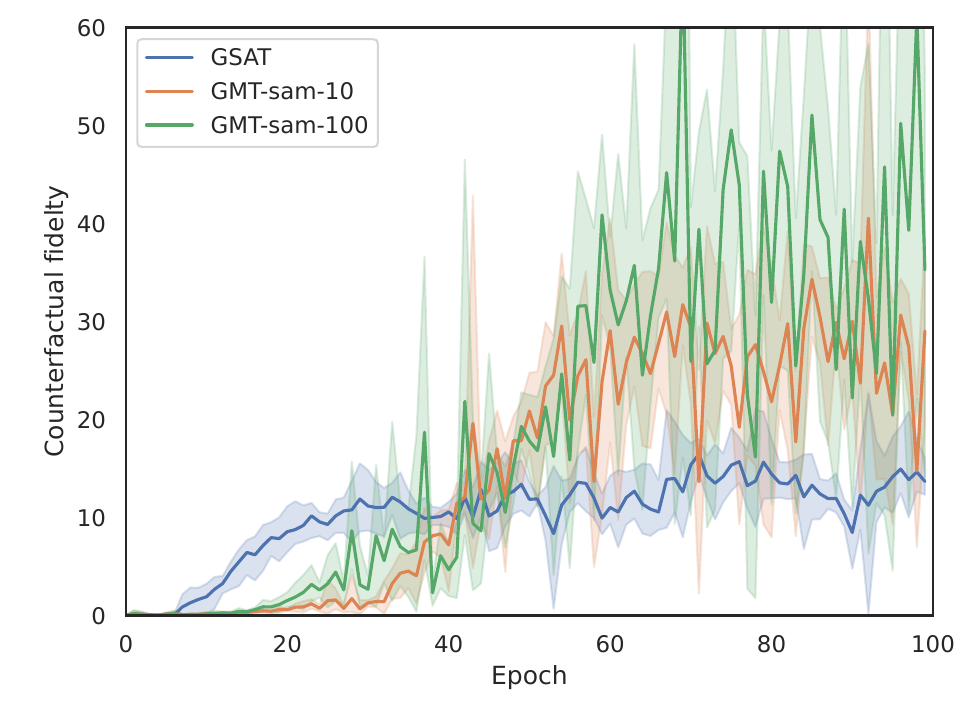}
    }
    \subfigure[\smt on BA-2Motifs test set.]{
        \includegraphics[width=0.31\textwidth]{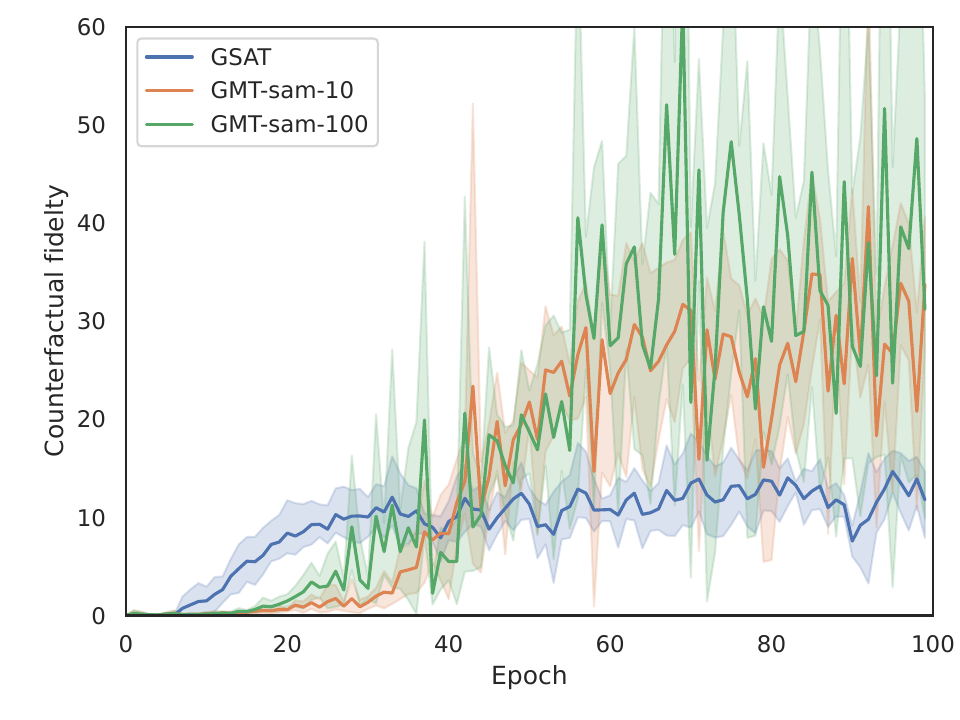}
    }
    \caption{Comparison of \gsat and the simulated \smt in counterfactual fidelity on BA-2Motifs measured via KL divergence.}
    \label{fig:cf_ba_kl_appdx}
\end{figure}

\begin{figure}[ht]
    \centering
    \subfigure[\smt on Mutag trainset.]{
        \includegraphics[width=0.31\textwidth]{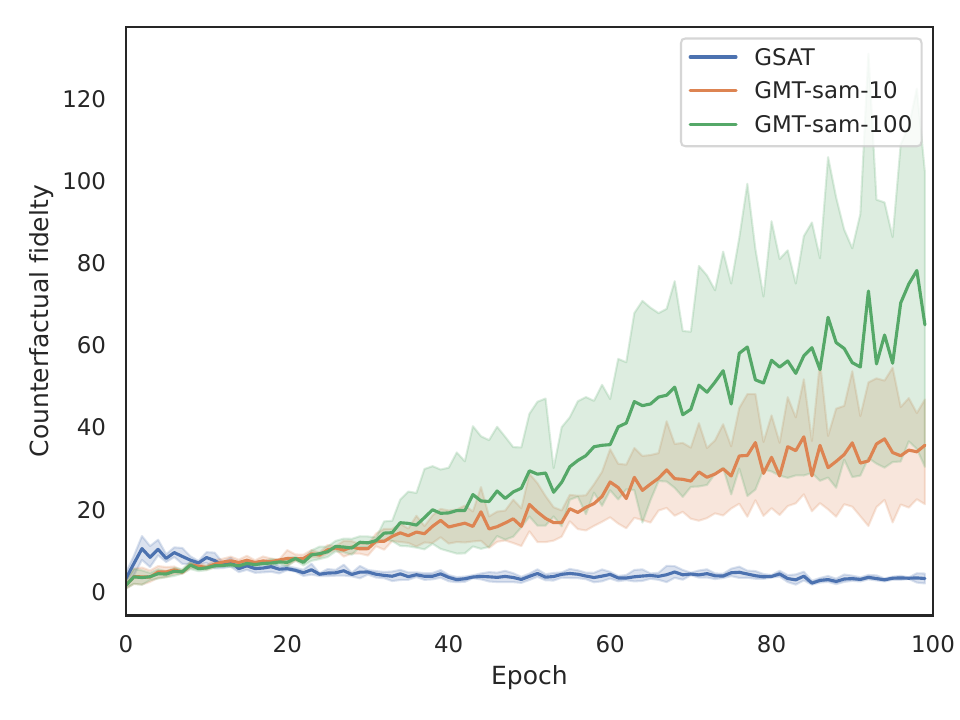}
    }
    \subfigure[\smt on Mutag validation set.]{
        \includegraphics[width=0.31\textwidth]{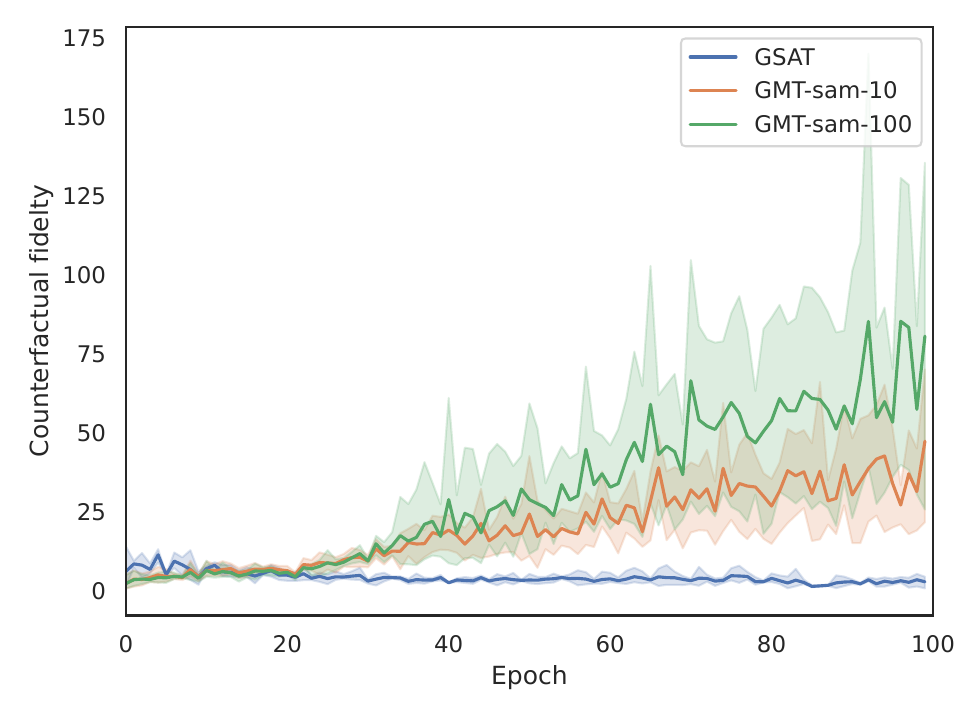}
    }
    \subfigure[\smt on Mutag test set.]{
        \includegraphics[width=0.31\textwidth]{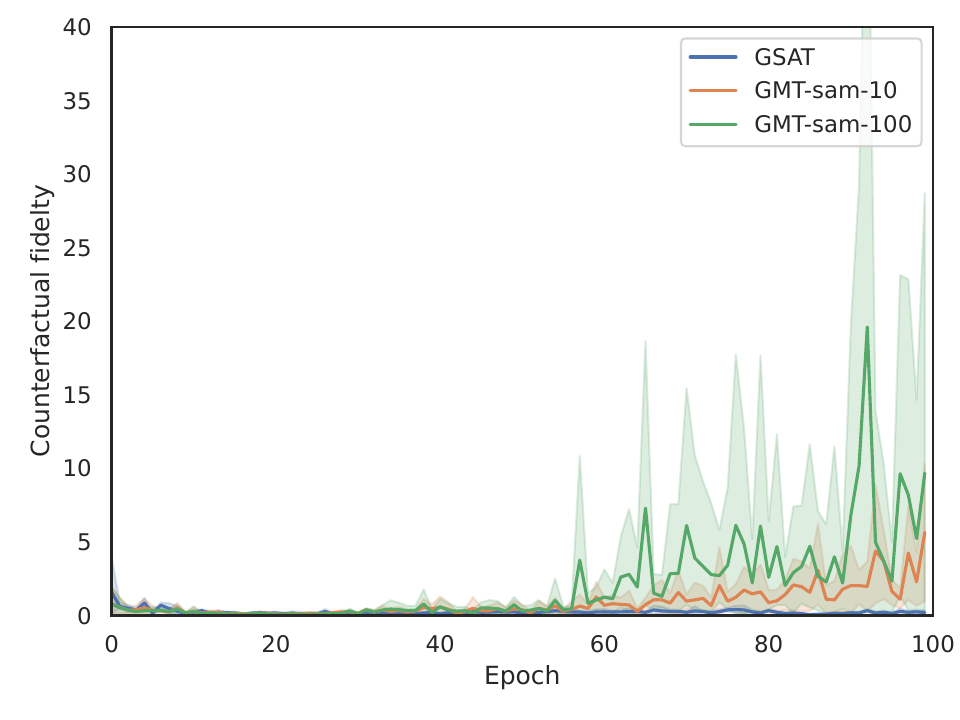}
    }
    \caption{Comparison of \gsat and the simulated \smt in counterfactual fidelity on Mutag measured via KL divergence.}
    \label{fig:cf_mu_kl_appdx}
\end{figure}

\begin{figure}[H]
    \centering
    \subfigure[\smt on BA-2Motifs trainset.]{
        \includegraphics[width=0.31\textwidth]{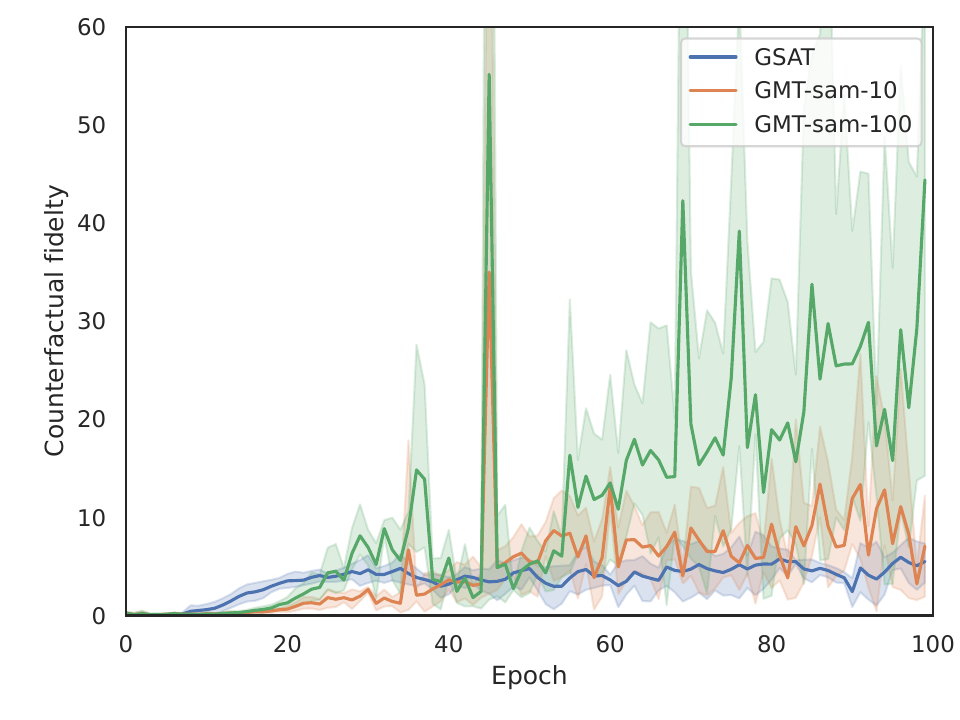}
    }
    \subfigure[\smt on BA-2Motifs valset.]{
        \includegraphics[width=0.31\textwidth]{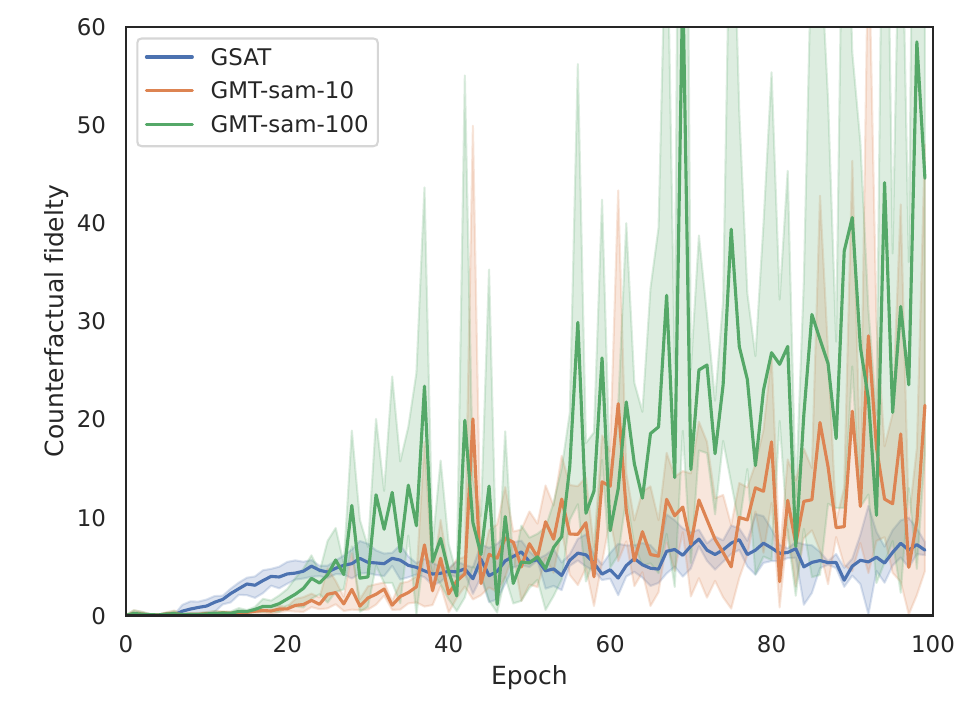}
    }
    \subfigure[\smt on BA-2Motifs test set.]{
        \includegraphics[width=0.31\textwidth]{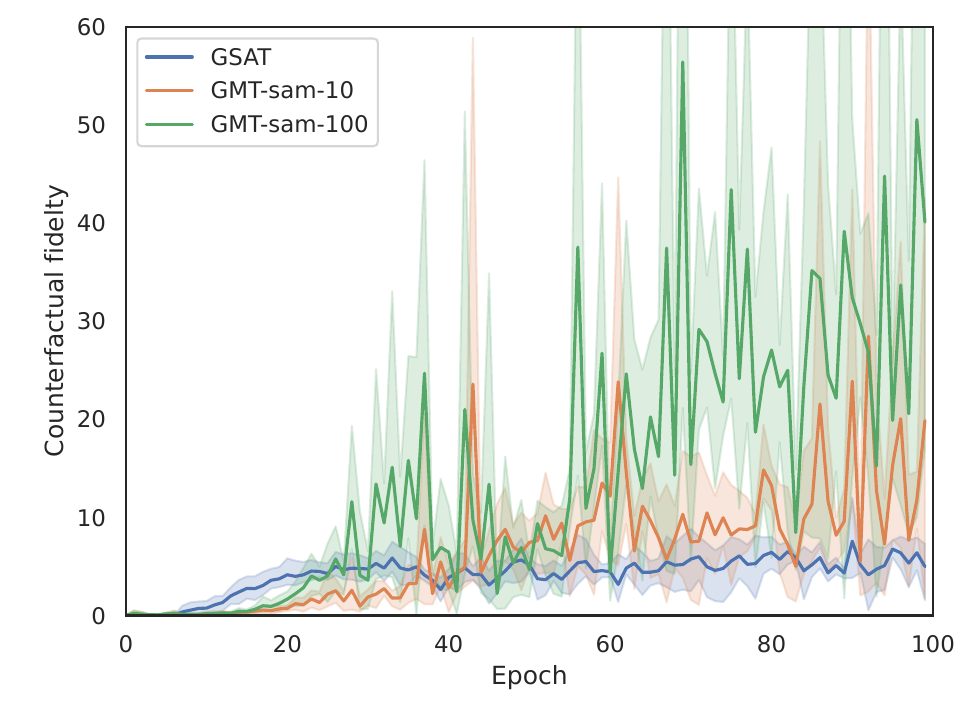}
    }
    \caption{Comparison of \gsat and the simulated \smt in counterfactual fidelity on BA-2Motifs measured via JSD divergence.}
    \label{fig:cf_ba_jsd_appdx}
\end{figure}

\begin{figure}[ht]
    \centering
    \subfigure[\smt on Mutag trainset.]{
        \includegraphics[width=0.31\textwidth]{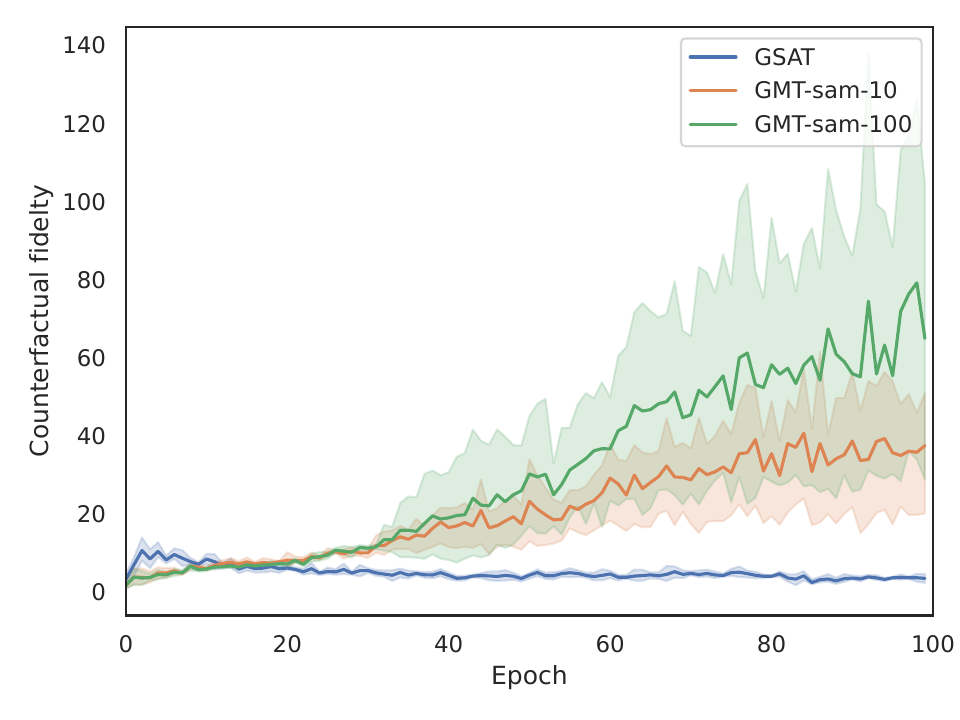}
    }
    \subfigure[\smt on Mutag validation set.]{
        \includegraphics[width=0.31\textwidth]{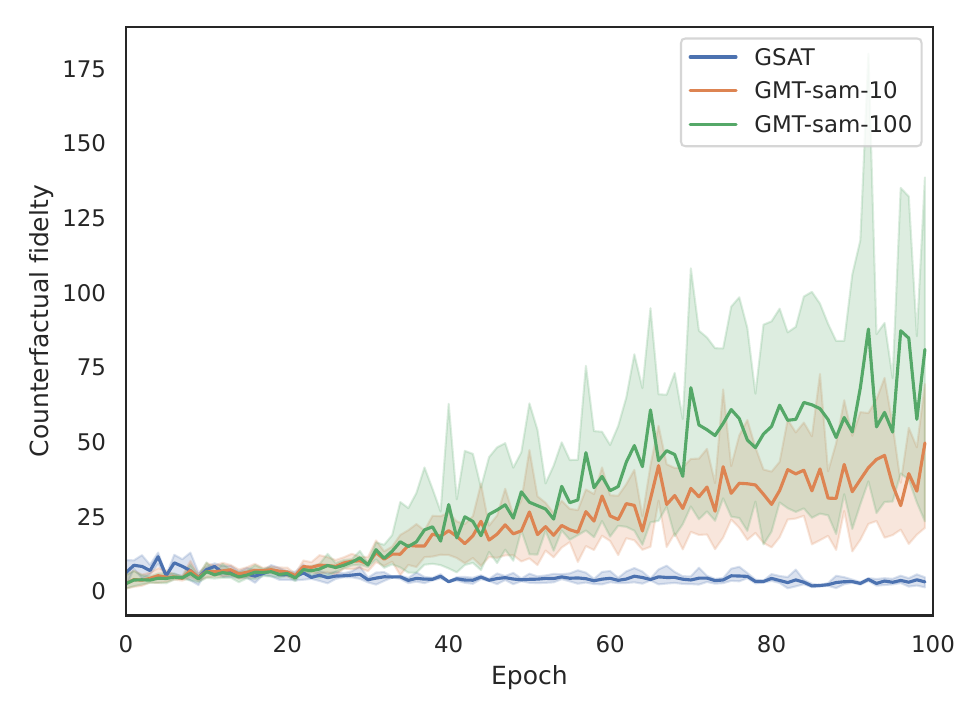}
    }
    \subfigure[\smt on Mutag test set.]{
        \includegraphics[width=0.31\textwidth]{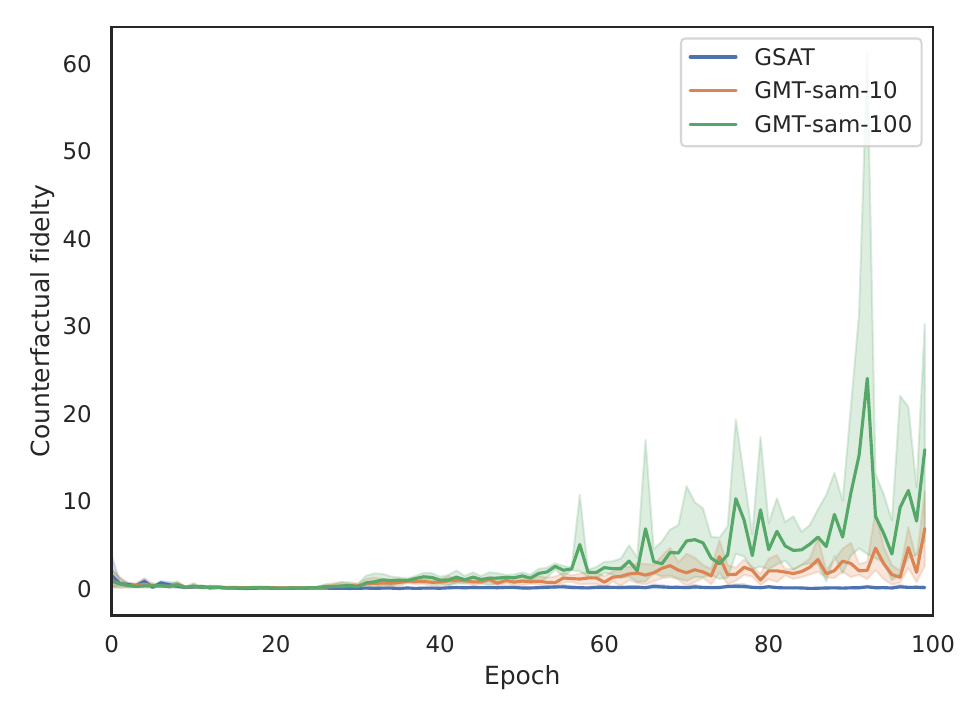}
    }
    \caption{Comparison of \gsat and the simulated \smt in counterfactual fidelity on Mutag measured via JSD divergence.}
    \label{fig:cf_mu_jsd_appdx}
\end{figure}

Shown as in Fig.~\ref{fig:cf_ba_kl_appdx},~\ref{fig:cf_mu_kl_appdx},
we plot the counterfactual fidelity results of \gsat and the simulated \smt via \ourss with $10$ and $100$  on BA-2Motifs and Mutag datasets measured via KL divergence.
Fig.~\ref{fig:cf_ba_jsd_appdx},~\ref{fig:cf_mu_jsd_appdx} show the counterfactual fidelity results of \gsat and the simulated \smt via \ourss with $10$ and $100$
on BA-2Motifs and Mutag datasets measured via JSD divergence.
It can be found that, the gap in counterfactual fidelity measured in KL divergence or JSD divergence can be even larger between \gsat and \smt,
growing up to $10$ times.
These results can serve as strong evidence for the degenerated interpretability caused by the failure of \smt approximation.


\begin{figure}[ht]
    \centering
    \subfigure[BA-2Motifs trainset.]{
        \includegraphics[width=0.31\textwidth]{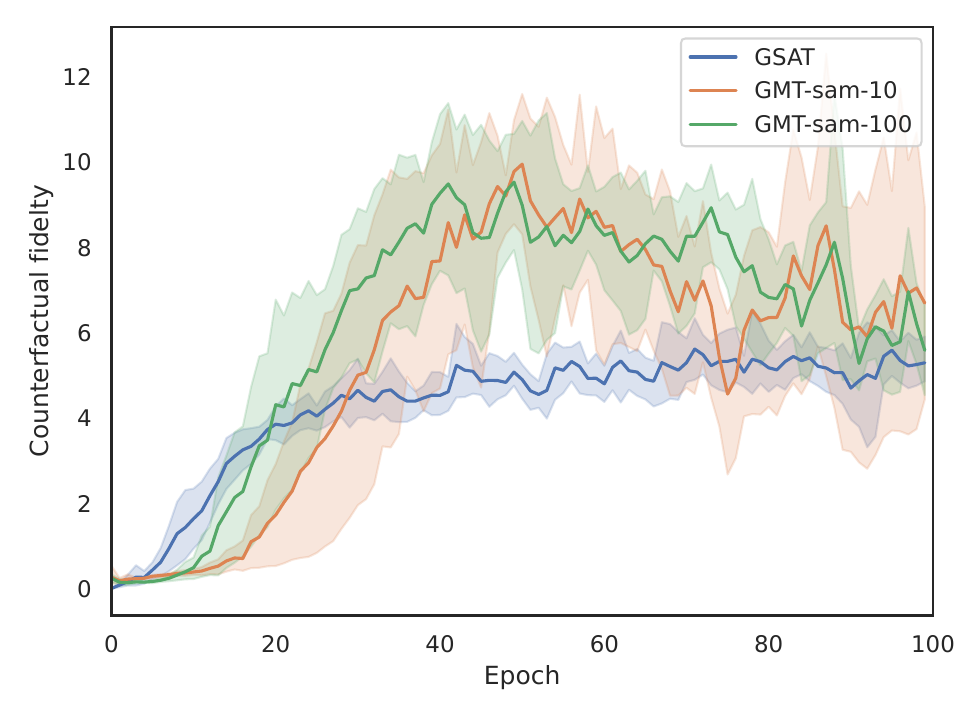}
    }
    \subfigure[BA-2Motifs valset.]{
        \includegraphics[width=0.31\textwidth]{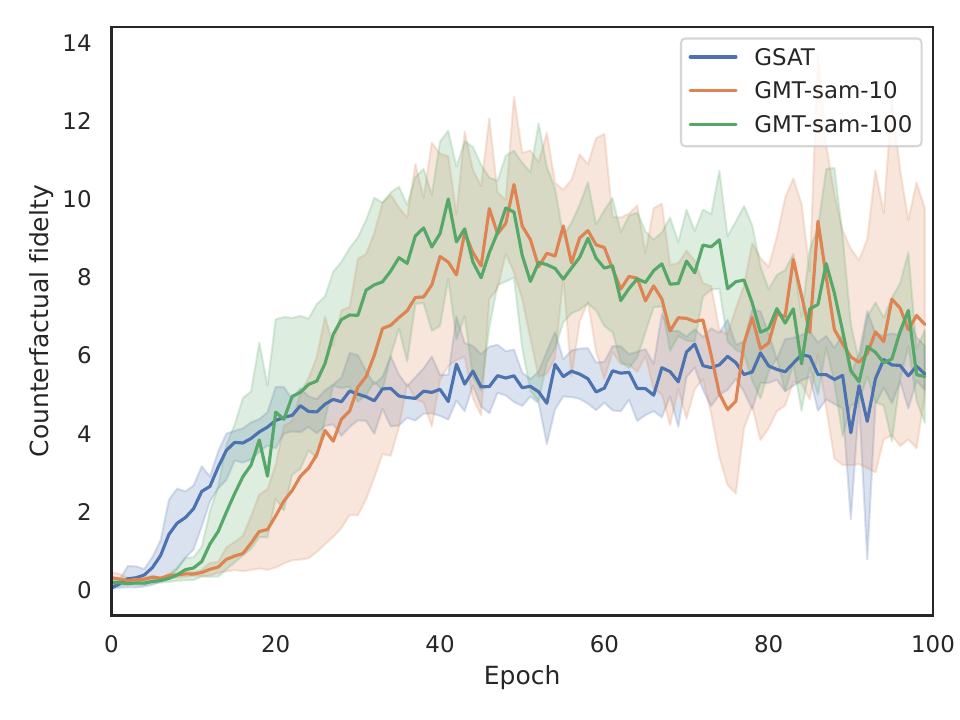}
    }
    \subfigure[BA-2Motifs test set.]{
        \includegraphics[width=0.31\textwidth]{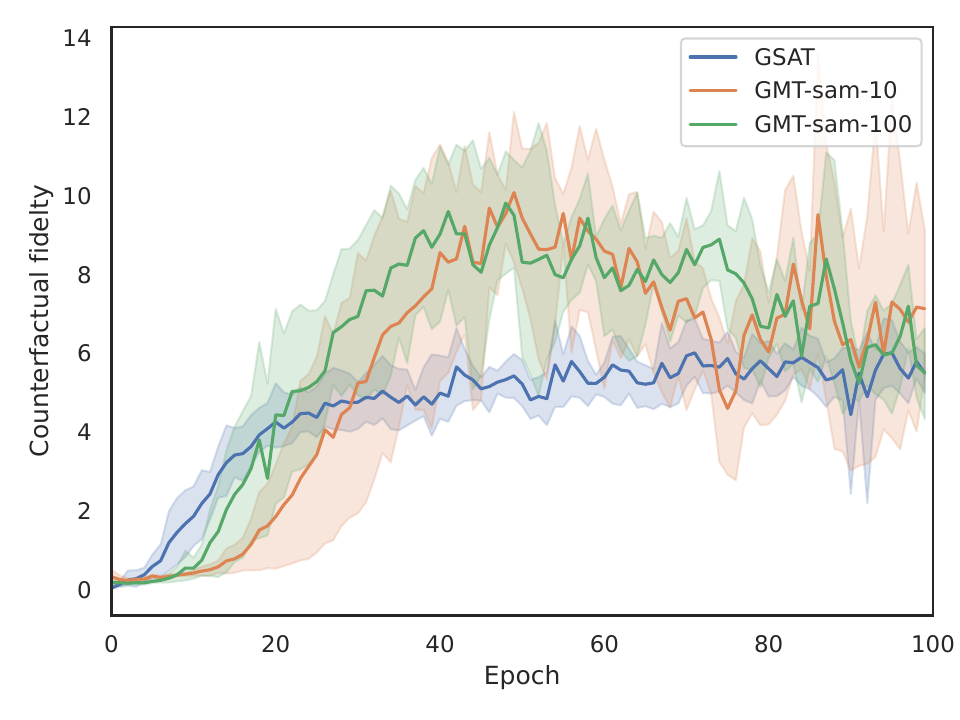}
    }
    \caption{The \ours optimization issue in terms of Comparison of \gsat and the simulated \smt in counterfactual fidelity on BA-2Motifs.}
    \label{fig:cf_ba_opt_appdx}
\end{figure}

\begin{figure}[H]
    \centering
    \subfigure[Mutag trainset.]{
        \includegraphics[width=0.31\textwidth]{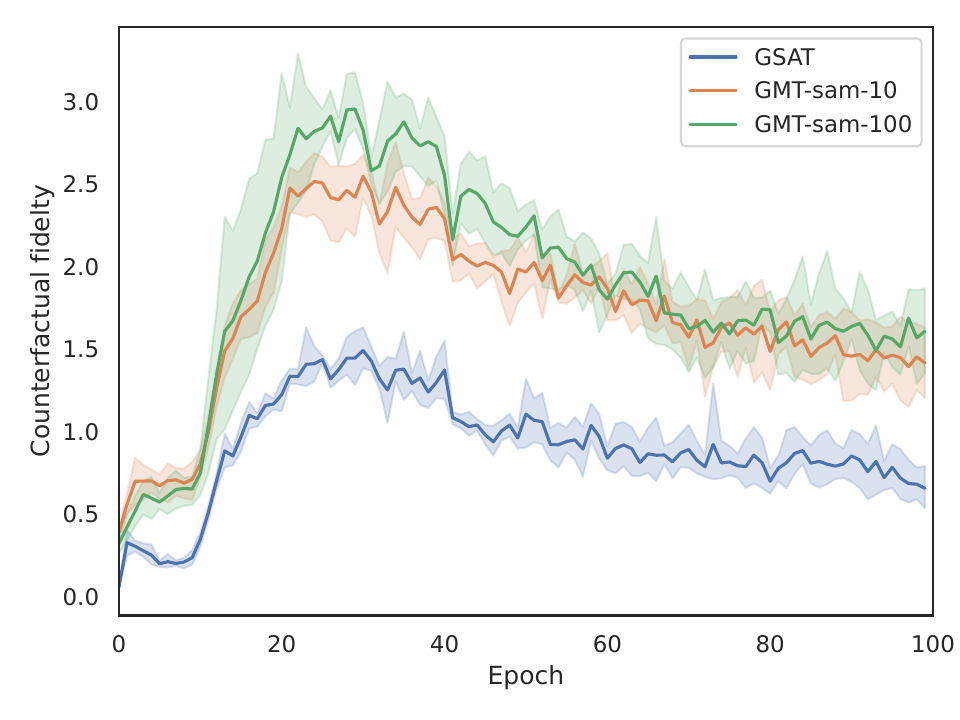}
    }
    \subfigure[Mutag validation set.]{
        \includegraphics[width=0.31\textwidth]{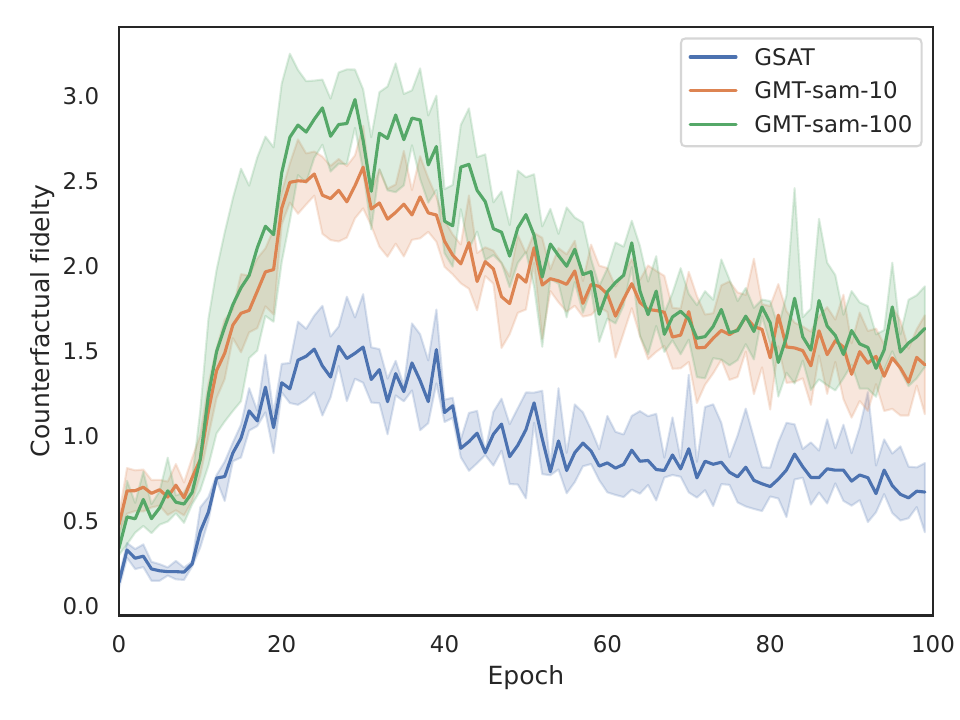}
    }
    \subfigure[Mutag test set.]{
        \includegraphics[width=0.31\textwidth]{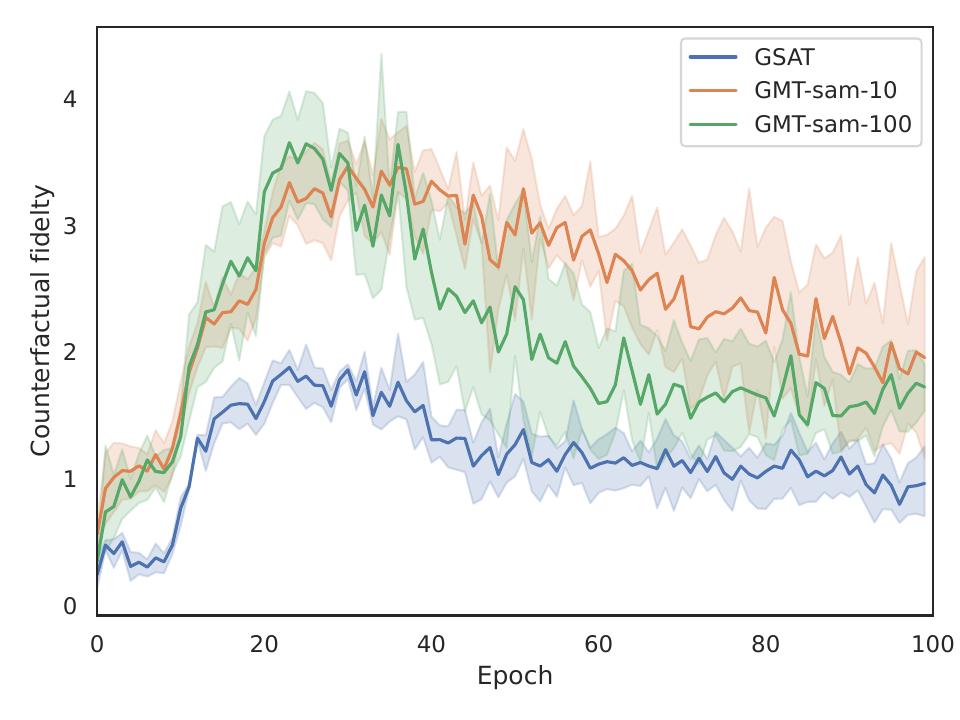}
    }
    \caption{The \ours optimization issue in terms of Comparison of \gsat and the simulated \smt in counterfactual fidelity on Mutag.}
    \label{fig:cf_mu_opt_appdx}
\end{figure}

Shown as in Fig.~\ref{fig:cf_ba_opt_appdx},~\ref{fig:cf_mu_opt_appdx},
we plot the counterfactual fidelity results of \gsat and the simulated \smt via \ourss with $10$ and $100$  on BA-2Motifs and Mutag datasets.
Compared to previous results, the \ourss in Fig.~\ref{fig:cf_ba_opt_appdx},~\ref{fig:cf_mu_opt_appdx} does not use any warmup strategies
that may suffer from the optimization issue as discussed in Sec.~\ref{sec:gmt_impl_dis_appdx}.
It can be found that, at the begining of the optimization, \ourss demonstrates increasing counterfactual fidelity.
However, as the optimization keeps proceeding, the counterfactual fidelity of \ourss will degenerate, because of fitting to the trivial solution of the \gsat objective.
Consequently, the interpretation results will degenerate too at the end of the optimization.

\subsection{\smt approximation gap analysis}
\label{sec:smt_gap_viz_appdx}

\begin{figure}[H]
    \centering
    \subfigure[BA-2Motifs trainset.]{
        \includegraphics[width=0.31\textwidth]{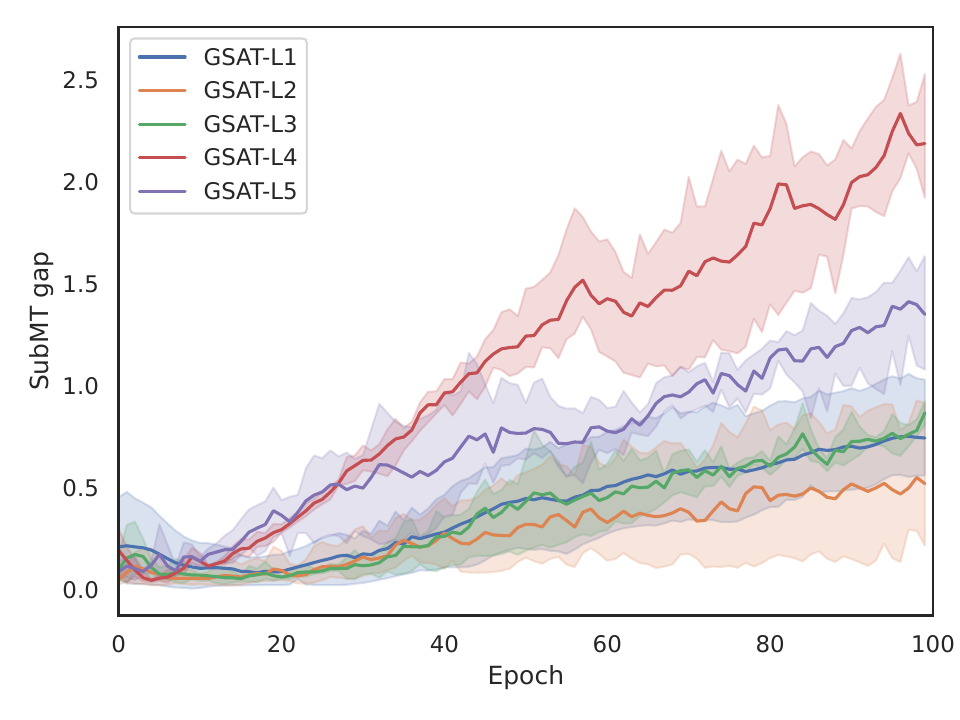}
    }
    \subfigure[BA-2Motifs valset.]{
        \includegraphics[width=0.31\textwidth]{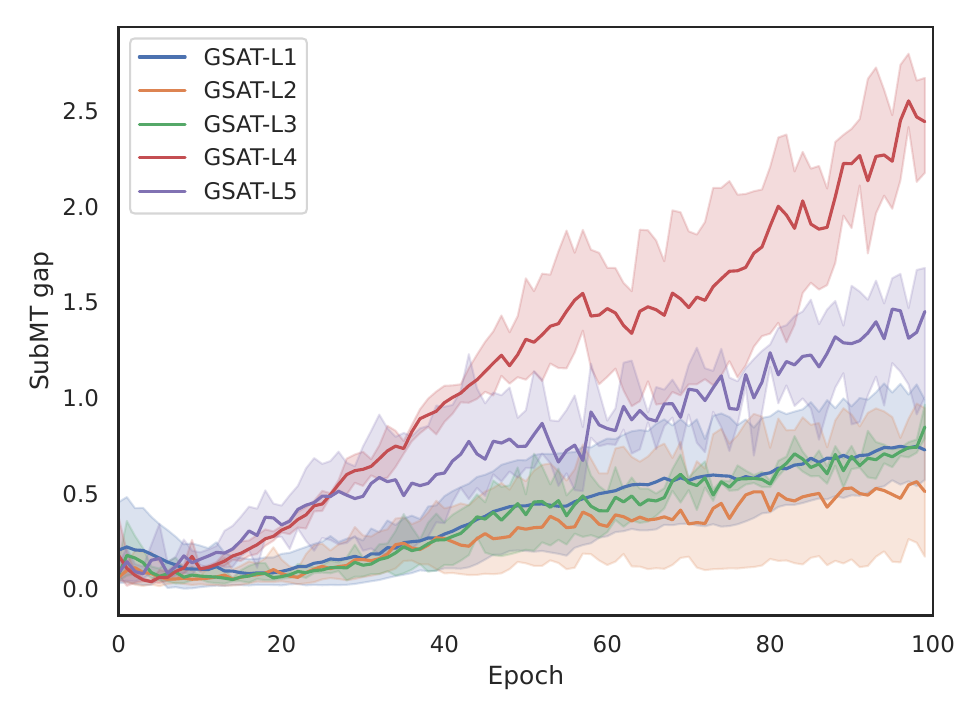}
    }
    \subfigure[BA-2Motifs test set.]{
        \includegraphics[width=0.31\textwidth]{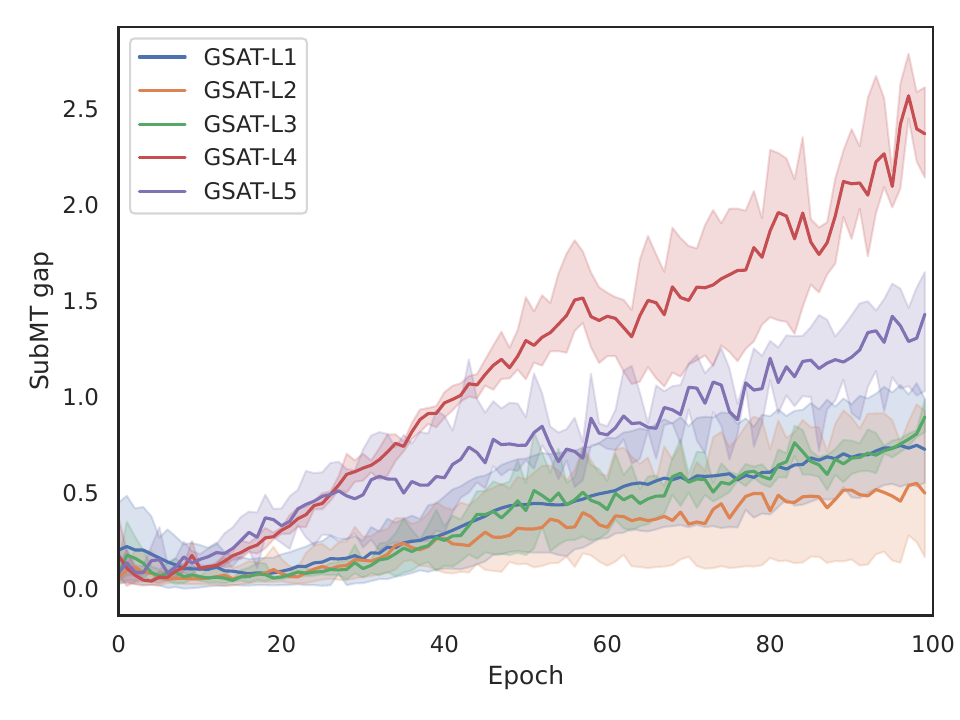}
    }
    \caption{The \smt approximation gap of \gsat with SGC on BA-2Motifs.}
    \label{fig:gap_ba2_sgc_appdx}
\end{figure}
\begin{figure}[H]
    \centering
    \subfigure[BA-2Motifs trainset.]{
        \includegraphics[width=0.31\textwidth]{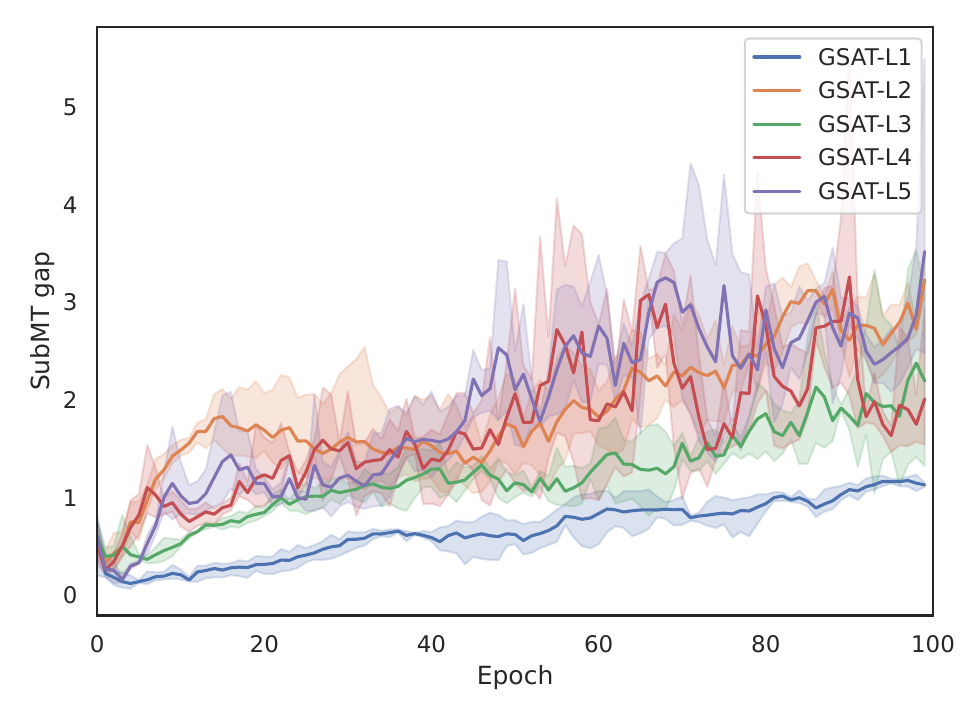}
    }
    \subfigure[BA-2Motifs valset.]{
        \includegraphics[width=0.31\textwidth]{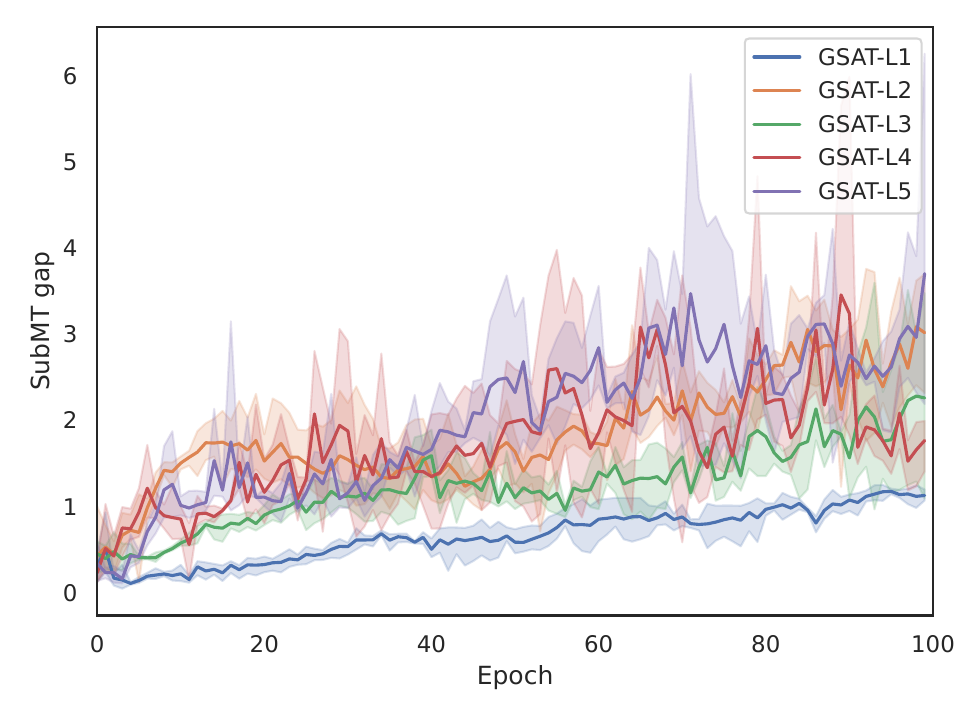}
    }
    \subfigure[BA-2Motifs test set.]{
        \includegraphics[width=0.31\textwidth]{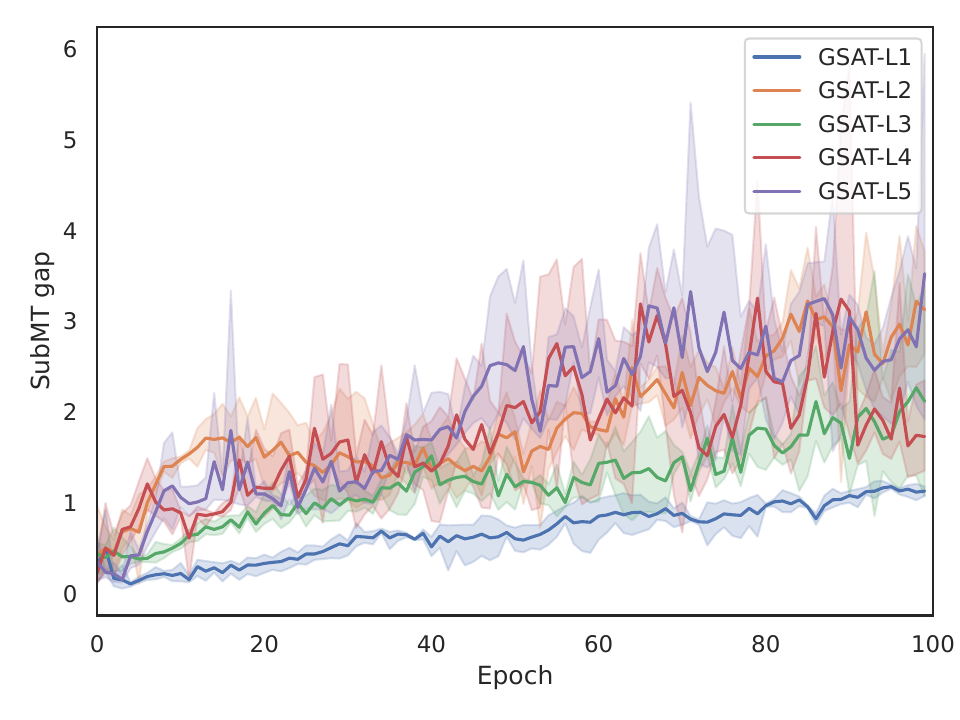}
    }
    \caption{The \smt approximation gap of \gsat with GIN on BA-2Motifs.}
    \label{fig:gap_ba2_gin_appdx}
\end{figure}
\begin{figure}[H]
    \centering
    \subfigure[Mutag trainset.]{
        \includegraphics[width=0.31\textwidth]{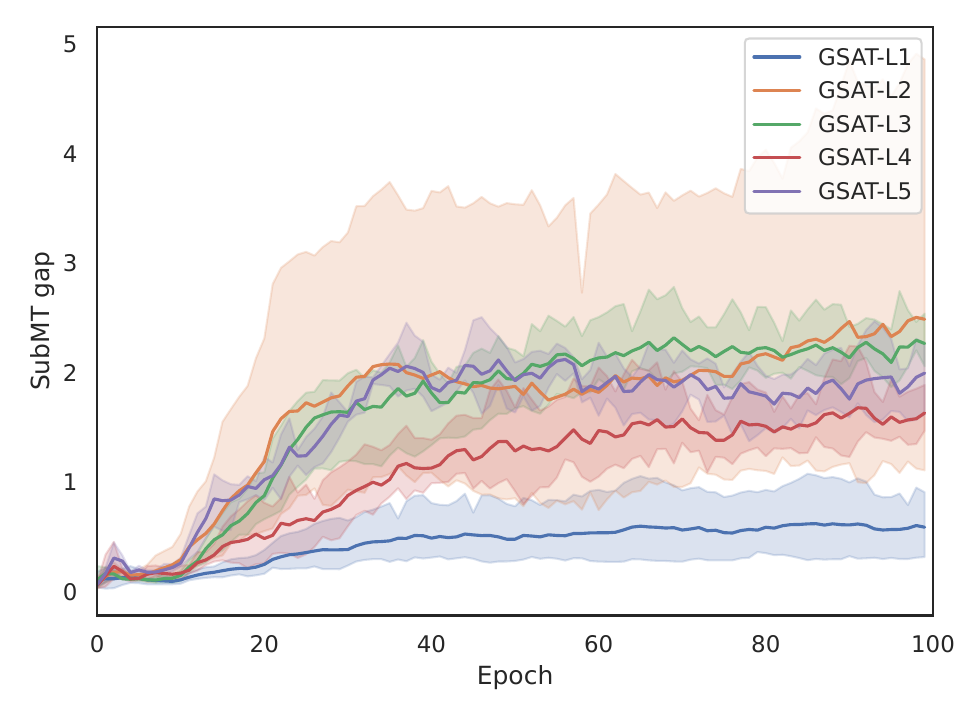}
    }
    \subfigure[Mutag validation set.]{
        \includegraphics[width=0.31\textwidth]{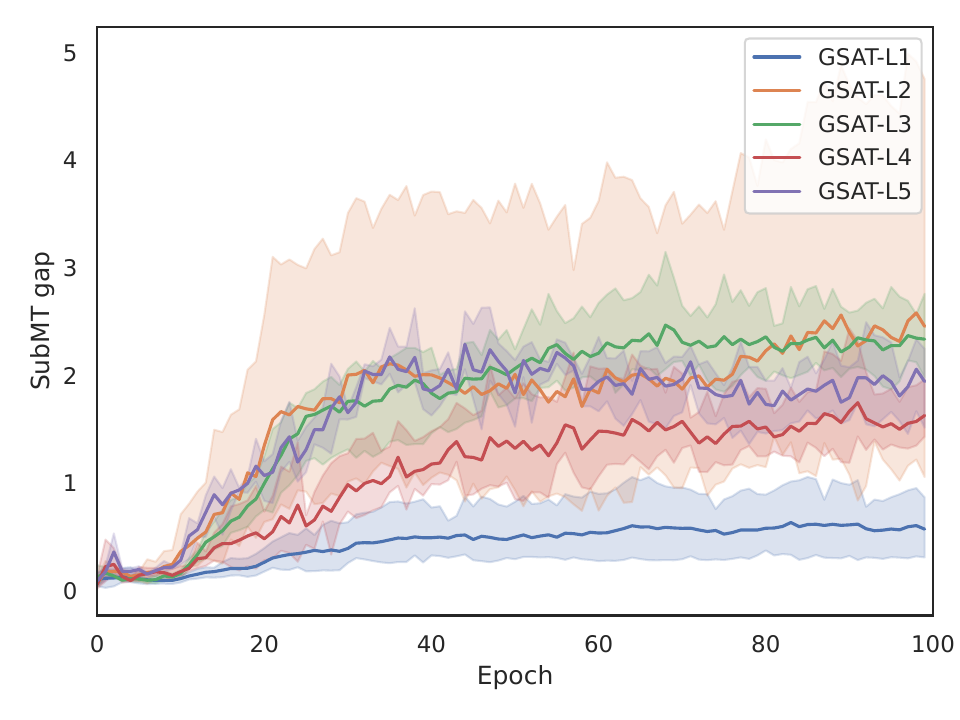}
    }
    \subfigure[Mutag test set.]{
        \includegraphics[width=0.31\textwidth]{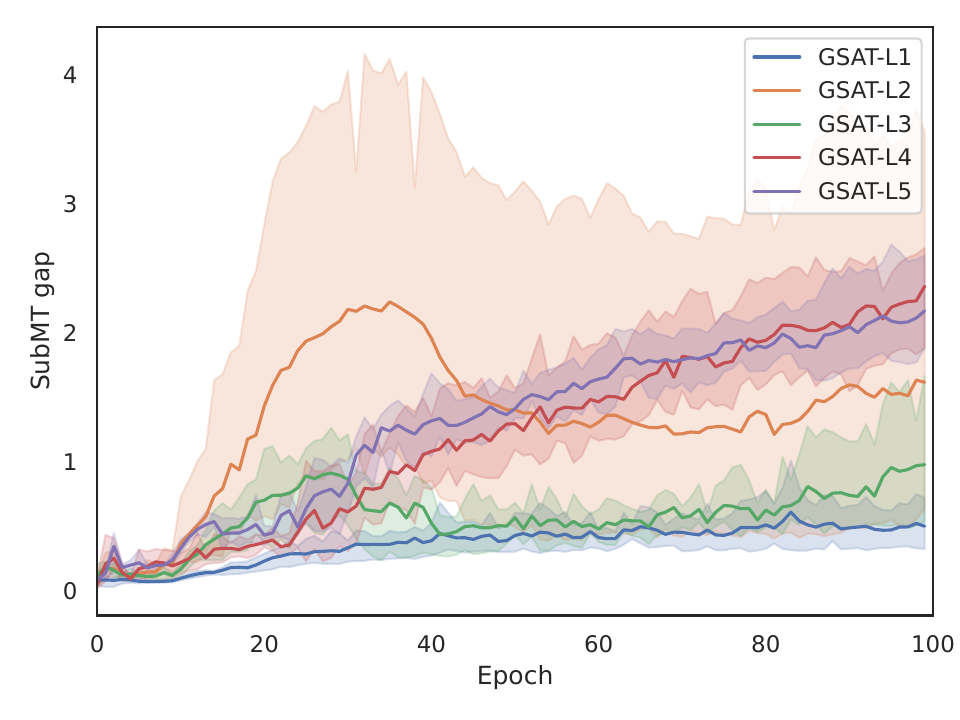}
    }
    \caption{The \smt approximation gap of \gsat with SGC on Mutag.}
    \label{fig:gap_mu_sgc_appdx}
\end{figure}
\begin{figure}[H]
    \centering
    \subfigure[Mutag trainset.]{
        \includegraphics[width=0.31\textwidth]{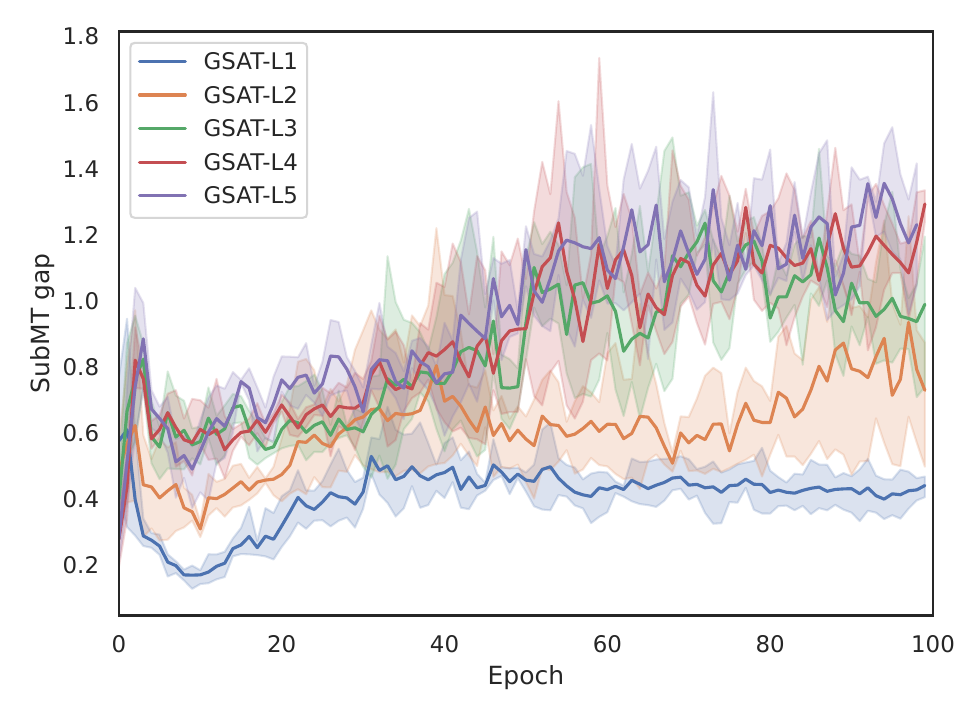}
    }
    \subfigure[Mutag validation set.]{
        \includegraphics[width=0.31\textwidth]{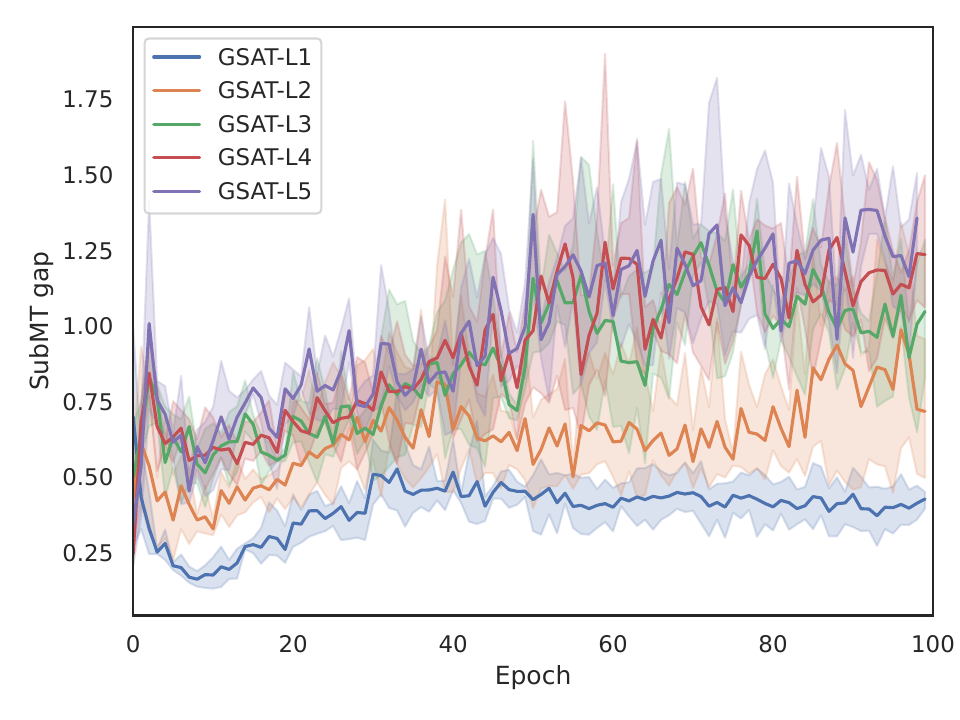}
    }
    \subfigure[Mutag test set.]{
        \includegraphics[width=0.31\textwidth]{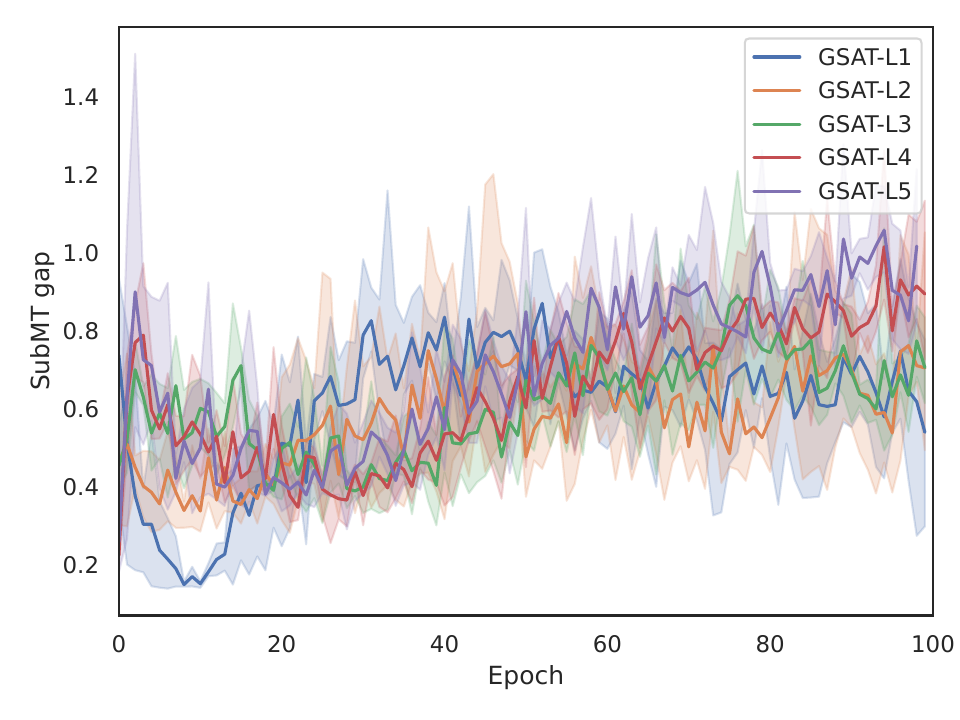}
    }
    \caption{The \smt approximation gap of \gsat with GIN on Mutag.}
    \label{fig:gap_mu_gin_appdx}
\end{figure}

Fig.~\ref{fig:gap_ba2_sgc_appdx} and ~\ref{fig:gap_ba2_gin_appdx}, Fig.~\ref{fig:gap_mu_sgc_appdx} and ~\ref{fig:gap_mu_gin_appdx} demonstrate the \smt approximation gaps of \gsat implemented
in GIN and SGC on BA\_2Motifs and Mutag respectively.
To fully verify Proposition~\ref{thm:submt_gap_appdx}, we range the number of layers of GIN and SGC from $1$ to $5$.
It can be found that the results are well aligned with Proposition~\ref{thm:submt_gap_appdx}.
When the number of layers is $1$, the \smt approximation gap is smallest, because of more ``linearity'' in the network.
While along with the growing number of GNN layers, the network becomes more ``unlinear'' such that the \smt approximation gap will be larger.

\begin{table}[t]
\caption{Results on node-level classification tasks.}
\label{tab:node}
\small\centering\sc
\resizebox{\textwidth}{!}{
\begin{tabular}{@{}lllllll@{}}\toprule
 &  & { \textbf{Citeseer}} & { \textbf{Pubmed}} & { \textbf{Coauthor-CS}} & { \textbf{Coauthor-Physics}} &                                              \\\midrule

                                 & { Num Nodes}                             & { 2,708}                                     & { 3,327}                                   & { 19,717}                                       & { 18,333}                                            & { 34,493}                \\

                                 & { Num Edges}                             & { 10,556}                                    & { 9,228}                                   & { 88,651}                                       & { 163,788}                                           & { 495,924}               \\\midrule

{ SunnyGNN}                          & { Prediction Acc. ($\uparrow$)}                   & { 81.28\std{0.75}}                               & { 67.44\std{2.07}}                             & { 77.10\std{0.46}}                                  & { 85.16\std{1.53}}                                       & { 92.58\std{0.26}}           \\

                                 & { Counterfactual Fid. ($\uparrow$)}               & { 1.25\std{0.34}}                                & { 0.59\std{0.21}}                              & { 1.83\std{0.17}}                                   & { 6.72\std{1.58}}                                        & { 9.13\std{0.92}}            \\

                                 & { Sufficiency Fid. ($\uparrow$)}                  & {\textbf{73.42\std{3.55}}}                      & { 58.32\std{0.64}}                             & { 57.48\std{4.73}}                                  & { 77.64\std{7.52}}                                       & { 85.84\std{02.78}}           \\\midrule

{ GSAT}                              & { Prediction Acc. ($\uparrow$)}                   & { 81.14\std{0.51}}                               & { 68.18\std{1.23}}                             & {\textbf{77.42\std{0.25}}}                         & { 85.30\std{0.65}}                                       & {\textbf{92.68\std{0.50}}}  \\

                                 & { Counterfactual Fid. ($\uparrow$)}               & { 1.66\std{0.14}}                                & { 0.52\std{0.20}}                              & { 2.77\std{0.11}}                                   & { 5.98\std{0.82}}                                        & { 8.73\std{1.19}}            \\

                                 & { Sufficiency Fid. ($\uparrow$)}                  & { 71.54\std{2.82}}                               & {\textbf{61.06\std{4.63}}}                    & { 56.12\std{4.71}}                                  & { 77.64\std{7.04}}                                       & { 85.88\std{2.63}}           \\\midrule

{ GMT-lin}                           & { Prediction Acc. ($\uparrow$)}                   & {\textbf{81.52\std{0.75}}}                      & {\textbf{68.56\std{1.11}}}                    & {\textbf{77.42\std{0.21}}}                         & {\textbf{85.66\std{0.60}}}                              & { 92.54\std{0.45}}           \\

                                 & { Counterfactual Fid. ($\uparrow$)}               & { 1.74\std{0.22}}                                & { 0.60\std{0.17}}                              & { 2.62\std{0.20}}                                   & { 5.40\std{2.54}}                                        & { 16.35\std{7.88}}           \\

                                 & { Sufficiency Fid. ($\uparrow$)}                  & { 71.96\std{3.51}}                               & { 59.04\std{1.64}}                             & { 54.88\std{5.07}}                                  & { 70.86\std{8.63}}                                       & { 70.4\std{18.99}}           \\\midrule

{ GMT-sam}                           & { Prediction Acc. ($\uparrow$)}                   & { 80.76\std{0.97}}                               & { 68.02\std{0.66}}                             & { 76.92\std{0.21}}                                  & { 84.22\std{1.00}}                                       & { 92.58\std{0.35}}           \\

                                 & { Counterfactual Fid. ($\uparrow$)}               & {\textbf{1.88\std{0.26}}}                       & {\textbf{0.62\std{0.08}}}                     & {\textbf{2.81\std{0.26}}}                          & {\textbf{6.90\std{2.41}}}                               & {\textbf{18.18\std{11.29}}} \\

                                 & { Sufficiency Fid. ($\uparrow$)}                  & { 72.66\std{3.00}}                               & { 60.00\std{1.45}}                             & {\textbf{61.76\std{1.53}}}                         & {\textbf{78.60\std{5.67}}}                              & {\textbf{85.96\std{3.52}}} \\\bottomrule
\end{tabular}}
\end{table}
\subsection{XGNNs on node-level classification}
\label{sec:node_xgnn}
We extend our studies to node-level classification, in order to verify our previous discussion that the results can also generalize to node-level tasks if we convert the node-level tasks into graph classification based on the ego-graphs of the respective central nodes.

Specifically, we follow the experimental setup of a recent work called SunnyGNN~\citep{sunny_gnn} to evaluate the interpretation and prediction performances. Since ground truth labels are not available, we consider two interpretation metrics:
\begin{itemize}
    \item Counterfactual Fidelity proposed in our work: the higher the better.
    \item Sufficiency Fidelity modified from~\citep{sunny_gnn}: the higher the better. Note that the original evaluation of interpretation in~\citep{sunny_gnn} considers a post-hoc setting, and we modified it for the intrinsic interpretation setting. \citet{sunny_gnn} also proposes Necessity Fidelity, and we omit it as it is close to $0$ for all methods.
\end{itemize}
For \ourss, we do not conduct retraining for clarity, which could further improve the prediction performance as demonstrated in Table~\ref{tab:retrain}. 
In addition, we consider five datasets including Cora~\citep{cora}, Citeseer~\citep{citeseer}, Pubmed~\citep{Sen2008CollectiveCI}, Coauthor-CS and Coauthor-Physics~\citep{Shchur2018PitfallsOG} from Microsoft Academic Graph.\footnote{\url{https://www.microsoft.com/en-us/research/project/microsoft-academic-graph/}}
The results on the node-level classification tasks are given in Table~\ref{tab:node}. From the table, we can find that, compared to the state-of-the-art XGNNs in node classification, \ourss and \oursl achieve a competitive prediction performance while bringing significant improvements in terms of interpretation performance, aligned with our discussion.

\subsection{Software and Hardware}
\label{sec:exp_software_appdx}
We implement our methods with PyTorch~\citep{pytorch} and PyTorch Geometric~\citep{pytorch_geometric} 2.0.4.
We ran our experiments on Linux Servers installed with V100 graphics cards and CUDA 11.3.

\subsection{Interpretation Visualization}
\label{sec:inter_viz_appdx}
To better understand the results, we provide visualizations of the learned interpretable subgraphs by \gsat and \ourss in the Spurious-Motif datasets, as well as the learned interpretable subgraphs by \ourss in OGBG-Molhiv dataset.

\begin{figure}[H]
    \centering
    \subfigure[Spurious-Motif class $0$ under bias$=0.5$ by \gsat.]{
        \includegraphics[width=\textwidth]{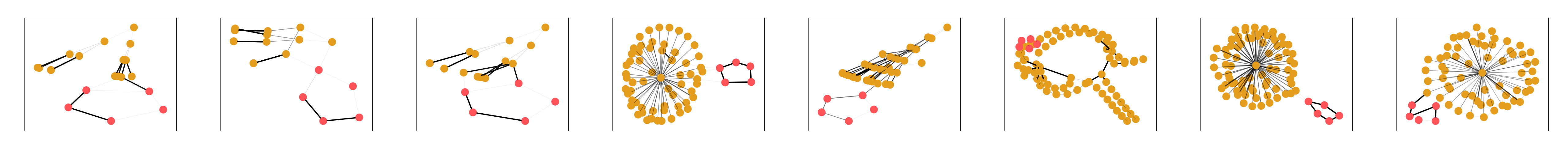}
    }
    \subfigure[Spurious-Motif class $0$ under bias$=0.5$ by \ourss.]{
        \includegraphics[width=\textwidth]{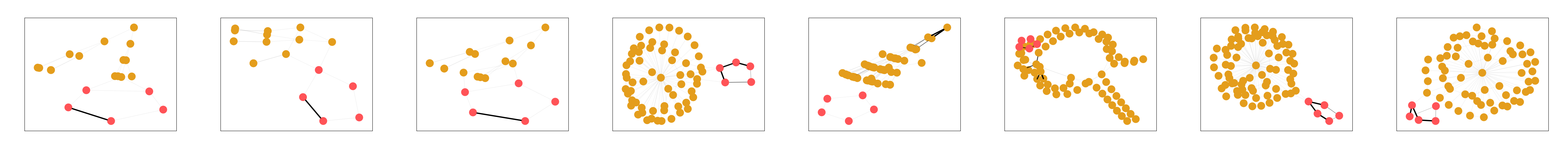}
    }
    \subfigure[Spurious-Motif class $1$ under bias$=0.5$ by \gsat.]{
        \includegraphics[width=\textwidth]{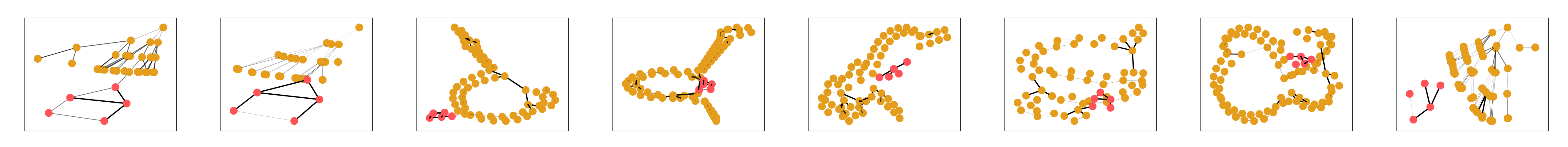}
    }
    \subfigure[Spurious-Motif class $1$ under bias$=0.5$ by \ourss.]{
        \includegraphics[width=\textwidth]{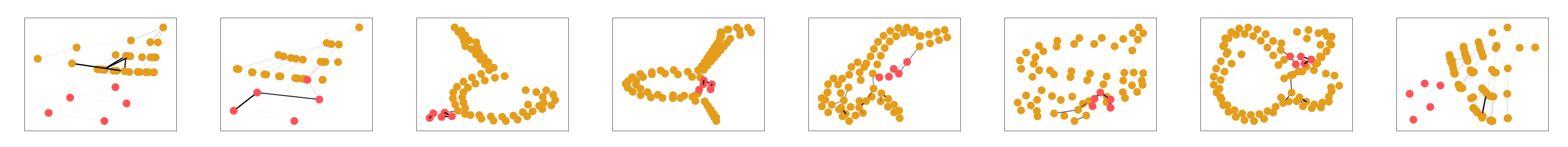}
    }
    \subfigure[Spurious-Motif class $2$ under bias$=0.5$ by \gsat.]{
        \includegraphics[width=\textwidth]{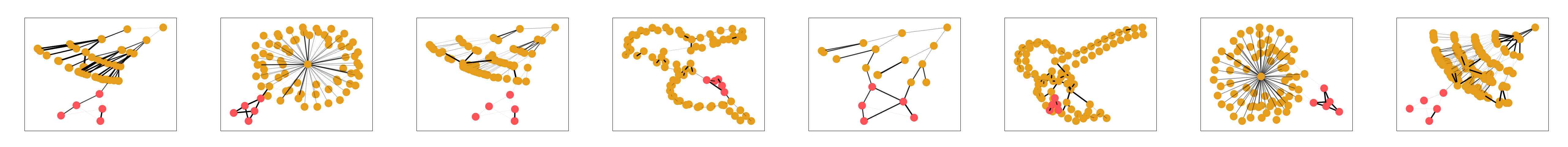}
    }
    \subfigure[Spurious-Motif class $2$ under bias$=0.5$ by \ourss.]{
        \includegraphics[width=\textwidth]{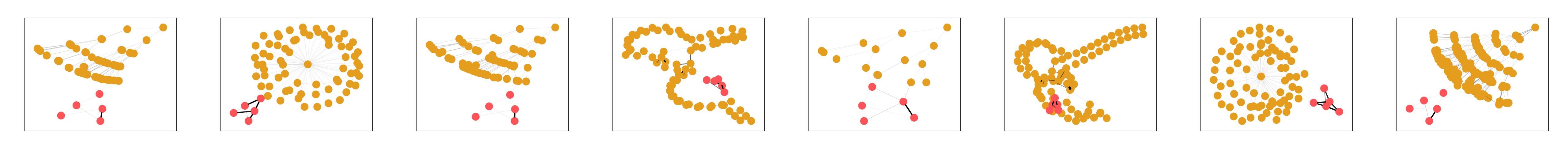}
    }
    \caption{
        Learned interpretable subgraphs by \gsat and \ourss on Spurious-Motif $b=0.5$.}
    \label{fig:sp5_viz_appdx}
\end{figure}

\begin{figure}[H]
    \centering
    \subfigure[Spurious-Motif class $0$ under bias$=0.5$ by \gsat.]{
        \includegraphics[width=\textwidth]{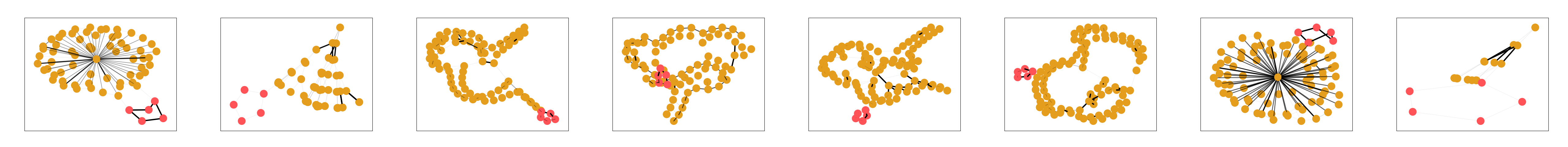}
    }
    \subfigure[Spurious-Motif class $0$ under bias$=0.5$ by \ourss.]{
        \includegraphics[width=\textwidth]{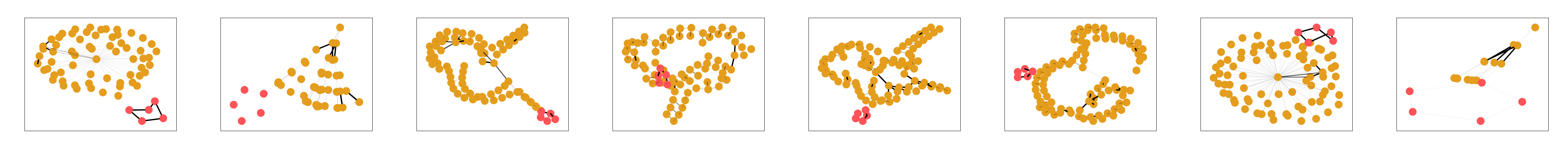}
    }
    \subfigure[Spurious-Motif class $1$ under bias$=0.5$ by \gsat.]{
        \includegraphics[width=\textwidth]{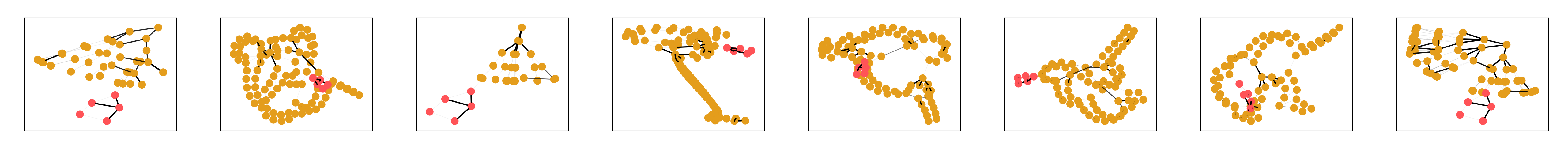}
    }
    \subfigure[Spurious-Motif class $1$ under bias$=0.5$ by \ourss.]{
        \includegraphics[width=\textwidth]{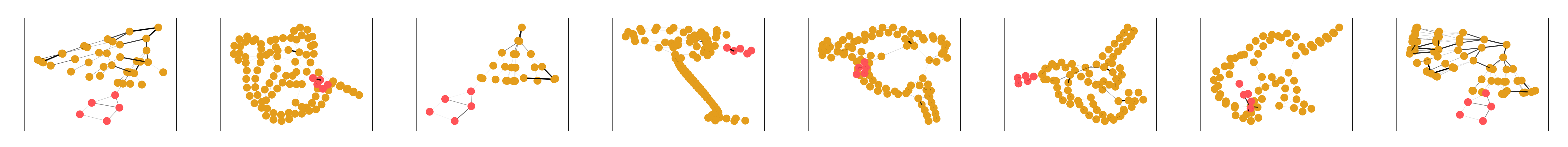}
    }
    \subfigure[Spurious-Motif class $2$ under bias$=0.5$ by \gsat.]{
        \includegraphics[width=\textwidth]{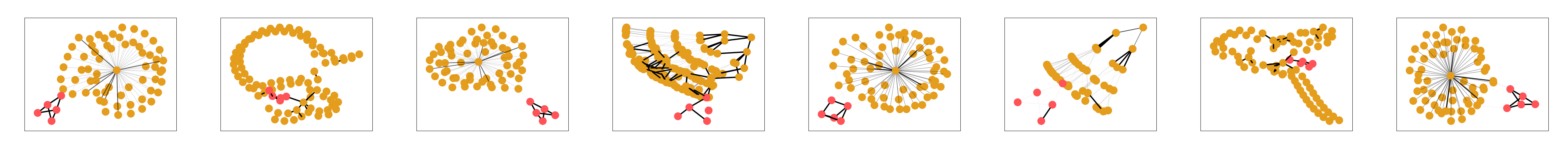}
    }
    \subfigure[Spurious-Motif class $2$ under bias$=0.5$ by \ourss.]{
        \includegraphics[width=\textwidth]{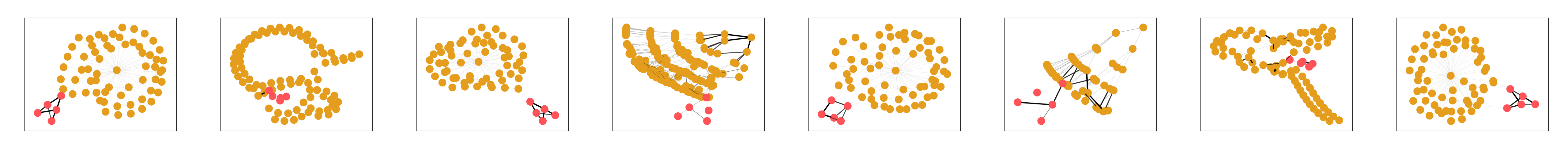}
    }
    \caption{
        Learned interpretable subgraphs by \gsat and \ourss on Spurious-Motif $b=0.7$.}
    \label{fig:sp7_viz_appdx}
\end{figure}

\begin{figure}[H]
    \centering
    \subfigure[Spurious-Motif class $0$ under bias$=0.5$ by \gsat.]{
        \includegraphics[width=\textwidth]{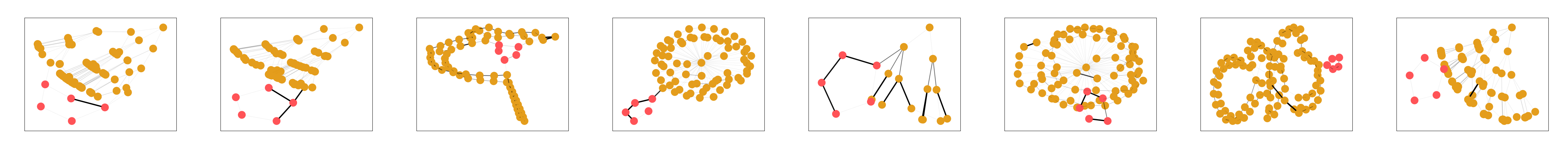}
    }
    \subfigure[Spurious-Motif class $0$ under bias$=0.5$ by \ourss.]{
        \includegraphics[width=\textwidth]{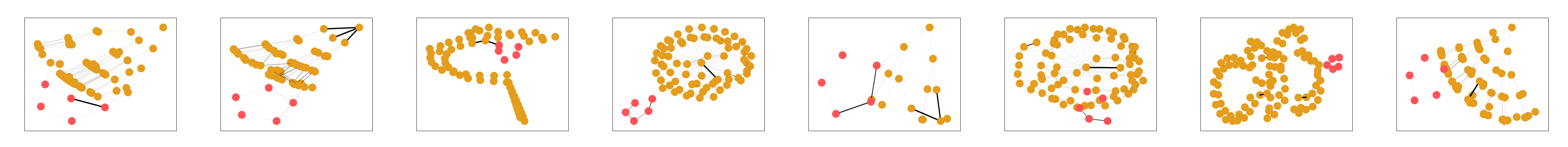}
    }
    \subfigure[Spurious-Motif class $1$ under bias$=0.5$ by \gsat.]{
        \includegraphics[width=\textwidth]{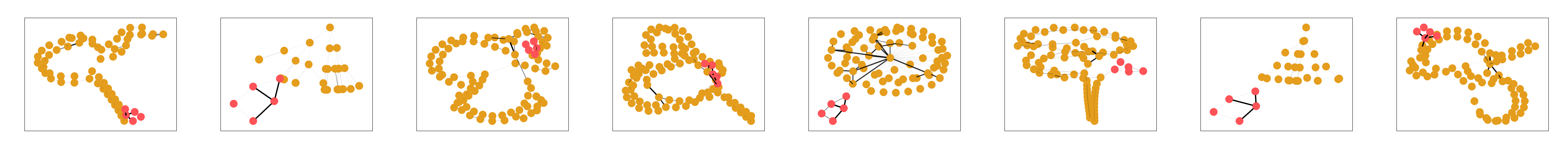}
    }
    \subfigure[Spurious-Motif class $1$ under bias$=0.5$ by \ourss.]{
        \includegraphics[width=\textwidth]{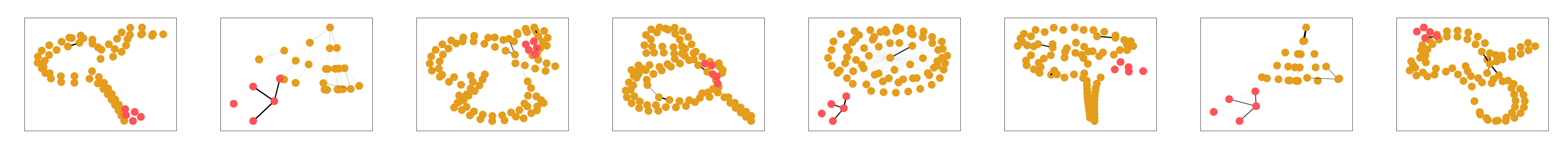}
    }
    \subfigure[Spurious-Motif class $2$ under bias$=0.5$ by \gsat.]{
        \includegraphics[width=\textwidth]{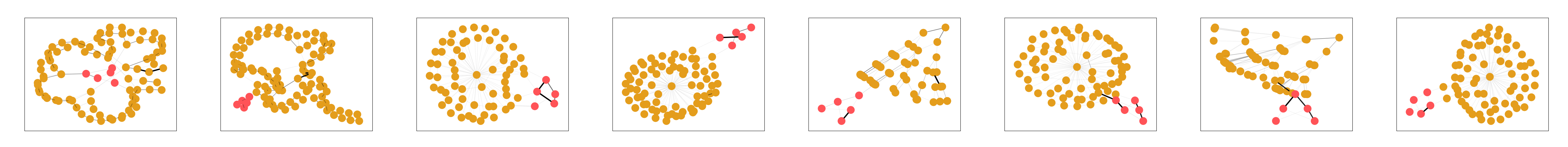}
    }
    \subfigure[Spurious-Motif class $2$ under bias$=0.5$ by \ourss.]{
        \includegraphics[width=\textwidth]{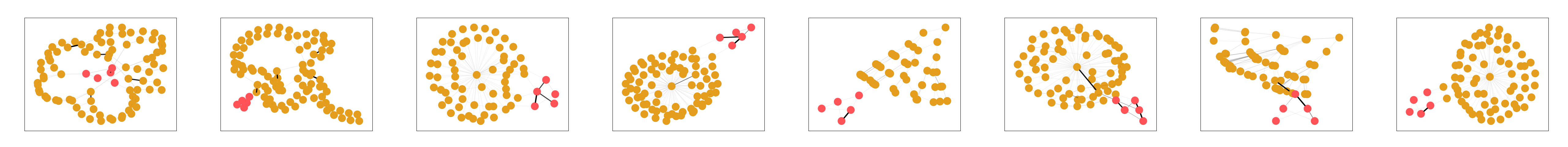}
    }
    \caption{
        Learned interpretable subgraphs by \gsat and \ourss on Spurious-Motif $b=0.9$.}
    \label{fig:sp9_viz_appdx}
\end{figure}

The results on Spurious-Motif datasets are given in Fig.~\ref{fig:sp5_viz_appdx},~\ref{fig:sp7_viz_appdx},\ref{fig:sp9_viz_appdx} for $b=0.5$, $b=0.7$ and $b=0.9$, respectively. The red nodes are the ground-truth interpretable subgraphs. It can be found that \ourss indeed learns the interpretable subgraph better than \gsat, which also explains the excellent OOD generalization ability of \ourss on Spurious Motif datasets.

\begin{figure}[H]
    \centering
    \subfigure[OGBG-Molhiv class $0$ by \ourss.]{
        \includegraphics[width=\textwidth]{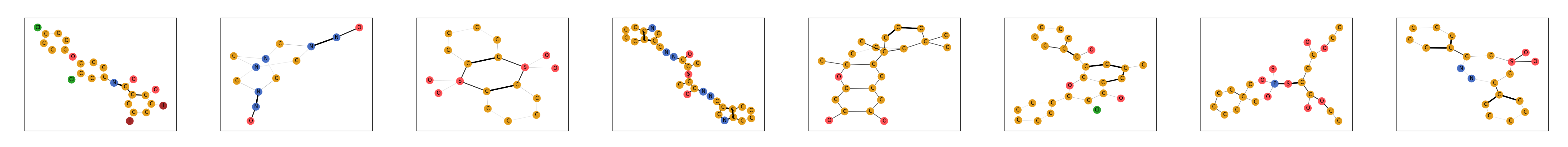}
    }
    \subfigure[OGBG-Molhiv class $1$ by \ourss.]{
        \includegraphics[width=\textwidth]{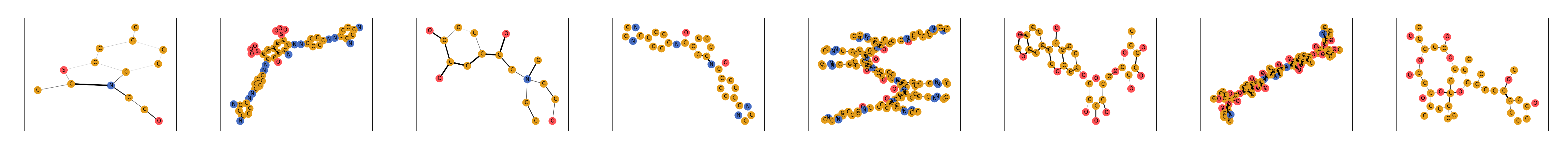}
    }
    \caption{
        Learned interpretable subgraphs by \ourss on OGBG-Molhiv.}
    \label{fig:oh_viz_appdx}
\end{figure}

In addition, we also provide the visualization of interpretable subgraphs learned by \ourss on OGBG-Molhiv, given in Fig.~\ref{fig:oh_viz_appdx}.

